\documentclass[12pt]{article}

\usepackage{scicite}

\usepackage{url}
\usepackage{booktabs}
\usepackage{hyperref}

\usepackage{times}
\usepackage{amsmath}
\usepackage{amsfonts}
\usepackage{amssymb,amsthm}
\usepackage{mathrsfs}
\usepackage{indentfirst,latexsym,bm}
\usepackage[hang]{subfigure}
\usepackage{graphicx}
\usepackage{listings}
\usepackage{multirow,tabularx}
\usepackage{xcolor}
\usepackage{pict2e}
\usepackage[all, cmtip]{xy}
\usepackage{makecell}
\usepackage{enumerate}
\usepackage{longtable}
\usepackage[flushleft]{threeparttable}
\usepackage[mathlines]{lineno}
\usepackage{cuted}
\newcommand{\eps}{\varepsilon}
\renewcommand{\vec}[1]{\boldsymbol{#1}}
\renewcommand{\d}{\mathop{}\!\mathrm{d}}
\newcommand{\eqnref}[1]{Eq.~\eqref{#1}}
\newcommand{\figref}[1]{Fig.~\ref{#1}}
\newcommand{\tabref}[1]{Table~\ref{#1}}

\newtheorem{theorem}{Theorem}
\newtheorem{definition}{Definition}
\newenvironment{sciabstract}{%
\begin{quote} \bf}
{\end{quote}}

\title{Deciphering Interventional Dynamical Causality from Non-intervention Complex Systems}

\newcommand{\authorwrap}[1]{%
	\begin{minipage}{\textwidth}
		\centering
		#1
	\end{minipage}
}

\author
{\authorwrap{Jifan Shi$^{1\ast}$, Yang Li$^{2}$, Juan Zhao$^{3}$, Siyang Leng$^{1,4}$, Rui Bao$^{5}$, Kazuyuki Aihara$^{2\ast}$, Luonan Chen$^{6,7\ast}$, Wei Lin$^{1\ast}$}\\
\\
\normalsize{\authorwrap{$^{1}$Research Institute of Intelligent Complex Systems \& CISOR, Fudan University, Shanghai 200433, China}}\\
\normalsize{\authorwrap{$^{2}$International Research Center for Neurointelligence, The University of Tokyo Institutes for Advanced Study, The University of Tokyo, Tokyo 113-0033, Japan}}\\
\normalsize{\authorwrap{$^{3}$School of Pharmacy, Shanghai University of Traditional Chinese Medicine, Shanghai 201203, China}}\\
\normalsize{\authorwrap{$^{4}$Institute of AI and Robotics, College of Intelligent Robotics and Advanced Manufacturing, Fudan University, Shanghai 200433, China}}\\
\normalsize{\authorwrap{$^{5}$Frontiers Science Center for Deep Ocean Multispheres and Earth System, Key Laboratory of Marine Chemistry Theory and Technology, Ministry of Education, Ocean University of China, Qingdao 266100, China}}\\
\normalsize{\authorwrap{$^{6}$School of Mathematical Sciences and School of AI, Shanghai Jiao Tong University, Shanghai 200240, China}}\\
\normalsize{\authorwrap{$^{7}$Key Laboratory of Systems Health Science of Zhejiang Province, Hangzhou Institute for Advanced Study, University of Chinese Academy of Sciences, Chinese Academy of Sciences,  Hangzhou 310024, China}}\\
\\
\normalsize{\authorwrap{$^\ast$To whom correspondence should be addressed.\\ E-mail:  jfshi@fudan.edu.cn; kaihara@g.ecc.u-tokyo.ac.jp; lnchen@sibs.ac.cn; wlin@fudan.edu.cn.}}
}

\date{}

\begin{document} 


\baselineskip24pt


\maketitle 

%\linenumbers


\begin{sciabstract}
	Detecting and quantifying causality is a focal topic in the fields of science, engineering, and interdisciplinary studies. However, causal studies on non-intervention systems attract much attention but remain extremely challenging. Delay-embedding technique provides a promising approach. In this study, we propose a framework named Interventional Dynamical Causality (IntDC) in contrast to the traditional Constructive Dynamical Causality (ConDC). ConDC, including Granger causality, transfer entropy and convergence of cross-mapping, measures the causality by constructing a dynamical model without considering interventions. A computational criterion, Interventional Embedding Entropy (IEE), is proposed to measure causal strengths in an interventional manner. IEE is an intervened causal information flow but in the delay-embedding space. Further, the IEE theoretically and numerically enables the deciphering of IntDC solely from observational (non-interventional) time-series data, without requiring any knowledge of dynamical models or real interventions in the considered system. In particular, IEE can be applied to rank causal effects according to their importance and construct causal networks from data. 
	We conducted numerical experiments on Logistic dynamics, coupled-Henon maps, and chaotic neural networks to demonstrate that IEE can find causal edges accurately, eliminate effects of confounding, and quantify causal strength robustly over traditional indices. We also applied IEE to real-world tasks, including estimating neural connectomes of \textit{C. elegans}, detecting COVID-19 transmission networks in Japan, and investigating regulatory networks surrounding key circadian genes. IEE performed as an accurate and robust tool for causal analyses solely from the observational data.  The IntDC framework and IEE algorithm provide an efficient approach to the study of causality from time series in diverse non-intervention complex systems.
\end{sciabstract}

% In setting up this template for *Science* papers, we've used both
% the \section* command and the \paragraph* command for topical
% divisions.  Which you use will of course depend on the type of paper
% you're writing.  Review Articles tend to have displayed headings, for
% which \section* is more appropriate; Research Articles, when they have
% formal topical divisions at all, tend to signal them with bold text
% that runs into the paragraph, for which \paragraph* is the right
% choice.  Either way, use the asterisk (*) modifier, as shown, to

\section{Introduction}

Directional or indirectional interactions among different components give rise to a variety of complex phenomena in nature and social society. The causality or the causal effect is one of the most attractive directional relations originating from the non-reversibility of time. Modeling, detecting, and quantifying causality from observational data are crucial for describing, interpreting, predicting, and even controlling complex systems. 

Statisticians believe that causality is contained and obtainable in universal random variables regardless of the time label. Neyman-Rubin's potential outcome framework \cite{neyman1923application, rubin1978bayesian, sekhon2008neyman}, the Wright's structural equation model \cite{wright1921correlation,wright1934method} and the Pearl's causal diagram model \cite{pearl1995causal, pearl2009causal} are the most famous statistical approaches, which have been proved to be of mathematical equivalence \cite{pearl2000models}. Especially, the instrumental variable method is widely applied in the study of causality among economics, environments, and other disciplines \cite{angrist1996identification}. Statistical methods seek to discern binary causal relations among random variables on a directed acyclic graph, without relying on temporal data. However, in general complex dynamical systems, the causation must precede the effect, feedback is common, and the causal strength needs to be quantitatively measured. Time should play definitely a key role.

Dynamics-oriented researchers have proposed fruitful algorithms for measuring the observational causality from time series, commonly referred to as the dynamical causality \cite{shi2022embedding}. The celebrated Granger causality (GC) \cite{granger1969investigating, lusch2016inferring} employs a linear model and uses improvement of predictability over time to illustrate the causality. Transfer entropy (TE) \cite{schreiber2000measuring} generalizes GC to the nonlinear case by quantifying the prediction uncertainty through Shannon entropy. Neither GC nor TE addresses the ``non-separability'' problem, which means that removing the causal variable from the system inevitably influences the dynamics of downstream variables \cite{sugihara2012detecting,  leng2020partial, ying2022continuity, shi2022embedding, liang2008information, liang2016information}. To measure causality in non-separable systems, numerous approaches have emerged in the last decade within the framework of delay embedding. These include convergence of cross mapping (CCM) \cite{sugihara2012detecting, ye2015distinguishing}, partial cross mapping (PCM) \cite{leng2020partial}, continuity scaling \cite{ying2022continuity}, inverse continuity \cite{pecora1995statistics}, topological expansion \cite{harnack2017topological}, joint distance distribution \cite{amigo2018detecting}, embedding entropy (EE) \cite{shi2022embedding} and other indices \cite{stavroglou2020unveiling, tao2023detecting}. Besides, Runge et al. designed the PCMCI method to further remove ``common drivers'' by selecting parent nodes iteratively\cite{runge2019detecting}. Friston et al. proposed the dynamical causal modeling (DCM), which is a Bayesian fitting from data with pre-selected models \cite{friston2003dynamic}. DCM was originally developed for modeling neural dynamics \cite{stephan2007dynamic}. Krakovsk\'{a} et al. conducted a comparative analysis of six methods for detecting causality in bivariate systems\cite{krakovska2018comparison}.

Nonetheless, most indices designed for quantifying causality from time series primarily focus on estimating directional causal relationships at the constructive level, referred to as constructive dynamical causality (ConDC). Traditionally, to detect and quantify causality at the interventional level, which is termed interventional dynamical causality (IntDC) in this study, requires intelligent modulation and manipulation of the dynamical system. By allowing external intervention to the system and recording data under different perturbations, frameworks such as perturbation cascade inference (PCI) \cite{stepaniants2020inferring} and dynamical causal effect (DCE) \cite{smirnov2014quantification, smirnov2020transfer} provide relevant computational schemes. To detect asymmetry information transfer in known two-dimensional dynamical systems, Liang and Kleeman proposed an analytic approach named Liang-Kleeman information flow \cite{liang2005information, liang2014unraveling, liang2016information}, which serves as a prototype for IntDC. This method quantifies causality by freezing one variable as a parameter, and evaluating the resulting outcomes using the Frobenius-Perron operator. The Liang information flow has been further extended to multivariate, stochastic, and quantum systems \cite{liang2021normalized, hristopulos2024information, yi2022quantum, pires2024general}. However, due to ethical or practical limitations, many real systems should be analyzed without any external intervention. An essential problem is how to measure IntDC solely from the observational data but at the interventional level.

In this study, we propose the IntDC framework and introduce a criterion named Interventional Embedding Entropy (IEE), which aims to identify and quantify IntDC between variables solely from the observational time series. Actually, IEE is rigorously derived by the theory of the delay embedding. Numerical IEE does not require specific prior knowledge of dynamics, and additional perturbation to the system is also unnecessary. Compared to ConDC indices, such as GC, TE, and CCM, IEE designed for IntDC has the capability to rank the importance of causal effects and construct directional causal networks more effectively. We demonstrate numerical experiments on both simulated examples and real datasets, including estimating neural connectomes of \textit{C.\,elegans}, evaluating COVID-19 transmission in Japan, and constructing regulatory networks surrounding key circadian genes.

\section{Methods}

\subsection{Constructive dynamical causality}
The definition of ConDC for a complex system in the original time-series space is given as follows:
\begin{definition}[Constructive dynamical causality, ConDC \cite{shi2022embedding}]\label{def1}
	For a complex dynamical system
	\begin{align}
	\vec{x}_{t+1} = \vec{f}(\vec{x}_t, \vec{x}_{t-1}, \dots, \vec{x}_{t-p}) +\vec{\eps}_t, \label{eq_multix}
	\end{align}
	where $\vec{x} = (x^{(1)}, x^{(2)}, \dots, x^{(n)})^T$ represents the system with $n$ components, $\vec{f} = (f_1, f_2, \dots, f_n)^T$ is a vector function, $p$ is referred to the memory time, and $\vec{\eps}_t$ is an independent noise term, there exists ConDC from component $x^{(j)}$ to $x^{(i)}$, if $\exists k\in \{1, 2, \dots, p\}$ such that $\partial f_i/\partial x^{(j)}_{t-k} \neq 0$ for almost any $t$.
\end{definition}
For simplicity, we consider a discrete two-variable dynamical system as an example:
\begin{equation}
\begin{cases}
x_{t+1} = g(x_t, x_{t-1}, \dots, x_{t-p}, \varepsilon_{x,t}),\\
y_{t+1} = f(x_t, x_{t-1}, \dots, x_{t-p}, y_t, y_{t-1}, \dots, y_{t-p}, \varepsilon_{y,t}),
\end{cases}\label{eq_xy}
\end{equation}
where $x, y$ are two variables, $p$ denotes the time step during which causality is considered,  and $\varepsilon_{\cdot, t}$ stand for small noise terms. According to Definition \ref{def1}, there exists ConDC from $x$ to $y$ since the evolving equation of $y_{t+1}$ depends on the historical behavior of $x$, while there is no ConDC from $y$ to $x$ as the dynamics of $x_{t+1}$ is independent of $y$. GC and TE measure the predictability of $f$ in the original time-series space to quantify the ConDC from $x$ to $y$ through linear regression and entropy uncertainty, respectively (\figref{fig1_main}(B)). The causal strength of GC/TE is based on the assumption of separability, which means that removing $x$ from the system does not influence the observed data of $y$. However, in general coupled systems, the separability is not satisfied, as removing one variable changes the values of affected ones \cite{sugihara2012detecting, leng2020partial, ying2022continuity}.

To address the challenge of detecting ConDC in universal non-separable systems, various approaches from the delay-embedding space have been proposed. In the delay-embedding framework, the bivariate system \eqnref{eq_xy} is assumed to evolve into an attractive manifold with an inner dimension $d$. 
Let the time-delayed vectors of $x$ and $y$ be 
\begin{equation}
\vec{X}_t = (x_t, x_{t-1}, \dots, x_{t-L})^T\in\mathcal{M}_X\subseteq \mathbb{R}^{L+1}, \label{eq_delayX}
\end{equation}
and
\begin{equation}
\vec{Y}_{t+1} = (y_{t+1}, y_t, y_{t-1}, \dots, y_{t-L+1})^T\in \mathcal{M}_Y\subseteq\mathbb{R}^{L+1}, \label{eq_delayY}
\end{equation}
where $L$ is the time-delayed length, $\mathcal{M}_X$ and $\mathcal{M}_Y$ represent the manifolds formed by $\vec{X}_t$ and $\vec{Y}_t$, respectively. According to the seminal stochastic version of Takens' embedding theorem \cite{takens1981detecting, sauer1991embedology, stark1997takens, cummins2015efficacy}, we can obtain the following theorem:
\begin{theorem}[ConDC in delay-embedding space]\label{thm1}
	If $x$ is the ConDC of $y$ in dynamics \eqnref{eq_xy}, and $\vec{X}_t$, $\vec{Y}_t$ are the time-delayed vectors, respectively, then there exists a smooth projection operator $\vec{F}$  in generic sense such that
	\begin{equation}
	\vec{X}_t = \vec{F}(\vec{Y}_{t+1}), \label{eq_proYX}
	\end{equation}
	when the time-delayed length satisfies $L\geqslant 2d$, where $d$ is the inner dimension of the attractive manifold.
\end{theorem}
The detailed derivation can be referred to the Supplementary Text. The ``generic sense'' means that a smooth projection exists for a dense and open set of all possible time-delayed ways\cite{takens1981detecting, sauer1991embedology}. Equation \eqref{eq_proYX} shows that the causal variable $\vec{X}_t\in \mathcal{M}_X$ can be reconstructed by the effect variable $\vec{Y}_{t+1}\in\mathcal{M}_Y$. However, as $\vec{F}$ is not reversible in generic sense, we can not determine $\vec{Y}_{t+1}$ only by the information from $\vec{X}_t$. According to Theorem \ref{thm1}, the causal dependence in $f$ between $x$ and $y$ is transformed into the reconstructability of $\vec{F}$ (\figref{fig1_main}(D)). Theorem\ref{thm1}, established for the dynamical system in \eqnref{eq_xy}, can be generalized to multi-variable systems as described in \eqnref{eq_multix}, allowing for feedback interactions between different variables. Instead of fitting a model by removing the causal variable as in GC and TE, detecting the existence and quantifying the continuity characteristics of $\vec{F}$ in the delay-embedding space can be adequate for causal identification, especially for universal non-separable systems. Related algorithms include CCM, PCM, and EE.

Figures~\ref{fig1_main}(A), (B) and (D) summarize dynamical causality at the constructive level, where the ConDC from variable $x$ to $y$ is estimated only from the historical observed time series.  But fitting \eqnref{eq_xy} or \eqnref{eq_proYX} from data is criticized as mere association or prediction \cite{altman2015points,shojaie2022granger}.

\begin{figure*}[!h]
	\centering
	\includegraphics[width = 0.92\textwidth]{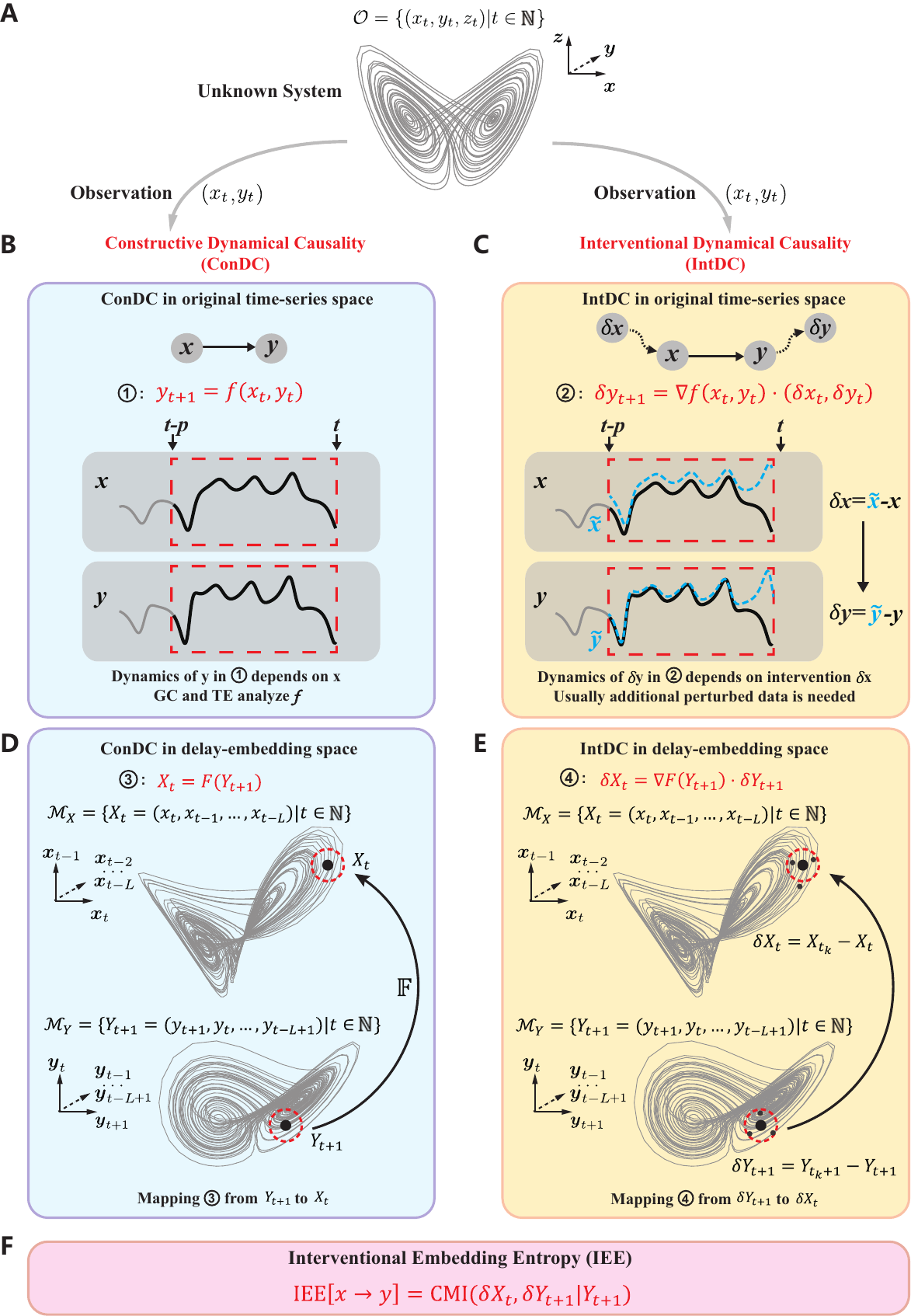}
\end{figure*}
\begin{figure}[t]%\ContinuedFloat
	\caption{Illustration for the constructive dynamical causality (ConDC) and interventional dynamical causality (IntDC). (A) An illustrative example of a complex system where $(x_t, y_t)$ are observed time series. The $z_t$ stands for the other variable that is not the primary focus of analysis. (B) ConDC from $x$ to $y$ indicates that the dynamics of $y$ depends on $x$. Algorithms such as Granger causality (GC) and transfer entropy (TE) infer the ConDC in the original time-series space. (C) IntDC from $x$ to $y$ indicates that the deviation of $y$, i.e. $\delta y$, depends on an intervention on $x$, i.e. $\delta x$. Usually additional data from a perturbed system $(\tilde{x}_t, \tilde{y}_t)$ is necessary for detecting IntDC in the original time-series space. The parameter $p$ is the time lag after intervention occurs. (D) ConDC can be detected in the embedding space by cross mapping $\mathbb{F}$. The causal vector $\vec{X}_t$ can be reconstructed by the effect vector $\vec{Y}_{t+1}$. Parameter $L$ is time delay used in the delay-embedding. (E) IntDC can be modeled in the embedding space. The interventional causal vector $\delta \vec{X}_t$ should be reconstructable by the effect vector $\delta \vec{Y}_{t+1}$ around $\vec{Y}_{t+1}$. In numerical computation, $\delta \vec{Y}_{t+1}$ is approximated by the distance from $\vec{Y}_{t+1}$ to its neighbors $\vec{Y}_{t_k+1}$, while $\delta \vec{X}_t$ is determined using $\vec{X}_{t_k}$ corresponding to the same time index $t_k$. (F) Interventional embedding entropy (IEE) measures the average information retention contained in $\delta \vec{Y}_{t+1}$ from $\delta \vec{X}_t$, quantifying the IntDC from variable $x$ to $y$. Notably, IEE does not require additional perturbations to the system and can be inferred from observational data solely.} \label{fig1_main}
\end{figure}

\subsection{Interventional dynamical causality}
The detection of hidden essential-level causality necessitates intervention or manipulation to the system \cite{pearl2000models,stepaniants2020inferring,	korb2004varieties, amblard2012relation}. In this study, we propose the definition of IntDC for a complex system in the original time-series space as the following:
	\begin{definition}[Interventional dynamical causality, IntDC]
		For a complex dynamical system \eqnref{eq_multix}, one component $x^{(j)}$ is the IntDC of another component $x^{(i)}$, if $\exists k\in \{1, 2, \dots, p\}$ such that $\delta x^{(i)}_{t+1} = \tilde{x}^{(i)}_{t+1} - x^{(i)}_{t+1}$ depends on $\delta x^{(j)}_{t-k} = \tilde{x}^{(j)}_{t-k} - x^{(j)}_{t-k}$ for almost any $t$, where $\tilde{x}^{(j)}_{t-k}$ is the intervened dynamics of $x^{(j)}$, and $\tilde{x}^{(i)}_{t+1}$ is the dynamics of affected $x^{(i)}$ after the intervention on $x^{(j)}$.
	\end{definition}
	\noindent The intervention $\delta x^{(j)}_{t-k}$ can be either a pulse stimulation or  persistent disturbances to the system.

\figref{fig1_main}(C) illustrates the IntDC in the time-series space for the simplified bivariate system \eqnref{eq_xy}, involving variables $x$ and $y$. To measure the IntDC, data before and after the intervention, i.e. $(x_t, y_t)$ and $(\tilde{x}_t, \tilde{y}_t)$, are typically required, as in PCI and DCE. PCI measures the distances of  every node in the network from each perturbed node to reconstruct the directed causal network \cite{stepaniants2020inferring}. DCE classified causal coupling into four situations based on the type of effects (stationary statistic change, phase orbit change) and the type of interventions (state space interventions, parametric interventions) \cite{smirnov2014quantification}.
These conventional algorithms rely on reproducible dynamics or data availability under different settings.

To address challenges of causal inference in widespread non-intervention, non-linear and non-separable systems, we turn to the study of intervened dynamics of \eqnref{eq_proYX} in the delay-embedding space. Denote $\delta\vec{X}_t = \widetilde{\vec{X}}_t -\vec{X}_t$ and $\delta\vec{Y}_{t+1} = \widetilde{\vec{Y}}_{t+1} - \vec{Y}_{t+1}$, where $\widetilde{\vec{X}}_t$ and $\widetilde{\vec{Y}}_{t+1}$ are the time-delayed vectors of $x$ and $y$ after the intervention, respectively. We can obtain the following theorem:
\begin{theorem}[IntDC in delay-embedding space]\label{thm2}
	If $x$ is the IntDC of $y$ in dynamics \eqnref{eq_xy}, and $\vec{X}_t, \vec{Y}_t$ are the time-delayed vectors, respectively, then there exists a smooth projection operator $\vec{F}$ in generic sense such that
	\begin{equation}
	\delta \vec{X}_t = \nabla_{\vec{Y}}\vec{F}(\vec{Y}_{t+1})\cdot \delta\vec{Y}_{t+1}, \label{eq_IntDE}
	\end{equation}
	when the time-delayed length satisfies $L\geq 2d$ and the intervention is sufficiently small, where $d$ is the inner dimension of the attractive manifold.
\end{theorem}
Theorem \ref{thm2} is directly deduced by Theorem \ref{thm1}. However, \eqnref{eq_IntDE} provides insights on the causality under intervention. It indicates that $\delta\vec{Y}_{t+1}$ contains complete information of $\delta\vec{X}_t$ in the neighborhood of $\vec{Y}_{t+1}$ and can reconstruct $\delta\vec{X}_t$, if there exists IntDC from the causal variable $x$ to the effect variable $y$ (\figref{fig1_main}(E)). The conclusion in \eqnref{eq_IntDE} of Theorem \ref{thm2}, derived from the dynamics in \eqnref{eq_xy}, remains valid for any pair of components exhibiting IntDC in the multi-variable system of \eqnref{eq_multix}, and feedback between different components is permitted. Figures~\ref{fig1_main}(A), (C), and (E) give an overview of the IntDC framework.

\subsection{Interventional embedding entropy}
We introduce the IEE criterion, specifically designed for quantifying IntDC solely from the observational data in the delay-embedding space.
\begin{definition}[Interventional embedding entropy, IEE]
	The IEE criterion is proposed to measure the average information retention or reconstructability from the effect $\delta \vec{Y}_{t+1}$ to the causation $\delta \vec{X}_t$, formulated  as
		\begin{equation}
		\mathrm{IEE}[x\rightarrow y] := \mathrm{CMI}(\delta\vec{X}_t, \delta\vec{Y}_{t+1}| \vec{Y}_{t+1}). \label{eq_cs1}
		\end{equation}
\end{definition}
\noindent In \eqnref{eq_cs1}, CMI is the conditional mutual information \cite{cover1999elements}, i.e.
	\begin{equation}
	\text{CMI}(\vec{x}, \vec{y} | \vec{z}) = \iiint p(\vec{x}, \vec{y}, \vec{z})\log \frac{p(\vec{x}, \vec{y}|\vec{z})}{p(\vec{x}|\vec{z})p(\vec{y}|\vec{z})}\d \vec{x}\d\vec{y}\d\vec{z}.\label{eq_eqCMI}
	\end{equation}
IEE provides is a quantitative measure of causal strength that is comparable across different edges within the same dynamical system. When dealing with data, the $\delta \vec{Y}_{t+1}$ is estimated by the distance from $\vec{Y}_{t+1}$ to its neighbored points $\vec{Y}_{t_k+1}$ in the embedding space, and  $\delta \vec{X}_t$ is calculated by $\vec{X}_{t_k}$ with the same time label $t_k$ (Figs.~\ref{fig1_main}(E) and (F)). Thus, \eqnref{eq_cs1} can be numerically approximated only from the observational data, especially for non-intervention systems. 
\tabref{tab1_IEEalg} presents the numerical algorithm for computing IEE, whose detailed descriptions and computational procedures are provided in the Supplementary Text.

\subsection{Connections with statistical causal models and assumptions of ConDC/IntDC}
The concept of dynamical causality in \eqnref{eq_multix} develops from traditional structural causal models, which typically studies structure equations of the form $Y = f(X, \varepsilon, \theta)$, where $X$ and $Y$ are random variable, $f$ is a function parameterized by $\theta$, and $\varepsilon$ represents noise or residue terms. In \cite{shi2022embedding}, the framework of ConDC is systematically studied with three key generalizations beyond classical structural equation modeling: (i) Temporal information is fully considered by dynamical systems, ensuring that causes precede effects in time; (ii) Feedback or causal loops between variables are permitted under the consideration of time delays, relaxing the restriction of directed acyclic graphs (DAGs); (iii) Quantitative measures of causal strength are provided beyond qualitative identification, enabling direct comparisons of causality between different variables. The IntDC proposed in the study further integrates the concept of interventions and potential outcomes within the dynamical framework\cite{pearl2000models}. IntDC aims to quantify causal strengths at the interventional level directly from observational data (without the need for additional interventions to the system, which are often impractical or infeasible in real-world datasets).

The following assumptions (A1)-(A8) clarify key connections between dynamical causality and statistical causal inference:

(A1) Temporal Order assumption: Causal information is embedded in time-series data and causes must precede their effects;

(A2) Causal Sufficiency (Unconfoundedness) assumption: All common drivers (confounders) are observed such that the causal graph accurately represents relationships among the observed variables;

(A3) Causal Markov assumption: Each variable $x_i$ is independent of its nondescendants given its direct causes (parents);

(A4) Faithfulness assumption: The causal graph structure can represent the conditional independence contained in the joint probability density;

(A5) Stationarity assumption: The dynamical system evolves into a stable attractive manifold, allowing stable causal relations to be measured from observational data;

(A6) Intervention Ignorability assumption: Given the system dynamics \eqnref{eq_multix}, initial conditions $\vec{x}_0$ and noise $\vec{\eps}_t$, the intervention on variable $x^{(j)}$ at time $t-k$ is independent of the potential outcome of $x^{(i)}$ at time $t+1$, i.e. $x^{(i)}_{t+1}\text{$\perp\!\!\!\perp$} \delta x^{(j)}_{t-k}  | (\vec{f}, \vec{x}_0, \vec{\eps}_t)$ for $k = 1, 2, \dots, p$;

(A7) Stable Infinitesimal Intervention assumption: The average information retention from $x^{(j)}$ to $x^{(i)}$ in dynamics \eqnref{eq_multix} remains stable under sufficiently small interventions $\delta x^{(j)}$;

(A8) Consistency assumption: The outcome $\tilde{x}^{(i)}_{t+1}$ in \eqnref{eq_multix} can be precisely determined under known dynamics $(\vec{f}, \vec{x}_0, \vec{\eps}_t)$ and a fixed intervention $\delta x^{(j)}_{t-k}$.

In practice, (A2) can be relaxed if the primary goal is to quantify the total causal strength (including both direct and indirect causality) between observed variables. In the Results section, we demonstrate that the IEE criterion is capable of mitigating the impact of confounding variables, thereby providing robust causal estimates even when some confounders are unobserved. Some of these assumptions for causal discovery from observational time-series data have also been discussed by Runge et al. (\cite{runge2019detecting, runge2019inferring}). 

GC, TE, and CCM are all criteria for ConDC, which estimate causal strengths by fitting constructive dynamical models (i.e. \eqnref{eq_xy} in the time-series space or \eqnref{eq_proYX} in the delay-embedding space). GC employs linear vector regression to fit $\vec{f}$ in \eqnref{eq_xy}. TE extends GC to nonlinear cases by quantifying information transfer between variables. CCM constructs a local-linear cross-mapping $\vec{F}$ in \eqnref{eq_proYX} by the time-delayed embedding to capture causality. For a detailed comparison of these ConDC methods, readers are referred to Table 1 in \cite{shi2022embedding}.

\section{Results}
\subsection{IEE captures the numerical behavior of causality in Logistic systems robustly}
To assess the numerical performance of measuring IntDC, we first applied the IEE algorithm on the following two-node Logistic dynamics

	\begin{equation}
	\begin{cases}
	x_{t+1} = 3.7 \big[ (1-\beta_{yx})x_t\left(1-x_t\right) + \beta_{yx}y_t\left(1-y_t\right) \big] + \eps_{x, t},\\
	y_{t+1} = 3.7 y_t \big[ 1 - (1-\beta_{xy})y_t - \beta_{xy}x_t \big] + \eps_{y, t},
	\end{cases}
	\end{equation}

\noindent where $\eps_{\cdot, t}$ represents independent Gaussian noises, and parameters $\beta_{xy}$ and $\beta_{yx}$ modulate the IntDC strength from $x$ to $y$ and from $y$ to $x$, respectively (see further details in SM).

IEE can capture the numerical behavior of IntDC in the two-node Logistic system. When we set $\beta_{xy} \equiv  0$ and let $\beta_{yx}$ increase from $0$ to $0.3$. we observed a monotonic increase in $\text{IEE}[y\rightarrow x]$ (the red line with squares in \figref{fig_2log}(A)), while $\text{IEE}[x\rightarrow y]$ remained around zero (the blue line with dots in \figref{fig_2log}(A)).  When we set $\beta_{xy} \equiv 0.1$ and let $\beta_{yx}$ increase from $0$ to $0.3$, $\text{IEE}[y\rightarrow x]$ exhibited a monotonic increase (the red line with squares in \figref{fig_2log}(B)), while $\text{IEE}[x\rightarrow y]$ stayed above zero (the blue line with dots in \figref{fig_2log}(B)). The gray dashed lines in Figs.~\ref{fig_2log}(A-B) represent $0.01$ for reference. A detailed comparison with GC, TE and CCM is presented in figs.S1 and S2. The mean values and standard deviations of the indices over 100 simulation runs are provided in tables~S1-S4. Fig.~S1(b) and fig.~S2(b) indicate that GC exhibited non-monotonic behavior. $\text{TE}[y\rightarrow x]$ demonstrated a decreasing trend when $\beta_{yx}$ increased, specifically for $\beta_{yx}\in (0, 0.02]$ in fig.~S1(c)/table~S1 and $\beta_{yx}\in (0, 0.03]$ in fig.~S2(c)/table~S3, which may lead to false-negative issues (ignorance of existence causality). $\text{CCM}[x\rightarrow y]$ remained significantly different from zero even when $\beta_{xy}=0$ (in the range $\beta_{yx}\in[0.15, 0.3]$ in fig.~S1(d)/table~S2). $\text{CCM}[y\rightarrow x]$ was also significantly different from zero when $\beta_{yx}=0$ (with $\beta_{xy}=0.1$ in fig.~S2(d)/table~S3). These results suggest that CCM may suffer from the false-positive problem (incorrect identification of non-existence causality).

Within suitable parameter ranges, IEE demonstrates stability and robustness. We validated the robustness of the IEE algorithm under various conditions, including different delayed lags $L$, the numbers of nearest neighbors $K$, lengths of time series $N$, and noise standard deviations $\sigma$. The causal strengths measured by IEE exhibited a consistent descending trend as $\beta_{yx}$ decreased  from $0.15$ to $0.05$ through $0.125, 0.1, 0.075$, as expected (Figs.~\ref{fig_2log}(D-G)). These results support that IEE reliably preserves the correct ranking of causal strengths across different parameter settings with relatively stable variance. 

\subsection{IEE eliminates the influence of confounding variables}
\begin{figure}[!htbp]
	\centering
	\includegraphics[width=0.95\textwidth]{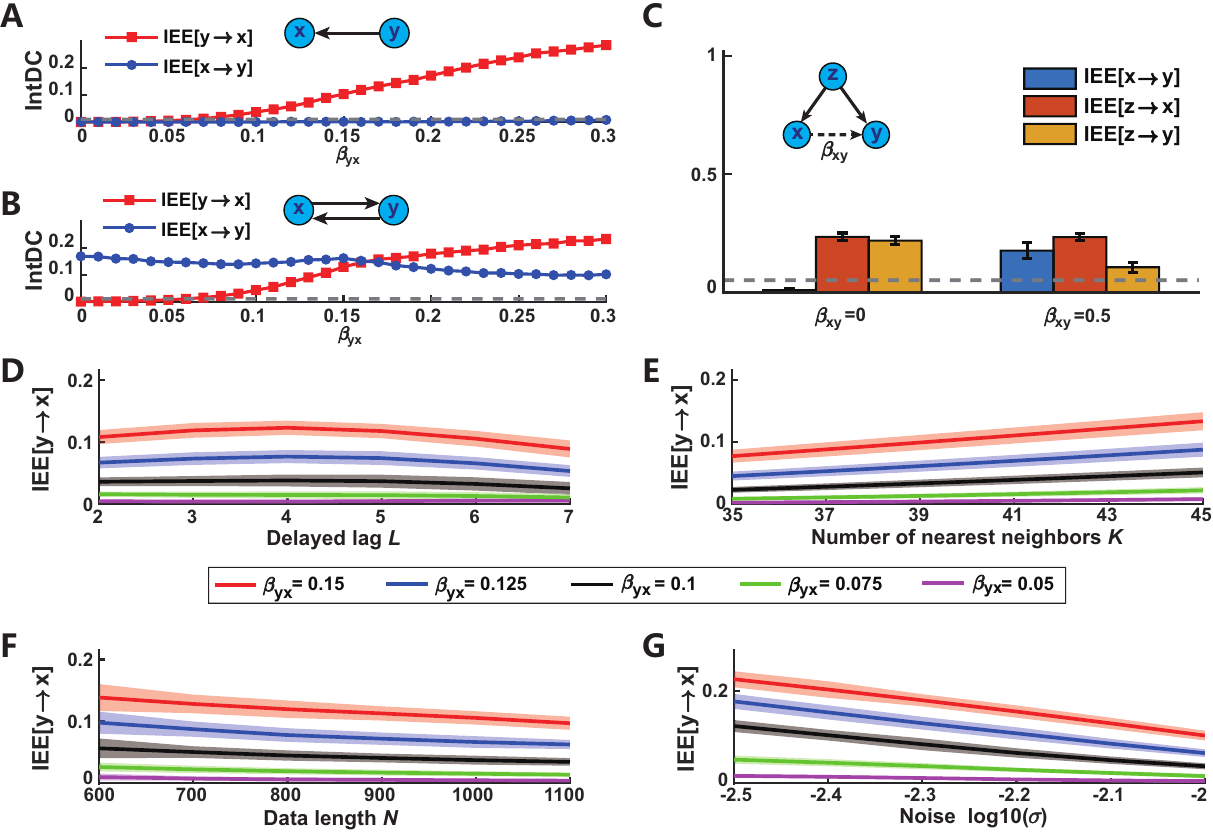}
	\caption{Performance of IEE on Logistic systems. (A) and (B) used the two-node Logistic system. The $\beta_{xy}$ and $\beta_{yx}$ control the coupling coefficients from $x$ to $y$ and from $y$ to $x$, respectively. (A) shows the mean values of IEE when $\beta_{xy}=0$ and $\beta_{yx}$ ranges from $0$ to $0.3$ over 100 simulations. $\text{IEE}[y\rightarrow x]$ (the red line with squares) increases monotonically, while $\text{IEE}[x\rightarrow y]$ (the blue line with dots) stays around zero. (B)  is under $\beta_{xy}=0.1$, in which $\text{IEE}[x\rightarrow y]$ (the blue line with dots) is significantly positive. The gray dashed lines in (A) and (B) represent the constant $0.01$ as reference. (C) shows the performance of IEE in the three-node Logistic system, where $z$ acts as a confounding variable. When $\beta_{xy}=0$, $\text{IEE}[x\rightarrow y]$ is almost zero correctly. IEE designed for measuring IntDC is not affected by confounders, which usually leads to false positives in ConDC. The results for $\beta_{xy}=0.5$ is shown for a comparison. The gray dashed line is $0.05$ for reference. (D-G) demonstrate the robustness of IEE under different delayed lags $L$, the numbers of nearest neighbors $K$, data lengths $N$, and noise standard deviations $\sigma$ in the two-node logistic system, respectively. The parameter $\beta_{xy}\equiv 0$, and $\beta_{yx}$ takes values of $0.15$ (red), $0.125$ (blue), $0.1$ (black), $0.075$ (green), and $0.05$ (magenta). Lines denote mean values under 50 simulations, and shaded areas represent the standard deviation. IEE consistently preserves the correct ranking of IntDC across various settings, with relatively stable variance.} \label{fig_2log}
\end{figure}

IEE remains applicable and accurate even in presence of confounding variables. We simulated a three-node Logistic dynamics $(x_t, y_t, z_t)$, where $z$ acted as a confounder (see details in SM). There were consistently  non-zero causal effects from $z$ to $x$ and from $z$ to $y$, while $\beta_{xy}$ adjusted the causal strength from $x$ to $y$ (\figref{fig_2log}(C)). When $\beta_{xy}=0$, $\text{IEE}[x\rightarrow y]$ was almost zero and significantly different from $\text{IEE}[z\rightarrow x]$ and $\text{IEE}[z\rightarrow y]$ (\figref{fig_2log}(C)); but conventional ConDC indices such as GC/TE/CCM produced false-positive causality from $x$ to $y$ due to the presence of confounder $z$ (fig.~S3). According to the IntDC framework, an intervention on $x$ (i.e. $\delta x$) will not induce an indirect change in $y$ (i.e. $\delta y$) through $z$; in other words, there is no intervention-induced pathway through $x\leftarrow z\rightarrow y$. However, traditional ConDC indices by fitting dynamical models may falsely infer an association between $x$ and $y$ due to the influence of the confounder $z$. IEE, specially designed for measuring IntDC, remained unaffected by the confounder $z$ and yielded accurate results.

\subsection{IEE enables causal network reconstruction with quantification of causal strengths}
\begin{figure}[!htbp]
	\centering
	\includegraphics[width = 0.95\textwidth]{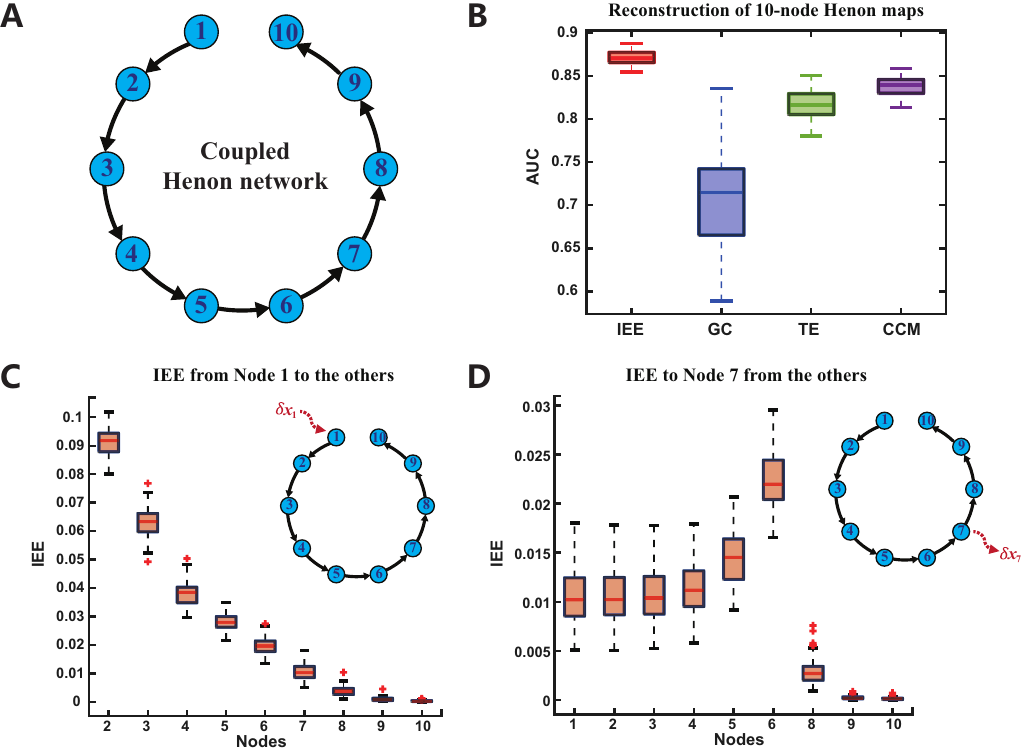}
	\caption{Performance of IEE on the 10-node coupled Henon-map dynamics. (A) illustrates the network structure of the dynamics. There are directed causality from each node to its subsequent node. (B) presents a boxplot of the Area Under Curve (AUC) values for the network reconstruction. IEE ($0.871\pm 0.008$) outperforms GC ($0.707\pm 0.069$), TE ($0.816\pm0.018$), and CCM ($0.837\pm0.012$) significantly in accuracy and stability. (C) demonstrates the IEE from Node 1 to the other nine nodes, while (D) shows the IEE received by Node 7 from the other nodes. IEE accurately captures the interventional information flow and IntDC in the cascade topological network.} \label{fig_3henon}
\end{figure}

IEE has the capability to reconstruct the causal network structure and rank the causal influence between nodes in complex networks at the interventional level. We used a 10-node coupled Henon maps as an example, where each node serves as the dynamical cause to its subsequent node (\figref{fig_3henon}(A)). Further details regarding the dynamics can be found in SM.

To verify the effectiveness of IEE in reconstructing causal networks, we calculated the Area Under Curve (AUC) values based on multiple simulated time series.  The AUC value (mean $\pm$ standard deviation) of IEE was $0.871\pm 0.008$,  significantly higher than conventional ConDC indices, such as GC with $0.707\pm 0.069$, TE with $0.816\pm0.018$, and CCM with $0.837\pm0.012$ (boxplots in \figref{fig_3henon}(B)). The AUCs of IEE exhibited both a higher mean and a lower variance relative to other approaches. Additionally, receiver operating characteristic (ROC) curves for the four indices were presented in fig.~S4.

To show the capability for ranking the importance of causal influences, we conducted two tests. We first calculated the IEE from Node 1 to the other nine nodes. Consistently with the actual scenario, IEE displayed sequential decrease in IntDC originating from Node 1 across Nodes 2 to 10 (\figref{fig_3henon}(C)). Then, we measured the IEE received by Node 7 from the other nine nodes. As expected, Nodes 1-6 exhibited ascending causal strengths on Node 7, whereas Nodes 8-10 had little influence on Node 7 (\figref{fig_3henon}(D)). IEE accurately discerned the IntDCs, with Node 6 showing the strongest value. A comparison showed that GC and TE failed to rank the influence from Node 1 accurately in the first test (fig.~S5), while CCM suffered from the false-positive causal detection from Node 8 to Node 7 in the second test (fig.~S6).

\subsection{IEE quantifies IntDC without requiring additional experimental interventions}
IEE is specifically designed for calculating IntDC solely from observed time-series data without requiring additional perturbations to the system. To validate its accuracy, we conducted a comparison against true perturbed deviations using chaotic neural networks (chNNs) as a model system\cite{aihara1990chaotic, adachi1997associative}. 

\begin{figure}[!htbp]
	\centering
	\includegraphics[width = 0.8\textwidth]{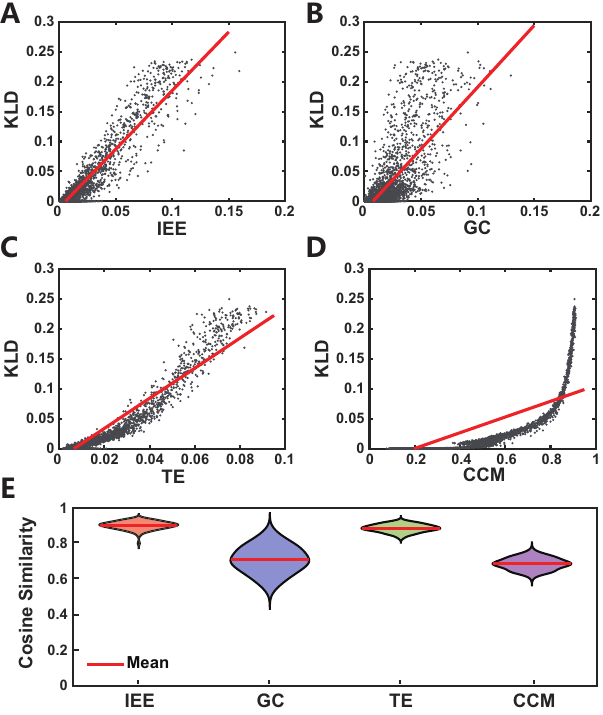}
	\caption{The performance of IEE when measuring the influence of perturbations on the chaotic neural networks (chNNs). The Kullback-Leibler divergence (KLD) is utilized to measure the true influence or IntDC between neurons. In (A-D) , $1000$ samples for different perturbations on chNNs are represented by gray dots, along with the linear regressions (the red lines) between KLD and IEE/GC/TE/CCM, respectively. (E) displays the violin plots of cosine similarity between KLD and IEE/GC/TE/CCM. The mean values and standard deviations of the cosine similarity are IEE ($0.904\pm0.024$),  GC ($0.711\pm0.076$), TE ($0.888\pm0.022$), and CCM ($0.687\pm0.033$). IEE can linearly reflect the KLD with high similarity, indicating its effectiveness in accurately quantifying IntDC without additional perturbed data on chNNs.} \label{fig_4chNN}
\end{figure}

The chNN comprises an output variable $\linebreak \vec{x} = (x_1, x_2, \dots, x_{N_{\text{node}}})^T$ and two internal variables, with $N_{\text{node}}$ representing the  number of neurons in the network. We chose $N_{\text{node}}=10$ in simulations. Details on the dynamics can be found in SM. We denoted the observed time-series data from the stationary system without intervention as $\vec{x}_{\text{obs}}$. When the neuron $i$ was removed (set as a constant zero), the perturbed data were denoted as $\vec{x}_{\text{per}}^{(i)}$. Removing one node was taken as an intervention to the system. The stationary probability density functions of the $j$th neuron before and after the intervention were represented by $p_j(\vec{x}_{\text{obs}})$ and $p_{j}(\vec{x}_{\text{per}}^{(i)})$, respectively. We used the Kullback-Leibler divergence (KLD), i.e.
\begin{align}
%\begin{aligned}
D_{ij}&\triangleq\text{KLD}[p_j(\vec{x}_{\text{obs}})||p_j(\vec{x}_{\text{per}}^{(i)})]\nonumber\\
& = \iint p_j(\vec{x}_{\text{obs}}) \log \frac{p_j(\vec{x}_{\text{obs}})}{p_j(\vec{x}_{\text{per}}^{(i)})}\d \vec{x}_{\text{obs}}\d\vec{x}_{\text{per}}^{(i)},
%\end{aligned}
\end{align}
to quantify the true influence of the intervention, i.e. ground truth of the IntDC,  from $i$ to $j$. Liang's Information flow adopts a similar intervention concept by treating a variable as fixed \cite{liang2014unraveling, liang2016information}. By conducting $100$ randomly simulated chNNs with interventions (in each simulation, one node is removed from each chNN with  $N_{\text{node}}$ nodes), we recorded $1000$ KLD values and compared them with the results of IEE (calculated from $\vec{x}_{\text{obs}}$ solely). We observed that $\text{IEE}[x_i\rightarrow x_j]$ exhibited a positive linear correlation with the KLD $D_{ij}$ (with $R^2 = 0.869$) in \figref{fig_4chNN}(A), where gray dots are the $1000$ samples of causal edges between neurons and the red line stands for the linear regression. Results for GC (with $R^2=0.572$), TE (with $R^2=0.912$), and CCM (with $R^2=0.622$) indicated that GC and CCM could not linearly reflect KLD (Figs.~\ref{fig_4chNN}(B-D)). Furthermore, we drew the violin plots of cosine similarity $S_c$ (see SM) between KLD and the four causal indices on $100$ chNNs (\figref{fig_4chNN}(E)).  IEE ($0.904\pm0.024$) had the highest mean similarity with KLD, compared to GC ($0.711\pm0.076$), TE ($0.888\pm0.022$), and CCM ($0.687\pm0.033$). These results present that IEE can accurately quantify IntDC from $\vec{x}_{\text{obs}}$ alone, alleviating the need for additional interventions to the dynamics.

\subsection{Application of IEE in inferring neural connectomes of \textit{C.\,elegans}}
\begin{figure}[!htbp]
	\centering
	\includegraphics[width = 0.95\textwidth]{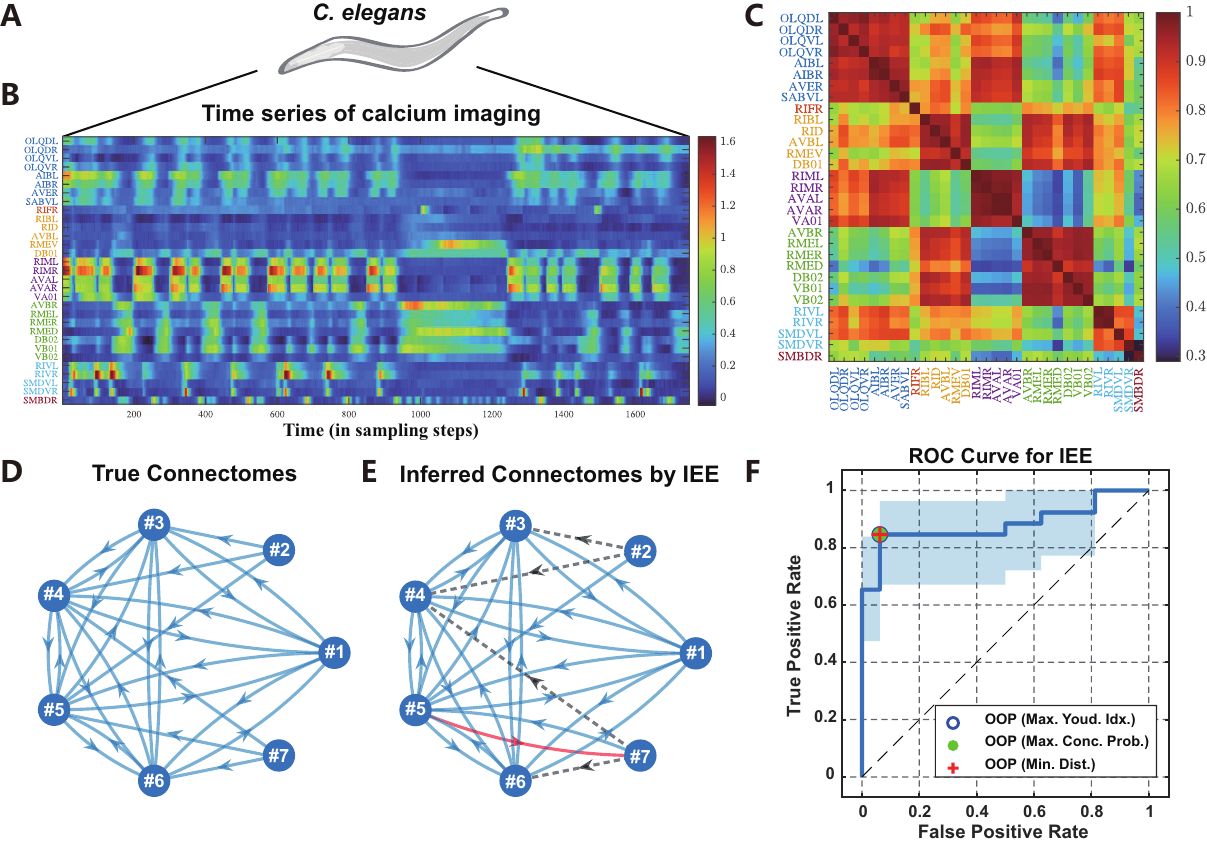}
	\caption{Application of the IEE on inferring the neural connectomes of \textit{C.\,elegans}. (A) is an illustration of \textit{C.\,elegans}. (B) displays the calcium imaging time series of $31$ key neurons, with their names listed on the left and colors representing different clusters. (C) is the cosine similarity matrix between neurons, aiding the clustering process. (D) is the true connectomes between $7$ clusters, while (E) is the inferred causal network at the the optimal operating point (OOP) determined by IEE. The red edge represents a false positive, and the black dashed edges represent false negatives. (F) shows the ROC curve (the blue line) for IEE. The same OOP is obtained under three criteria, i.e. the maximum Youden index (the blue circle),  the maximum concordance probability (the green dot), and the minimum distance to the point $(0, 1)$ (the red cross). The shaded area around the ROC curve represents the $95\%$ confidence interval obtained through bootstrapping. The AUC value of IEE is $0.882$.}\label{fig_5Celegans}
\end{figure}

We applied the IEE criterion to infer the neural connectomes of \textit{Caenorhabditis elegans} (\textit{C.\,elegans}), a model organism known for its comprehensively studied nervous system \cite{Cook2019}. Calcium fluorescence imaging time series data from $31$ neurons with specific functions of a freely moving \emph{C.\,elegans} were collected from \textit{Kato et al.} \cite{Kato2015} (see Figs.~\ref{fig_5Celegans}(A-B) and fig.~S7). The cosine similarity between different neurons were calculated and the neurons were clustered into $7$ clusters after data preprocessing (see \figref{fig_5Celegans}(C), Supplementary Text and table~S5). We presented the ground truth of neural connectomes between clusters in a directed graph (\figref{fig_5Celegans}(D)), which was detected by electron microscopy \cite{Cook2019, Banerjee2023, EmmonsLab2020}. The IEE values between different clusters were calculated to infer the IntDC network. The ROC curve of IEE got a high AUC value $0.882$ (\figref{fig_5Celegans}(F)). By maximum Youden index, maximum concordance probability, and minimum distance to the point $(0,1)$, IEE gave the same optimal operating point (OOP). At the OOP, IEE provided an inferred connectomic network with only 1 false positive and 4 false negatives (\figref{fig_5Celegans}(E)). We also compared IEE with other ConDC indices, i.e. GC/TE/CCM, which revealed that IEE exhibited the highest AUC value, best OOP, largest similarity to the ground truth, lowest false positive at the OOP, and lowest false negative at the OOP (see \tabref{tab2_Celegans}, figs.~S8-S10, and table~S6). 

Additionally, we inferred the connectomic network using the PCMCI algorithm with two approaches: the partial correlation (PCMCI-ParCorr) and CMI test (PCMCI-CMI). PCMCI used an independence test with the only parameter being the significance level $\alpha_{\text{PC}}$. The false positive edges for PCMCI-ParCorr ($\alpha_{\text{PC}}=0.05$) and PCMCI-CMI ($\alpha_{\text{PC}}=0.01$) are $5$ and $6$, respectively. Both approaches have $4$ false negative edges. A detailed description of the procedure and results can be found in the Supplementary Text, figs.~S10(e-f) and table S6.

These findings demonstrate the efficacy of IEE in reconstructing the neural connectomes of \textit{C.\,elegans}. 

\subsection{Application of IEE to COVID-19 transmission in Japan}
We validated that IEE was a promising indicator for assessing the transmission dynamics of infectious diseases. Our study collected the daily confirmed COVID-19 cases from all $47$ prefectures in Japan, spanning a duration of 1209 days from January 16, 2020 to May 8, 2023 (fig.~S11). 

First, we calculated the IEE indices for IntDC among different prefectures. IEE effectively ranked the influence across prefectures, with the top five affected areas by Tokyo being Kanagawa (with IEE value $0.601$), Chiba (with IEE value $0.568$), Saitama (with IEE value $0.534$), Aichi (with IEE value $0.453$), and Osaka (with IEE value $0.435$). Remarkably, this ranking coincides with the geographical proximity and socio-economic connections to Tokyo. Kanagawa, Chiba and Saitama are in the same metropolitan area with Tokyo, while Aichi and Osaka are far from Tokyo but connected to Tokyo by expressway and Shinkansen.

\begin{figure}[!htbp]
	\centering
	\includegraphics[width = 0.95\textwidth]{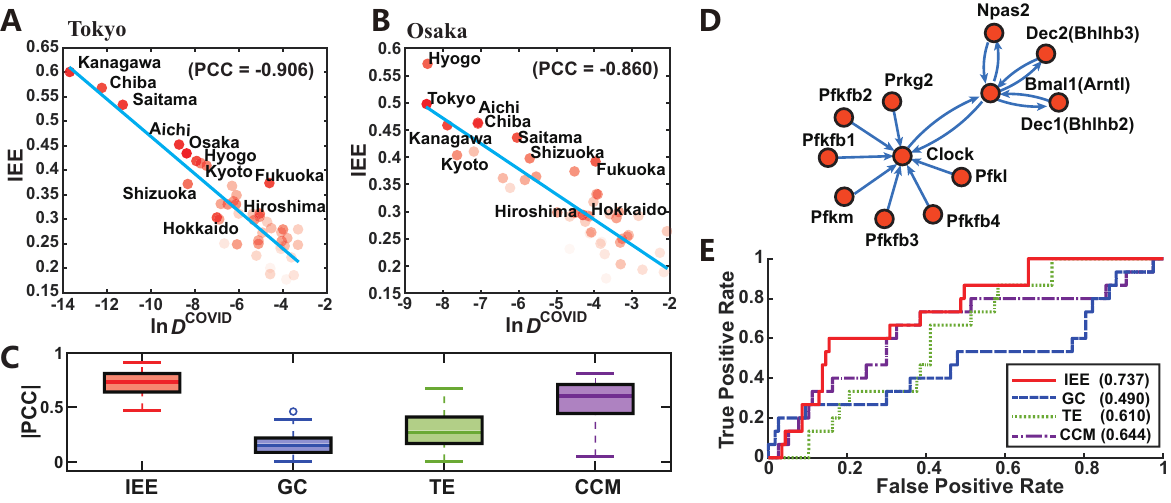}
	\caption{Application of the IEE on the COVID-19 transmission in Japan and the gene regulatory network (GRN) of circadian rhythms. (A) shows that IEE values from Tokyo to other prefectures are highly linear-correlated with $\ln D^{\text{COVID}}$, where $D^{\text{COVID}}$ is the effective distance obtained from the gravity model in consideration of various factors. (B) is the result for IEE values from Osaka to the other prefectures. (C) is the boxplot for $|\text{PCC}(c_{ij}^{\text{Inf}}, \ln D_{ij}^{\text{COVID}})|$ with average over fixed $i$, where PCC is the Pearson correlation coefficient and $c^{\text{Inf}}$ is the inferred causal strengths by IEE, GC, TE, or CCM. The PCC linearity of IEE ($0.724\pm0.116$) is significantly higher than that of GC ($0.161\pm0.099$), TE($0.284\pm0.168$), and CCM ($0.568\pm0.174$).  (D) shows the GRN surrounding \textit{Clock}, a key circadian gene. (E) demonstrates the ROC curves and AUC values of the IntDC index IEE, and three  ConDC indices GC/TE/CCM.} \label{fig_6CovidCirc}
\end{figure}

Then, IEE was validated to coincide with the effective distance matrix $D^{\text{COVID}}$ between prefectures. The $D^{\text{COVID}}\in\mathbb{R}^{47\times 47}$ was non-symmetric and designed by incorporating factors such as geometric distances, human mobility, population sizes, and infectious rates across prefectures (based on the gravity model \cite{Barbosa2018, Zipf1946}, see Supplementary Text and fig.~S11 for details). The element $D_{ij}^{\text{COVID}}$ served as a benchmark for quantifying COVID-19 transmission from  prefecture $i$ to $j$. The IEE values demonstrated a strong linear correlation with $\ln D^{\text{COVID}}$, particularly when assessing the causality from Tokyo/Osaka to other prefectures (with Pearson correlation coefficients, i.e. PCCs, $-0.906$ for Tokyo and $-0.860$ for Osaka, Figs.~\ref{fig_6CovidCirc}(A-B)). This correlation remained high for other regions (see fig.~S12), such as Aichi (with PCC value $-0.811$), Hokkaido (with PCC value $-0.751$), Fukuoka (with PCC value $-0.752$), and Okinawa (with PCC value $-0.811$).

Further, IEE outperformed ConDC indices in practicality. Notably, the absolute value of the PCC between IEE and $\ln D^{\text{COVID}}$ was $0.724$ substantially higher than those obtained by GC $0.161$, TE $0.284$, and CCM $0.568$ (average over specific prefecture, see the second to last column in \tabref{tab3_covidCorr} and the boxplot in \figref{fig_6CovidCirc}(C)). More detailed comparisons to support the effectiveness of IEE were presented in \tabref{tab3_covidCorr} (with specific illustration for Tokyo in fig.~S13).

These results underscore the reliability and efficacy of IEE in providing quantitative insights into COVID-19 transmission dynamics in Japan. Importantly, IEE performs as a simple and valuable tool for causal analyses solely from daily confirmed time-series data, without the need for complex models considering various factors such as geodesic distances, human mobility, population sizes, and infectious rates.

\subsection{Application of IEE to investigating circadian rhythms}
We investigated the gene regulatory networks (GRNs) related to key genes on the circadian rhythm. The time series of gene expressions were measured by microarray from cultured rat cells \cite{wang2009network, kawaguchi2007establishment}. Through decades of molecular and genetic studies, many key circadian genes, such as \textit{Clock, Bmal1(Arntl), Dec1(Bhlhb2), Dec2(Bhlhb3), Cry1, Cry2, Per1, Per2, Per3}, have been identified on mammals \cite{ko2006molecular, ueda2005system}. Figure \ref{fig_6CovidCirc}(D) displayed the GRN around \textit{Clock} at the protein level, where the transcription factor \textit{Clock} is phosphorylated by PFK family genes. We calculated the IEE/GC/TE/CCM to reconstruct the GRN (ROC curves and AUC values in \figref{fig_6CovidCirc}(E)). IEE designed for quantifying IntDC presented a higher AUC value ($0.737$) than ConDC indices ($0.490$ for GC, $0.610$ for TE, and $0.644$ for CCM). We also tested the indices on the GRN surrounding Cry1/Cry2 (fig.~S14), where IEE obtained the highest AUC value ($0.639$) and outperformed the others ($0.428$ for GC, $0.619$ for TE, and $0.614$ for CCM).

\section{Conclusions and discussions}
In summary, we have established a framework of IntDC based on dynamical systems theory and introduced the IEE criterion to quantify IntDC in this study. IEE measures the information flow between interventional causes and effects in the delay-embedding space, making it suitable for analyzing non-linear and non-separable systems\cite{sugihara2012detecting, shi2022embedding, liang2016information}. Moreover, IEE is able to infer IntDC solely from observational time-series data, without requiring additional perturbations to the system. This property makes IEE particularly suitable for non-intervention systems. Both theoretical derivations and numerical examples presented in this study provide strong evidence to support the effectiveness, accuracy and robustness of IEE in detecting IntDC and reconstructing networks. Furthermore, IEE serves as an effective tool for evaluating and ranking the causal dependence between variables within a dynamical system. Through real-world examples, we illustrated the promising applicability of IEE in diverse fields such as regulatory inference and disease transmission studies.

Information-flow-based methods have been developing rapidly in recent years \cite{liang2021normalized, hristopulos2024information, pires2024general}. Liang's information flow theory, albeit with full nonlinearity in its theoretical formalism, is yet to be implemented for practical applications on real-world data. Existing implementations still rely on assumptions such as independent white noise and linearity when evaluating information flow from data. Our IEE algorithm, inspired by the conception information flow, is specifically designed to handle fully nonlinear systems and has demonstrated applicability across a variety of real datasets. The proposed IntDC framework, together with the IEE algorithm, advances the frontier of causality research by providing a practical and effective tool for nonlinear causal inference.

Runge et al. \cite{runge2019inferring} summarized recent advances and key challenges in the field of causal inference and developed the platform \href{https://causeme.net}{https://causeme.net}, which serves as a valuable resource for researchers. Causal inference holds great promise for applications in earth system science and beyond. Our proposed IEE algorithm, together with the concept of IntDC, provides a novel and practical approach for investigating causality in complex nonlinear systems.

There are several important open issues that warrant further investigation and future development.

(1) The theory of IntDC and the IEE criterion can be extended to analyze causal relations between groups of variables. Specifically, in \eqnref{eq_xy}, the variables $x$ and $y$ can be generalized to multivariate vectors, enabling the investigation of complex interactions among multiple variables. 

(2) The data used for computing IEE can be non-uniformly sampled in time. In this study, we employed the standard time-delayed embedding \eqnref{eq_delayX} and \eqnref{eq_delayY}. However, according to the generalized Takens' embedding theorem, any set of $L+1$ observations of $x_t$ and $y_t$ can construct topological diffeomorphisms to  the manifolds $\mathcal{M}_X$ and $\mathcal{M}_Y$, respectively. As shown in \eqnref{eq_cs1}, only the local neighborhood structure in the embedding space is essential for the computation. Therefore, the IEE algorithm can be naturally extended to infer causality from datasets sampled at varying or irregular time intervals.

(3) To further distinguish direct and indirect causality, we can generalize IEE to its conditional version
	\begin{equation}
	\text{cIEE}[x\rightarrow y|z] := \text{CMI}(\delta\vec{X}_t, \delta\vec{Y}_{t+1}|\vec{Y}_{t+1},\vec{Z}_t, \delta\vec{Z}_t), \label{eq_cIEE}
	\end{equation}

\noindent where ``cIEE'' is short for the conditional IEE, $z$ is a third variable whose time-delayed vector is $\vec{Z}_t$, and $\delta\vec{Z}_t$ represents the intervened deviation on $\vec{Z}_t$. Equation \eqref{eq_cIEE} measures the direct IntDC from $x$ to $y$ conditioned on $z$. Further, in combination with the well-known PC algorithm \cite{spirtes1991algorithm} or PCMCI\cite{runge2019detecting}, cIEE offers a feasible way to remove high-order indirect causal edges iteratively and reconstruct the direct causal network.

(4) In \cite{sugihara2012detecting, ye2015distinguishing, krakovska2018comparison}, researchers have pointed out that time-delayed embedding algorithms are particularly suitable for deterministic dynamical systems with an attractive manifold. However, the coupling strength between variables could influence the accuracy of these methods. Stark et al. \cite{stark1997takens} extended Takens' embedding to stochastic systems, providing a theoretical basis for applying embedding approaches beyond purely deterministic dynamics. In the Results section of this study, we demonstrated the effectiveness of IEE for Logistic systems with Gaussian noises. Nevertheless, how systematic factors, such as coupling strength and noise characteristics,  affect the quantification of causal strength in more general settings remains an open question.
	
(5) Moreover, causal inference empowered by deep learning is emerging as a frontier in research. Notable advances include CausalEGM \cite{liu2022causalegm}, Causalformer\cite{lu2023attention}, intervened reservoir computing\cite{zhao2024detecting}, and reservoir cross mapping\cite{cao2025reservoir}. Integrating the IntDC framework with neural networks presents a promising direction for further exploration. On one hand, causality theory provides interpretability to artificial intelligence; on the other hand, large AI foundation models offer unprecedented capacity to uncover causal relationships in complex data . These developments signal the advent of a transformative ``big causality era'', in which data-driven discovery and causal reasoning evolve in synergy.

%BibTeX users: After compilation, comment out the following two lines and paste in
% the generated .bbl file. 

\bibliography{scibib}

\begin{thebibliography}{10}

\bibitem{neyman1923application}
J.~S. Neyman, {\it Statistical Science\/} {\bf 5}, 465 (1990).

\bibitem{rubin1978bayesian}
D.~B. Rubin, {\it The Annals of Statistics\/} {\bf 6}, 34 (1978).

\bibitem{sekhon2008neyman}
J.~S. Sekhon, {\it The Oxford Handbook of Political Methodology\/} {\bf 2}, 1
  (2008).

\bibitem{wright1921correlation}
S.~Wright, {\it Journal of Agricultural Research\/} {\bf 20}, 557 (1921).

\bibitem{wright1934method}
S.~Wright, {\it The Annals of Mathematical Statistics\/} {\bf 5}, 161 (1934).

\bibitem{pearl1995causal}
J.~Pearl, {\it Biometrika\/} {\bf 82}, 669 (1995).

\bibitem{pearl2009causal}
J.~Pearl, {\it Statistics Surveys\/} {\bf 3}, 96 (2009).

\bibitem{pearl2000models}
J.~Pearl, {\it Causality: models, reasoning, and inference\/} (Cambridge
  University Press, UK, 2000).

\bibitem{angrist1996identification}
J.~D. Angrist, G.~W. Imbens, D.~B. Rubin, {\it Journal of the American
  Statistical Association\/} {\bf 91}, 444 (1996).

\bibitem{shi2022embedding}
J.~Shi, L.~Chen, K.~Aihara, {\it Journal of the Royal Society Interface\/} {\bf
  19}, 20210766 (2022).

\bibitem{granger1969investigating}
C.~W.~J. Granger, {\it Econometrica\/} {\bf 37}, 424 (1969).

\bibitem{lusch2016inferring}
B.~Lusch, P.~D. Maia, J.~N. Kutz, {\it Physical Review E\/} {\bf 94}, 032220
  (2016).

\bibitem{schreiber2000measuring}
T.~Schreiber, {\it Physical Review Letters\/} {\bf 85}, 461 (2000).

\bibitem{sugihara2012detecting}
G.~Sugihara, {\it et~al.\/}, {\it Science\/} {\bf 338}, 496 (2012).

\bibitem{leng2020partial}
S.~Leng, {\it et~al.\/}, {\it Nature Communications\/} {\bf 11}, 1 (2020).

\bibitem{ying2022continuity}
X.~Ying, {\it et~al.\/}, {\it Research\/} {\bf 2022}, 9870149 (2022).

\bibitem{liang2008information}
X.~S. Liang, {\it Physical Review E—Statistical, Nonlinear, and Soft Matter
  Physics\/} {\bf 78}, 031113 (2008).

\bibitem{liang2016information}
X.~S. Liang, {\it Physical Review E\/} {\bf 94}, 052201 (2016).

\bibitem{ye2015distinguishing}
H.~Ye, E.~R. Deyle, L.~J. Gilarranz, G.~Sugihara, {\it Scientific Reports\/}
  {\bf 5}, 14750 (2015).

\bibitem{pecora1995statistics}
L.~M. Pecora, T.~L. Carroll, J.~F. Heagy, {\it Physical Review E\/} {\bf 52},
  3420 (1995).

\bibitem{harnack2017topological}
D.~Harnack, E.~Laminski, M.~Sch{\"u}nemann, K.~R. Pawelzik, {\it Physical
  Review Letters\/} {\bf 119}, 098301 (2017).

\bibitem{amigo2018detecting}
J.~M. Amig{\'o}, Y.~Hirata, {\it Chaos: An Interdisciplinary Journal of
  Nonlinear Science\/} {\bf 28}, 075302 (2018).

\bibitem{stavroglou2020unveiling}
S.~K. Stavroglou, A.~A. Pantelous, H.~E. Stanley, K.~M. Zuev, {\it Proceedings
  of the National Academy of Sciences U.S.A.\/} {\bf 117}, 7599 (2020).

\bibitem{tao2023detecting}
P.~Tao, {\it et~al.\/}, {\it Fundamental Research\/}  (2023).

\bibitem{runge2019detecting}
J.~Runge, P.~Nowack, M.~Kretschmer, S.~Flaxman, D.~Sejdinovic, {\it Science
  advances\/} {\bf 5}, eaau4996 (2019).

\bibitem{friston2003dynamic}
K.~J. Friston, L.~Harrison, W.~Penny, {\it Neuroimage\/} {\bf 19}, 1273 (2003).

\bibitem{stephan2007dynamic}
K.~E. Stephan, {\it et~al.\/}, {\it Journal of biosciences\/} {\bf 32}, 129
  (2007).

\bibitem{krakovska2018comparison}
A.~Krakovsk{\'a}, {\it et~al.\/}, {\it Physical Review E\/} {\bf 97}, 042207
  (2018).

\bibitem{stepaniants2020inferring}
G.~Stepaniants, B.~W. Brunton, J.~N. Kutz, {\it Physical Review E\/} {\bf 102},
  042309 (2020).

\bibitem{smirnov2014quantification}
D.~A. Smirnov, {\it Physical Review E\/} {\bf 90}, 062921 (2014).

\bibitem{smirnov2020transfer}
D.~A. Smirnov, {\it Physical Review E\/} {\bf 102}, 062139 (2020).

\bibitem{liang2005information}
X.~S. Liang, R.~Kleeman, {\it Physical review letters\/} {\bf 95}, 244101
  (2005).

\bibitem{liang2014unraveling}
X.~S. Liang, {\it Physical Review E\/} {\bf 90}, 052150 (2014).

\bibitem{liang2021normalized}
X.~S. Liang, {\it Entropy\/} {\bf 23}, 679 (2021).

\bibitem{hristopulos2024information}
D.~T. Hristopulos, {\it IEEE Transactions on Signal Processing\/} {\bf 72}, 839
  (2024).

\bibitem{yi2022quantum}
B.~Yi, S.~Bose, {\it Physical Review Letters\/} {\bf 129}, 020501 (2022).

\bibitem{pires2024general}
C.~A. Pires, D.~Docquier, S.~Vannitsem, {\it Physica D: Nonlinear Phenomena\/}
  {\bf 458}, 133988 (2024).

\bibitem{takens1981detecting}
F.~Takens, {\it Dynamical systems and turbulence, Warwick 1980\/} (Springer,
  1981), pp. 366--381.

\bibitem{sauer1991embedology}
T.~Sauer, J.~A. Yorke, M.~Casdagli, {\it Journal of Statistical Physics\/} {\bf
  65}, 579 (1991).

\bibitem{stark1997takens}
J.~Stark, D.~Broomhead, M.~Davies, J.~Huke, {\it Nonlinear Analysis: Theory,
  Methods and Applications\/} {\bf 30}, 5303 (1997).

\bibitem{cummins2015efficacy}
B.~Cummins, T.~Gedeon, K.~Spendlove, {\it SIAM Journal on Applied Dynamical
  Systems\/} {\bf 14}, 335 (2015).

\bibitem{altman2015points}
N.~Altman, M.~Krzywinski, {\it Nature methods\/} {\bf 12} (2015).

\bibitem{shojaie2022granger}
A.~Shojaie, E.~B. Fox, {\it Annual Review of Statistics and Its Application\/}
  {\bf 9}, 289 (2022).

\bibitem{korb2004varieties}
K.~B. Korb, L.~R. Hope, A.~E. Nicholson, K.~Axnick, {\it PRICAI 2004: Trends in
  Artificial Intelligence\/} (Springer, 2004), pp. 322--331.

\bibitem{amblard2012relation}
P.-O. Amblard, O.~J. Michel, {\it Entropy\/} {\bf 15}, 113 (2012).

\bibitem{cover1999elements}
T.~M. Cover, {\it Elements of information theory\/} (John Wiley \& Sons, 1999).

\bibitem{runge2019inferring}
J.~Runge, {\it et~al.\/}, {\it Nature communications\/} {\bf 10}, 2553 (2019).

\bibitem{aihara1990chaotic}
K.~Aihara, T.~Takabe, M.~Toyoda, {\it Physics Letters A\/} {\bf 144}, 333
  (1990).

\bibitem{adachi1997associative}
M.~Adachi, K.~Aihara, {\it Neural Networks\/} {\bf 10}, 83 (1997).

\bibitem{Cook2019}
S.~J. Cook, {\it et~al.\/}, {\it Nature\/} {\bf 571}, 63 (2019).

\bibitem{Kato2015}
S.~Kato, {\it et~al.\/}, {\it Cell\/} {\bf 163}, 656 (2015).

\bibitem{Banerjee2023}
A.~Banerjee, S.~Chandra, E.~Ott, {\it Proceedings of the National Academy of
  Sciences U.S.A.\/} {\bf 120}, e2216030120 (2023).

\bibitem{EmmonsLab2020}
{Emmons Lab}, {Hermaphrodite and Male Connectomes (Adjacency Matrices), Adults
  (corrected July 2020)}, available at
  \url{https://wormwiring.org/pages/adjacency.html} (2020). Accessed:
  2023-12-15.

\bibitem{Barbosa2018}
H.~Barbosa, {\it et~al.\/}, {\it Physics Reports\/} {\bf 734}, 1 (2018).

\bibitem{Zipf1946}
G.~K. Zipf, {\it American Sociological Review\/} {\bf 11}, 677 (1946).

\bibitem{wang2009network}
Y.~Wang, X.-S. Zhang, L.~Chen, {\it OMICS\/} {\bf 13}, 313 (2009).

\bibitem{kawaguchi2007establishment}
S.~Kawaguchi, {\it et~al.\/}, {\it Biochemical and Biophysical Research
  Communications\/} {\bf 355}, 555 (2007).

\bibitem{ko2006molecular}
C.~H. Ko, J.~S. Takahashi, {\it Human Molecular Genetics\/} {\bf 15}, R271
  (2006).

\bibitem{ueda2005system}
H.~R. Ueda, {\it et~al.\/}, {\it Nature Genetics\/} {\bf 37}, 187 (2005).

\bibitem{spirtes1991algorithm}
P.~Spirtes, C.~Glymour, {\it Social Science Computer Review\/} {\bf 9}, 62
  (1991).

\bibitem{liu2022causalegm}
Q.~Liu, Z.~Chen, W.~H. Wong, {\it arXiv preprint arXiv:2212.05925\/}  (2022).

\bibitem{lu2023attention}
Z.~Lu, {\it et~al.\/}, {\it arXiv preprint arXiv:2311.06928\/}  (2023).

\bibitem{zhao2024detecting}
J.~Zhao, {\it et~al.\/}, {\it Communications Physics\/} {\bf 7}, 232 (2024).

\bibitem{cao2025reservoir}
R.~Cao, {\it et~al.\/}, {\it Cell Reports Physical Science\/} {\bf 6}, 102683
  (2025).

\end{thebibliography}


\begin{thebibliography}{10}

\bibitem{Kato2015}
S.~Kato, {\it et~al.\/}, {\it Cell\/} {\bf 163}, 656 (2015).

\bibitem{Cook2019}
S.~J. Cook, {\it et~al.\/}, {\it Nature\/} {\bf 571}, 63 (2019).

\bibitem{Banerjee2023}
A.~Banerjee, S.~Chandra, E.~Ott, {\it Proceedings of the National Academy of
  Sciences U.S.A.\/} {\bf 120}, e2216030120 (2023).

\bibitem{EmmonsLab2020}
{Emmons Lab}, {Hermaphrodite and Male Connectomes (Adjacency Matrices), Adults
  (corrected July 2020)}, available at
  \url{https://wormwiring.org/pages/adjacency.html} (2020). Accessed:
  2023-12-15.

\bibitem{MHLW2023}
{Ministry of Health, Labour and Welfare}, {Trend in the number of newly
  confirmed cases (daily)}, in Visualizing the data: information on COVID-19
  infections, available at \url{https://covid19.mhlw.go.jp/en/} (2023).
  Accessed: 2023-10-05.

\bibitem{GSI}
{Geospatial Information Authority of Japan}, {Distances between prefectural
  offices}, in Geographic information of Japan (in Japanese), available at
  \url{https://www.gsi.go.jp/KOKUJYOHO/kenchokan.html} (2023). Accessed:
  2023-10-12.

\bibitem{MLIT2019}
{Ministry of Land, Infrastructure, Transport and Tourism}, {2015 Inter-Regional
  Travel Survey in Japan}, in National Trunk Line Passenger Net Flow Survey (in
  Japanese), available at\\
  \url{https://www.mlit.go.jp/sogoseisaku/soukou/sogoseisaku\_soukou\_fr\_000016.html}
  (2019). Accessed: 2023-10-11.

\bibitem{MLIT2017}
{Ministry of Land, Infrastructure, Transport and Tourism}, {2015 Passenger
  Regional Flow Survey Dataset}, in Portal Site of Official Statistics of Japan
  (eStat) (in Japanese), available at\\
  \url{https://www.e-stat.go.jp/stat-search/files?tclass=000001056011&cycle=8&year=20151}
  (2017). Accessed: 2023-10-04.

\bibitem{SBJ2022}
{Statistics Bureau of Japan}, {Population by Sex and Sex Ratio for Prefectures
  - Total population, Japanese population, October 1, 2021}, in Portal Site of
  Official Statistics of Japan (eStat) (in Japanese), available at\\
  \url{https://www.e-stat.go.jp/stat-search/files?page=1&toukei=00200524&tstat=000000090001}
  (2022). Accessed: 2024-02-15.

\bibitem{leng2020partial}
S.~Leng, {\it et~al.\/}, {\it Nature Communications\/} {\bf 11}, 1 (2020).

\bibitem{wang2009network}
Y.~Wang, X.-S. Zhang, L.~Chen, {\it OMICS\/} {\bf 13}, 313 (2009).

\bibitem{takens1981detecting}
F.~Takens, {\it Dynamical systems and turbulence, Warwick 1980\/} (Springer,
  1981), pp. 366--381.

\bibitem{sauer1991embedology}
T.~Sauer, J.~A. Yorke, M.~Casdagli, {\it Journal of Statistical Physics\/} {\bf
  65}, 579 (1991).

\bibitem{stark1997takens}
J.~Stark, D.~Broomhead, M.~Davies, J.~Huke, {\it Nonlinear Analysis: Theory,
  Methods and Applications\/} {\bf 30}, 5303 (1997).

\bibitem{cummins2015efficacy}
B.~Cummins, T.~Gedeon, K.~Spendlove, {\it SIAM Journal on Applied Dynamical
  Systems\/} {\bf 14}, 335 (2015).

\bibitem{kraskov2004estimating}
A.~Kraskov, H.~St{\"o}gbauer, P.~Grassberger, {\it Physical Review E\/} {\bf
  69}, 066138 (2004).

\bibitem{kozachenko1987sample}
L.~Kozachenko, N.~N. Leonenko, {\it Problemy Peredachi Informatsii\/} {\bf 23},
  9 (1987).

\bibitem{lindner2011trentool}
M.~Lindner, R.~Vicente, V.~Priesemann, M.~Wibral, {\it BMC Neuroscience\/} {\bf
  12}, 1 (2011).

\bibitem{aihara1990chaotic}
K.~Aihara, T.~Takabe, M.~Toyoda, {\it Physics Letters A\/} {\bf 144}, 333
  (1990).

\bibitem{adachi1997associative}
M.~Adachi, K.~Aihara, {\it Neural Networks\/} {\bf 10}, 83 (1997).

\bibitem{Granger1969}
C.~W.~J. Granger, {\it Econometrica\/} {\bf 37}, 424 (1969).

\bibitem{Schreiber2000}
T.~Schreiber, {\it Phys. Rev. Lett.\/} {\bf 85}, 461 (2000).

\bibitem{Sugihara2012}
G.~Sugihara, {\it et~al.\/}, {\it Science\/} {\bf 338}, 496 (2012).

\bibitem{Rota2015}
M.~Rota, L.~Antolini, M.~G. Valsecchi, {\it BMC Medical Research Methodology\/}
  {\bf 15}, 24 (2015).

\bibitem{runge2019detecting}
J.~Runge, P.~Nowack, M.~Kretschmer, S.~Flaxman, D.~Sejdinovic, {\it Science
  advances\/} {\bf 5}, eaau4996 (2019).

\bibitem{Shi2022}
J.~Shi, L.~Chen, K.~Aihara, {\it Journal of the Royal Society Interface\/} {\bf
  19}, 20210766 (2022).

\bibitem{Tao2023}
P.~Tao, {\it et~al.\/}, {\it Fundamental Research\/}  (2023). In press,
  available online 6 February 2023.

\bibitem{Barbosa2018}
H.~Barbosa, {\it et~al.\/}, {\it Physics Reports\/} {\bf 734}, 1 (2018).

\bibitem{Zipf1946}
G.~K. Zipf, {\it American Sociological Review\/} {\bf 11}, 677 (1946).

\bibitem{kawaguchi2007establishment}
S.~Kawaguchi, {\it et~al.\/}, {\it Biochemical and Biophysical Research
  Communications\/} {\bf 355}, 555 (2007).

\bibitem{morioka2008phase}
R.~Morioka, {\it et~al.\/}, {\it The Second International Symposium on
  Optimization and Systems Biology\/} (2008).

\bibitem{ma2014detecting}
H.~Ma, K.~Aihara, L.~Chen, {\it Scientific Reports\/} {\bf 4}, 7464 (2014).

\bibitem{ko2006molecular}
C.~H. Ko, J.~S. Takahashi, {\it Human Molecular Genetics\/} {\bf 15}, R271
  (2006).

\bibitem{ueda2005system}
H.~R. Ueda, {\it et~al.\/}, {\it Nature Genetics\/} {\bf 37}, 187 (2005).

\end{thebibliography}

\bibliographystyle{Science}

\section*{Acknowledgments}

\subsection*{Funding:}
 {This work was supported by the National Natural Science Foundation of China [grant numbers 42450192, 42450135, 12301620, 11925103 and 32100509], the National Key R\&D Program of China [grant number 2022YFC2704604], the Shanghai Municipal Data Bureau special fund for urban digital transformation [grant number 202401065], the AI for Science Foundation of Fudan University [grant number FudanX24AI041], the AI for Science Program of Shanghai Municipal Education Commission [grant number 24RGZN10], the Japan Science and Technology Agency Moonshot R\&D [grant number JPMJMS2021], the Institute of AI and Beyond of UTokyo, the International Research Center for Neurointelligence (WPI-IRCN) at The University of Tokyo Institutes for Advanced Study (UTIAS), the Cross-ministerial Strategic Innovation Promotion Program (SIP), the 3rd period of SIP [grant number JPJ012207].
 }

\subsection*{Competing interests: }
The authors declare no competing interest.
\subsection*{Data and materials availability: }
The code for IEE is available at \href{https://github.com/smsxiaomayi/IEE}{https://github.com/smsxiaomayi/IEE}.

%Here you should list the contents of your Supplementary Materials -- below is an example. 
%You should include a list of Supplementary figures, Tables, and any references that appear only in the SM. 
%Note that the reference numbering continues from the main text to the SM.
% In the example below, Refs. 4-10 were cited only in the SM.     
\section*{Supplementary materials}
{Materials and Methods

Supplementary Text

Figs. S1 to S14

Tables S1 to S6

References
}
% For your review copy (i.e., the file you initially send in for
% evaluation), you can use the {figure} environment and the
% \includegraphics command to stream your figures into the text, placing
% all figures at the end.  For the final, revised manuscript for
% acceptance and production, however, PostScript or other graphics
% should not be streamed into your compliled file.  Instead, set
% captions as simple paragraphs (with a \noindent tag), setting them
% off from the rest of the text with a \clearpage as shown  below, and
% submit figures as separate files according to the Art Department's
% instructions.

\clearpage
\begin{table}[!htbp]
	\centering
	\caption{Algorithm: Interventional Embedding Entropy (IEE).}\label{tab1_IEEalg}
	\begin{tabular}{lll}
		\Xhline{1.5pt}
		1. Given observed time series $\{x_{t_n}|t_n\geq 0\}, \{y_{t_n}| t_n\geq 0\}$.\\
		2. Construct time-delayed vectors $\vec{X}_{t_n}$ and $\vec{Y}_{t_n+1}$ by Eqs.~\eqref{eq_delayX}-\eqref{eq_delayY}.\\
		%	3. Get $K$ nearest neighbors $\vec{Y}_{t_k^{(n)}+1}$ around $\vec{Y}_{t_n+1}$, $k=1,2, \dots, K$.\\
		3. Get $K$ nearest neighbors $\vec{Y}_{t_k+1}$ around $\vec{Y}_{t_n+1}$, $k=1,2, \dots, K$.\\
		4. Sample $\delta\vec{Y}_{t_n+1}$ condition on $\vec{Y}_{t_n+1}$ as $\vec{Y}_{t_k+1}-\vec{Y}_{t_n}$.\\
		5. Sample $\delta\vec{X}_{t_n}$ condition on $\vec{Y}_{t_n+1}$ as $\vec{X}_{t_k}-\vec{X}_{t_n}$ with the same time labels $t_k$.\\
		%	6. Compute MI between $\delta\vec{X}_{t_n}|_{\vec{Y}_{t_n+1}}$ and $ \delta\vec{Y}_{t_n+1}|_{\vec{Y}_{t_n+1}}$ at $t_n$.\\
		6. Calculate \eqnref{eq_cs1} around $\vec{Y}_{t_n+1}$ and average over $n$ samples to  approximate $\text{IEE}[x\rightarrow y]$.\\
		\Xhline{1.5pt}
	\end{tabular}
	%	\addtabletext{nomenclature for the TSs refers to the numbered species in the table.}
\end{table}

\begin{table}[!htbp]
	\centering
	\caption{\label{tab2_Celegans}
		Comparison of causal indices on the inference of \emph{C. elegans} neural connectomes.}
	\begin{threeparttable}
		{
			\begin{tabular}{cllllcll}
				\toprule
%				\hline
				~& \multicolumn{4}{c}{Properties of ROC curves}
				&& \multicolumn{2}{c}{Cosine Similarity} \\
				\cline{2-5}\cline{7-8}
				& \footnotesize{AUC~~} 
				& \footnotesize{\makecell{Maximum \\Youden\\ index~~}}
				& \footnotesize{\makecell{Maximum \\concordance \\probability~~}}
				& \footnotesize{\makecell{Minimum \\distance to \\$(0, 1)$~~}}
				&& \footnotesize{to $\vec{C}$~~~~} & \footnotesize{to $\log(1+\vec{C})$~~}\\
				\midrule
				$\vec{C}^{\text{IEE}}$
				~&$\textbf{0.882}$   &$\textbf{0.784}$   &$\textbf{0.793}$   &$\textbf{0.166}$&&$\textbf{0.750}$ &~~$\textbf{0.905}$\\
				$\vec{C}^{\text{GC}}$
				~&$0.858$        &$0.721$        &$0.740$        &$0.198$        && $0.430$        &~~$0.678$      \\
				$\vec{C}^{\text{TE}}$
				~&$0.796$        &$0.620$        &$0.656$        &$0.269$        && $0.477$        &~~$0.726$      \\
				$\vec{C}^{\text{CCM}}$
				~&$0.880$        &$0.615$        &$0.639$        &$0.294$        && $0.644$        &~~$0.866$      \\
%				\hline
				\bottomrule
			\end{tabular}
		}
		\begin{tablenotes}[flushleft]\footnotesize
			\item $^\ast$ The matrices $\vec{C}^{\text{IEE}}/\vec{C}^{\text{GC}}/\vec{C}^{\text{TE}}/\vec{C}^{\text{CCM}}$ represent the inferred connectomes by the four indices. The matrix $\vec{C}$ is the ground truth of neural connectomes. The best value in each column is shown in bold.
		\end{tablenotes}
	\end{threeparttable}
\end{table}

\begin{table}[!htbp]
	\centering
	\caption{\label{tab3_covidCorr}
		Comparison of IEE/GC/TE/CCM in measuring causal strengths in the transmission of COVID-19 data in Japan.}
	\begin{threeparttable}
		{
			\begin{tabular}{ cl lllllll cc }
				\toprule
%				\hline\hline
				&& \multicolumn{7}{c}{ $|\text{PCC}\left(c^{\text{Inf}}_{ij}, \ln D^{\text{COVID}}_{ij} \right)|$
					for specific prefecture (fixed $i$)}
				&& \multicolumn{1}{c}{\small{
						\multirow{2}{*}{$|\text{PCC}\left(c^{\text{Inf}}_{ij}, \ln D^{\text{COVID}}_{ij} \right)|$}}}\\
				\cline{3-9}&
				& \multicolumn{1}{c}{\footnotesize{~Tokyo~}} 
				& \multicolumn{1}{c}{\footnotesize{~Osaka~}} 
				& \multicolumn{1}{c}{\footnotesize{~Aichi~}} 
				& \multicolumn{1}{c}{\footnotesize{~Hokkaido~}} 
				& \multicolumn{1}{c}{\footnotesize{~Fukuoka~}} 
				& \multicolumn{1}{c}{\footnotesize{~Okinawa~}}
				& \multicolumn{1}{c}{\footnotesize{~Average~}}
				&& \\[-3pt]
				&&\multicolumn{1}{c}{\scriptsize{~($i=13$)~}}
				& \multicolumn{1}{c}{\scriptsize{~($i=27$)~}}
				& \multicolumn{1}{c}{\scriptsize{~($i=23$)~}}
				& \multicolumn{1}{c}{\scriptsize{~($i=1$)~}}
				& \multicolumn{1}{c}{\scriptsize{~($i=40$)~}}
				& \multicolumn{1}{c}{~\scriptsize{~$(i=47)$~}}
				& \multicolumn{1}{c}{~\scriptsize{~over all $i$~}}
				&&\multicolumn{1}{c}{\scriptsize{for all $i,j$ with $i\neq j$}}\\
				\hline
				IEE&\hspace{5mm}&$\textbf{0.906}$    &$\textbf{0.860}$&$\textbf{0.811}$ &$\textbf{0.751}$ &$\textbf{0.752}$ & $\textbf{0.811}$ & $\textbf{0.724}$ && $\textbf{0.748}$\\
				GC&  &$0.462$         &$0.387$     &$0.236$      &$0.070$   &$0.125$   
				&$0.164$         &$0.161$ &&$0.166$     \\
				TE&  &$0.655$         &$0.308$     &$0.018$      &$0.445$   &$0.526$   
				&$0.538$         &$0.284$ &&$0.109$     \\
				CCM&  &$0.799$         &$0.795$     &$0.448$      &$0.052$   &$0.727$   
				&$0.477$         & $0.568$&&$0.585$     \\
%				\hline\hline
				\bottomrule
			\end{tabular}
		}
		\begin{tablenotes}[flushleft]\footnotesize
			\item $^\ast$ $i$ and $j$ (ranging from 1 to 47) stand for different prefectures in Japan. $c_{ij}^{\text{Inf}}$ is the causal strength from the prefecture $i$ to $j$ inferred by IEE, GC, TE, or CCM. $D_{ij}^{\text{COVID}}$ is the effective distance between prefectures $i$ and $j$. PCC is the Pearson correlation coefficient. The best value in each column is shown in bold.
		\end{tablenotes}
	\end{threeparttable}
\end{table}

\end{document}

%The next command sets up an environment for the abstract to your paper.
%\begin{figure}%[tbhp]
%\centering
%\includegraphics[width=.8\linewidth]{science.jpg}
%\end{figure}
{\centering{
		\title{Supplementary Materials for\\\vspace{1em}\textbf{Deciphering Interventional Dynamical Causality from Non-intervention Complex Systems}}}}
%\title{Quantifying interventional dynamical causality in complex networks}
\newcommand{\authorwrap}[1]{%
	\begin{minipage}{\textwidth}
		\centering
		#1
	\end{minipage}
}

\author
{\authorwrap{Jifan Shi, Yang Li, Juan Zhao, Siyang Leng, Rui Bao, Kazuyuki Aihara, Luonan Chen, Wei Lin}\\
	\normalsize{\authorwrap{\vspace{0.5em}Corresponding authors: jfshi@fudan.edu.cn (J.S.), kaihara@g.ecc.u-tokyo.ac.jp (K.A.), lnchen@sibs.ac.cn (L.C.), wlin@fudan.edu.cn (W.L.).}}
}

\date{}

\renewcommand{\thefigure}{S\arabic{figure}}  
\renewcommand{\theequation}{S\arabic{equation}} 
\renewcommand{\thetable}{S\arabic{table}} 
\newcommand{\f}{\frac}
\newcommand{\eps}{\varepsilon}
\newcommand{\EX}{\mathbb{E}}
\newcommand{\eqnref}[1]{Eq.~\eqref{#1}}
\newcommand{\figref}[1]{fig.~\ref{#1}}
\newcommand{\tabref}[1]{table~\ref{#1}}
\renewcommand{\d}{\mathop{}\!\mathrm{d}}
\renewcommand{\vec}[1]{\boldsymbol{#1}}
\let\oldalign\align
\let\oldendalign\endalign
\renewenvironment{align}{\linenomathNonumbers\oldalign}{\oldendalign\endlinenomath}

%% Comment out or remove this line before generating final copy for submission; this will also remove the warning re: "Consecutive odd pages found".
%\instructionspage  
\maketitle

%% Adds the main heading for the SI text. Comment out this line if you do not have any supporting information text.
%\SItext
\section*{The PDF file includes:}
%	\item Materials and Methods
{	Materials and Methods
	
	Supplementary Text
	
	Supplementary figs. S1 to S14
	
	Supplementary tables S1 to S6
	
	References
}	 

\newpage
%\linenumbers
\section{Materials and Methods}
The calcium fluorescence imaging data of  \textit{C.\,elegans} was collected from \cite{Kato2015}, and the neural connectomes  were available in \cite{Cook2019, Banerjee2023, EmmonsLab2020}. The daily confirmed COVID-19 cases of $47$ prefectures in Japan (1209 days) were obtained from the website of the Ministry of Health, Labour and Welfare of Japan \cite{MHLW2023}. To construct the effective distance matrix, the geodesic distances between prefectures were collected from \cite{GSI}. The net and gross human mobility data were from \cite{MLIT2019} and \cite{MLIT2017}, respectively. The population counts of each prefecture were from the statistics bureau of Japan \cite{SBJ2022}. The circadian rhythm dataset was collected from \cite{leng2020partial, wang2009network}.% and downloaded from \href{https://github.com/Partial-Cross-Mapping/circadian}{https://github.com/Partial-Cross-Mapping/circadian}.

\section{Supplementary Text}

\subsection{Proof of Theorem 1 (Eq.~(5)) in the maintext}	
	\begin{theorem}[ConDC in delay-embedding space, Theorem 1 in main text]
		If $x$ is the ConDC of $y$ in dynamics
		\begin{equation}
		\begin{cases}
		x_{t+1} = g(x_t, x_{t-1}, \dots, x_{t-p}, \varepsilon_{x,t}),\\
		y_{t+1} = f(x_t, x_{t-1}, \dots, x_{t-p}, y_t, y_{t-1}, \dots, y_{t-p}, \varepsilon_{y,t}),
		\end{cases}\label{eq_xy}
		\end{equation}
		and $\vec{X}_t$, $\vec{Y}_t$ are the time-delayed vectors, respectively, then there exists a smooth projection operator $\vec{F}$ in generic sense such that
		\begin{equation}
		\vec{X}_t = \vec{F}(\vec{Y}_{t+1}), \label{eq_proYX}
		\end{equation}
		when the time-delayed length satisfies $L\geqslant 2d$, where $d$ is the inner dimension of the attractive manifold.
	\end{theorem}
	
	\textbf{Proof:} For dynamics of \eqnref{eq_xy}, denote trajectories of the system as $\{\vec{m}_{x, y}(t) = (x_t, x_{t-1},\\ \dots, x_{t-p}, y_t, y_{t-1}, \dots, y_{t-p})^T| t\geq 0 \}$, whose attractive manifold is $\mathcal{O}$ when $t\rightarrow +\infty$. Because the dynamics of $x$ is autonomous, we denote the trajectories of $x$ as $\{\vec{m}_{x}(t) = (x_t, x_{t-1}, \dots, x_{t-p})^T\\| t\geq 0\}$ whose attractive manifold is $\mathcal{O}_X$ when $t\rightarrow +\infty$. The noise terms are set to be sampled from $\Sigma_{X,Y} = \{\vec{\eps}_{x, y, t} = (\eps_{x, t},\dots, \eps_{x, t-p}, \eps_{y, t}, \dots, \eps_{y, t-p})^T| t\geq 0\}$
	and 
	$\Sigma_X = \{\vec{\eps}_{x, t}=(\eps_{x, t}, \eps_{x, t-1}, \dots, \eps_{x, t-p})^T| t\geq 0\}$. According to the seminal stochastic version of Takens' embedding theorem \cite{takens1981detecting, sauer1991embedology, stark1997takens, cummins2015efficacy}, for open dense sets $\vec{\eps}_{x} \subseteq \Sigma_X$, $\vec{\eps}_{x, y} \subseteq \Sigma_Y$ and delay dimension $L\geqslant 2d$,  we can obtain
	\begin{eqnarray}
	&\vec{X}_t = \varphi_{X}^{\vec{\eps}_{x}}(\vec{m}_x(t)), \label{eq_emX}\\
	&\vec{Y}_{t+1} = \varphi_{Y}^{\vec{\eps}_{x,y}}(\vec{m}_{x, y}(t)),  \label{eq_emY}
	\end{eqnarray}
	where $\varphi_X^{\vec{\eps}_x}: \mathcal{O}_X\rightarrow \mathcal{M}_X$ and $\varphi_Y^{\vec{\eps}_{x,y}}: \mathcal{O}\rightarrow \mathcal{M}_Y$ are diffeomorphisms in the generic sense, i.e. $\varphi_X^{\vec{\eps}_x}$ and $\varphi_Y^{\vec{\eps}_{x,y}}$ are one-to-one with differentiable inverse maps. The $\mathcal{M}_X$ and $\mathcal{M}_Y$ represent manifolds formed by time-delayed vectors $\vec{X}_t$ and $\vec{Y}_t$, respectively. By definition, the mapping $\Pi: \vec{m}_{x, y}(t) \mapsto \vec{m}_{x}(t)$ is a projection. Together with Eqs.~\eqref{eq_emX} and \eqref{eq_emY}, we have
	\begin{equation}
	\vec{X}_t = \vec{F}_{\vec{\eps}_{x,y}}(\vec{Y}_{t+1}), \label{eq_proYX2}
	\end{equation}
	where $\vec{F}_{\vec{\eps}_{x,y}} = \varphi_X^{\vec{\eps}_x}\circ\Pi\circ(\varphi_Y^{\vec{\eps}_{x,y}})^{-1}$. For simplicity, we use the notation $\vec{F}$ in Eq.~(5) in the main text. The irreversibility of $\vec{F}$ is due to the projection operator $\Pi$. $\blacksquare$

\subsection{Numerical algorithm for IEE}
The interventional embedding entropy (IEE) criterion from $x$ to $y$ is 
\begin{equation}
	\text{IEE}[x\rightarrow y] := \text{CMI}(\delta\vec{X}_t, \delta\vec{Y}_{t+1}| \vec{Y}_{t+1}), \label{eq_cs1}
\end{equation}
where $\vec{X}_t, \vec{Y}_{t+1}$ are delay-embedding vectors, and $\delta\vec{X}_t, \delta\vec{Y}_{t+1}$ are corresponding deviations caused by interventions. To numerically approximate \eqnref{eq_cs1} solely from the observed data, we have

%\begin{linenomathNonumbers}
\begin{align}
%\begin{eqnarray}
	&\text{IEE}[x\rightarrow y] \nonumber\\
	&=\iiint p(\vec{Y}_{t+1})p(\delta\vec{X}_t, \delta\vec{Y}_{t+1}| \vec{Y}_{t+1})\log \frac{p(\delta\vec{X}_t, \delta\vec{Y}_{t+1} | \vec{Y}_{t+1})}{p(\delta\vec{X}_t|\vec{Y}_{t+1})p(\delta\vec{Y}_{t+1}|\vec{Y}_{t+1})} \d\delta\vec{X}_t\d\delta\vec{Y}_{t+1}\d\vec{Y}_{t+1}\nonumber\\
	&\approx \frac{1}{N}\sum_{n=1}^N\iint p(\delta\vec{X}_{t_n}, \delta\vec{Y}_{t_n+1}| \vec{Y}_{t_n+1})\log \frac{p(\delta\vec{X}_{t_n}, \delta\vec{Y}_{t_n+1} | \vec{Y}_{t_n+1})}{p(\delta\vec{X}_{t_n}|\vec{Y}_{t_n+1})p(\delta\vec{Y}_{t_n+1}|\vec{Y}_{t_n+1})} \d\delta\vec{X}_{t_n}\d\delta\vec{Y}_{t_n+1}\nonumber\\
	&=\frac{1}{N}\sum_{n=1}^N \text{MI}(\delta\vec{X}_{t_n}|_{\vec{Y}_{t_n+1}}, \delta\vec{Y}_{t_n+1}|_{\vec{Y}_{t_n+1}}), \label{eq_IEEapprox}
%\end{eqnarray}
\end{align}
%\end{linenomathNonumbers}

\noindent where $N$ is the total number of points in the delayed-embedding space, $\delta\vec{Y}_{t_n+1}|_{\vec{Y}_{t_n+1}}$ is sampled as $\vec{Y}_{t_k+1} - \vec{Y}_{t_n+1}$ representing the interventional effect around $\vec{Y}_{t_n+1}$, $\vec{Y}_{t_k+1}$ is the $k$th nearest neighbor of $\vec{Y}_{t_n+1}$ in the embedding space, $\delta\vec{X}_{t_n}|_{\vec{Y}_{t_n+1}}$ is sampled as $\vec{X}_{t_k} - \vec{X}_{t_n}$ with the same time label $t_k$ representing the interventional cause around $\vec{X}_{t_n}$, and $\text{MI}$ is the mutual information

\begin{equation}
\text{MI}(\vec{x}, \vec{y}) = \iint p(\vec{x}, \vec{y})\log \frac{p(\vec{x}, \vec{y})}{p(\vec{x})p(\vec{y})}\d\vec{x}\d\vec{y},
\end{equation}
which can be realized by the kNN algorithm in high dimensional cases \cite{kraskov2004estimating, kozachenko1987sample, lindner2011trentool}. We used the symbol $t_k$ instead of $t_{k}^{(n)}$ for simplicity, but we should remember that the time labels for nearest neighbors change for different points $\vec{Y}_{t_n+1}$. Table~1 in the main text lists steps for the IEE algorithm, and the code is available at \href{https://github.com/smsxiaomayi/IEE}{https://github.com/smsxiaomayi/IEE}.

\subsection{Two-node Logistic dynamics}
\begin{figure}[!htbp]
	\centering
	\includegraphics[width= \textwidth]{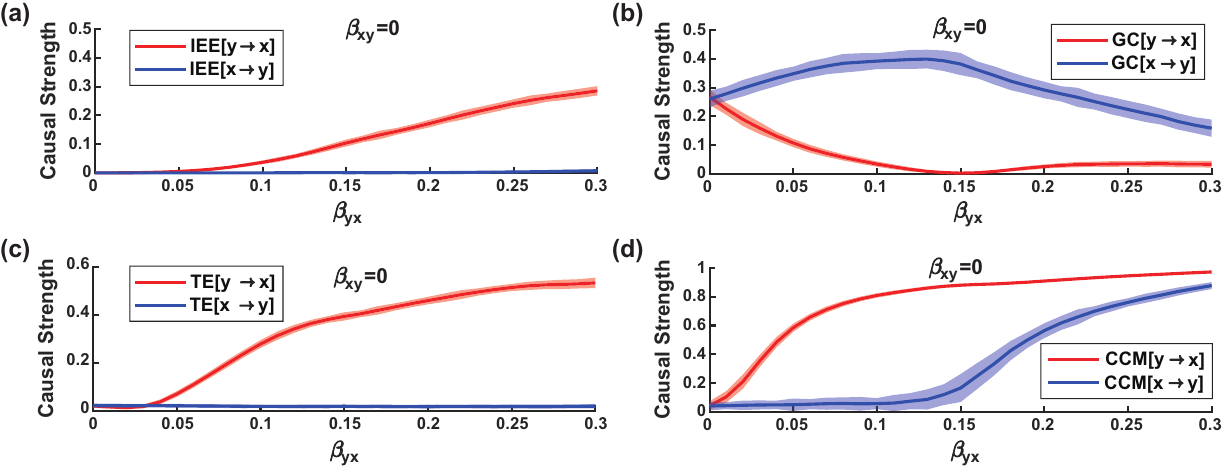}
	\caption{Comparison of (a) IEE, (b) GC, (c) TE, and (d) CCM on the two-node Logistic system ($\beta_{xy}=0$). The parameter $\beta_{yx}$ increases from 0 to 0.3. GC is non-monotonic. $\text{TE}[y\rightarrow x]$ decreases from $0$ to $0.025$ and being smaller than $\text{TE}[x\rightarrow y]$, which induces the false negative problem. CCM has the false positive problem for $x\rightarrow y$, especially when $\beta_{yx}>0.15$.}\label{fig_2log00}
\end{figure}

The two-node Logistic dynamics is
\begin{equation}
\begin{cases}
x_{t+1} = 3.7 \big[ (1-\beta_{yx})x_t\left(1-x_t\right) + \beta_{yx}y_t\left(1-y_t\right) \big] + \eps_{x, t},\\
y_{t+1} = 3.7 y_t \big[ 1 - (1-\beta_{xy})y_t - \beta_{xy}x_t \big] + \eps_{y, t},
\end{cases}
\end{equation}
where parameter $\beta_{xy}$ adjusts the causality from $x$ to $y$, $\beta_{yx}$ controls the causality from $y$ to $x$, and $\eps_{x, t}, \eps_{y,t}$ are independent Gaussian variables representing the noise. 
In Fig.~2(A) in the main text, we set $\beta_{xy}\equiv 0$ and let $\beta_{yx}$ change from $0$ to $0.30$. When $\beta_{yx}\neq 0$, there is no IntDC from $x$ to $y$ while there exists unidirectional IntDC from $y$ to $x$. For each $\beta_{yx}$, we simulated an ensemble with $100$ trajectories, and calculated $\text{IEE}[y\rightarrow x]$ and $\text{IEE}[x\rightarrow y]$ to obtain the statistics such as the mean value and standard deviation. Parameters were chosen as: data length (i.e. number of data points) $N = 1000$, delayed lag $L=2$, number of nearest neighbors $K = 40$, and noise standard deviation $\sigma = 0.01$ (i.e. normal distribution $\mathcal{N}(0, \sigma^2)$). The same time series were used to calculated GC, TE, and CCM, whose results are shown in \figref{fig_2log00}. Table~\ref{tab_s1} shows mean values and standard deviations of the four causal indices from $y$ to $x$ under $100$ simulations, and \tabref{tab_s2} is for the causality from $x$ to $y$.

\begin{figure}[!htbp]
	\centering
	\includegraphics[width= \textwidth]{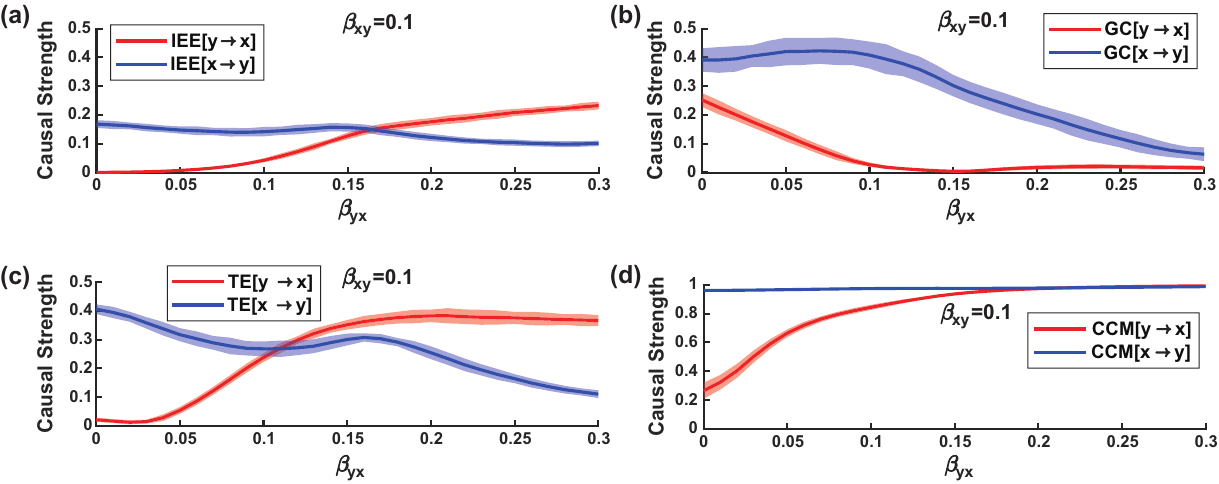}
	\caption{Comparison of (a) IEE, (b) GC, (c) TE, and (d) CCM on the two-node Logistic system ($\beta_{xy}=0.1$). The parameter $\beta_{yx}$ increases from 0 to 0.3. GC is non-monotonic. $\text{TE}[y\rightarrow x]$ decreases from $0$ to $0.025$ and after $0.2$. $\text{CCM}[y\rightarrow x]$ has the false positive problem even when $\beta_{yx}=0$.}\label{fig_2log01}
\end{figure}

In Fig.~2(B) in the main text, we set $\beta_{xy}\equiv 0.1$ and let $\beta_{yx}$ change from $0$ to $0.30$. When $\beta_{yx}\neq 0$, there is bidirectional IntDC between $x$ and $y$. For each $\beta_{yx}$, we simulated $100$ trajectories, and calculated $\text{IEE}[y\rightarrow x]$ and $\text{IEE}[x\rightarrow y]$. Parameters were chosen as: data length $N = 1000$, delayed lag $L=2$, number of nearest neighbors $K = 40$, and noise standard deviation $\sigma = 0.01$. The same time series were used to calculated GC, TE, and CCM, whose results are shown in \figref{fig_2log01}. Table~\ref{tab_s3} shows mean values and standard deviations of the four causal indices from $y$ to $x$ under $100$ simulations, and \tabref{tab_s4} is for the causality from $x$ to $y$.

In Fig.~2(D) in the main text, we set $\beta_{xy} \equiv 0$ and let $\beta_{yx} = 0.15, 0.125, 0.1, 0.075, 0.05$. Delayed lag $L$ changed from $2$ to $7$. For each $\beta_{yx}$ and $L$, we simulated $50$ trajectories and calculated $\text{IEE}[y\rightarrow x]$. Parameters were chosen as: data length $N = 1000$, number of nearest neighbors $K = 40$, and  noise standard deviation $\sigma = 0.01$.

In Fig.~2(E) in the main text, we set $\beta_{xy} \equiv 0$ and let $\beta_{yx} = 0.15, 0.125, 0.1, 0.075, 0.05$. Number of nearest neighbors $K$ changed from $35$ to $45$. For each $\beta_{yx}$ and $K$, we simulated $50$ trajectories and calculated $\text{IEE}[y\rightarrow x]$. Parameters were chosen as: data length $N = 1000$, delayed lag $L = 2$, and  noise standard deviation $\sigma = 0.01$.

In Fig.~2(F) in the main text, we set $\beta_{xy} \equiv 0$ and let $\beta_{yx} = 0.15, 0.125, 0.1, 0.075, 0.05$. Date length $N$ changed from $600$ to $1100$. For each $\beta_{yx}$ and $N$, we simulated $50$ trajectories and calculated $\text{IEE}[y\rightarrow x]$. Parameters were chosen as: delayed lag $L=2$, number of nearest neighbors $K = 40$, and  noise standard deviation $\sigma = 0.01$.

In Fig.~2(G) in the main text, we set $\beta_{xy} \equiv 0$ and let $\beta_{yx} = 0.15, 0.125, 0.1, 0.075, 0.05$.  Noise standard deviation $\sigma$ changed from $10^{-2.5}$ to $10^{-2}$. For each $\beta_{yx}$ and $\sigma$, we simulated $50$ trajectories and calculated $\text{IEE}[y\rightarrow x]$. Parameters were chosen as: data length $N = 1000$, delayed lag $L=2$, and number of nearest neighbors $K = 40$.

\subsection{Three-node Logistic dynamics} 
\begin{figure}[!htbp]
	\centering
	\includegraphics[width= \textwidth]{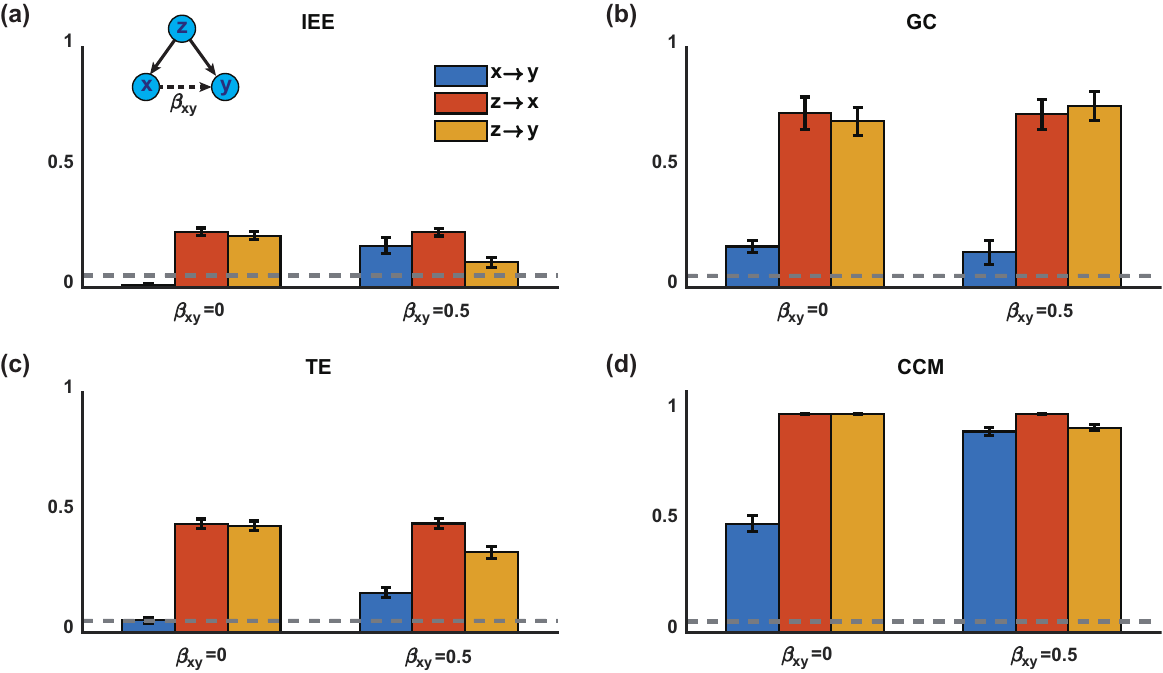}
	\caption{Comparison of (a) IEE, (b) GC, (c) TE, and (d) CCM on the three-node Logistic system. There are constant causal effects from $z$ to $x$ and from $z$ to $y$. The variable $z$ acts as a confounder. When  $\beta_{xy}=0$ there is no causality from $x$ to $y$, while causality exists from $x$ to $y$ when $\beta_{xy}=0.5$. The gray dashed line represents $0.05$ for reference. IEE for IntDC accurately distinguish the causal strength between $x$ and $y$, while the other three indices for ConDC suffer from false-positive detections when a confounding variable $z$ exists.}\label{fig_3logNodes}
\end{figure}

We used a three-node Logistic dynamics to test the performance of IEE when there exists confounding variable. The dynamics is
\begin{equation}
\begin{cases}
x_{t+1} = \gamma_xx_t[1-(1-\frac{\beta_{zx}}{\gamma_x})x_t-\frac{\beta_{zx}}{\gamma_x}z_t]+\eps_{x,t},\\
y_{t+1}= \gamma_yy_t[1-(1-\frac{\beta_{xy}+\beta_{zy}}{\gamma_y})y_t-\frac{\beta_{xy}}{\gamma_y}x_t-\frac{\beta_{zy}}{\gamma_y}z_t] +\eps_{y,t},\\
z_{t+1} = \gamma_zz_t(1-z_t)+\eps_{z,t},
\end{cases}
\end{equation}
where $\gamma_x = \gamma_y=\gamma_z = 3.7$, noises $\eps_{x,t}, \eps_{y,t}, \eps_{z,t}$ are independent Gaussian variables $\mathcal{N}(0, \sigma^2)$ with the standard deviation $\sigma = 0.001$, $\beta_{xy}$ adjusts the causal strength from $x$ to $y$, and parameters $\beta_{zx} = \beta_{zy} = 0.5$ represent the causality from $z$ to $x$ and $z$ to $y$, respectively. There are constant causality from $z$ to $x$ and from $z$ to $y$, and $z$ is a confounder. When $\beta_{xy} = 0$, we simulated $100$ trajectories and calculated IEE, GC, TE, and CCM for edges $x\rightarrow y$, $z\rightarrow x$, and $z\rightarrow y$. When $\beta_{xy} = 0.5$, we did the same computation. Parameters were chosen as: data length $N=1000$, delayed lag $L=2$, number of nearest neighbors $K = 20$. In \figref{fig_3logNodes}, we showed the results for the four indices, respectively. The result for IEE is also displayed in Fig.~2(C) in the main text. The causal strength from $x$ to $y$ were correct for IEE when $\beta_{xy}=0$, while the GC/TE/CCM exhibited false positives in some degree. The gray dashed line is $0.05$ for a reference.

\subsection{10-node coupled Henon maps}
The dynamics of the 10-node coupled Henon maps are
\begin{subequations}
\begin{eqnarray}
	x_{i, t+1} &=& 1 - ax_{i, t}^2 + bx_{i, t-1}  + \sigma\varepsilon_{i, t}, \qquad i = 1,\\
	x_{i, t+1} &=& 1-a(\beta x_{i-1, t}+(1-\beta)x_{i, t})^2 + bx_{i, t-1} + \sigma\varepsilon_{i, t}, \qquad i = 2, 3, \dots, 10,
\end{eqnarray}
\end{subequations}
where parameters $a = 1.4$, $b = 0.3$ are constant, $\sigma = 0.002$ is the noise standard deviation, noise term $\eps_{i, t}$ are independent standard Gaussian random variables, and $\beta = 0.6$ is the coupling coefficient representing the causality from variable $x_{i}$ to $x_{i+1}$. We sampled $500$ time points for each trial from initial point $x_{i, 1} = 0.5,~i=1,2,\dots, 10$.

\begin{figure}[!htbp]
	\centering
	\includegraphics[width = 0.95\textwidth]{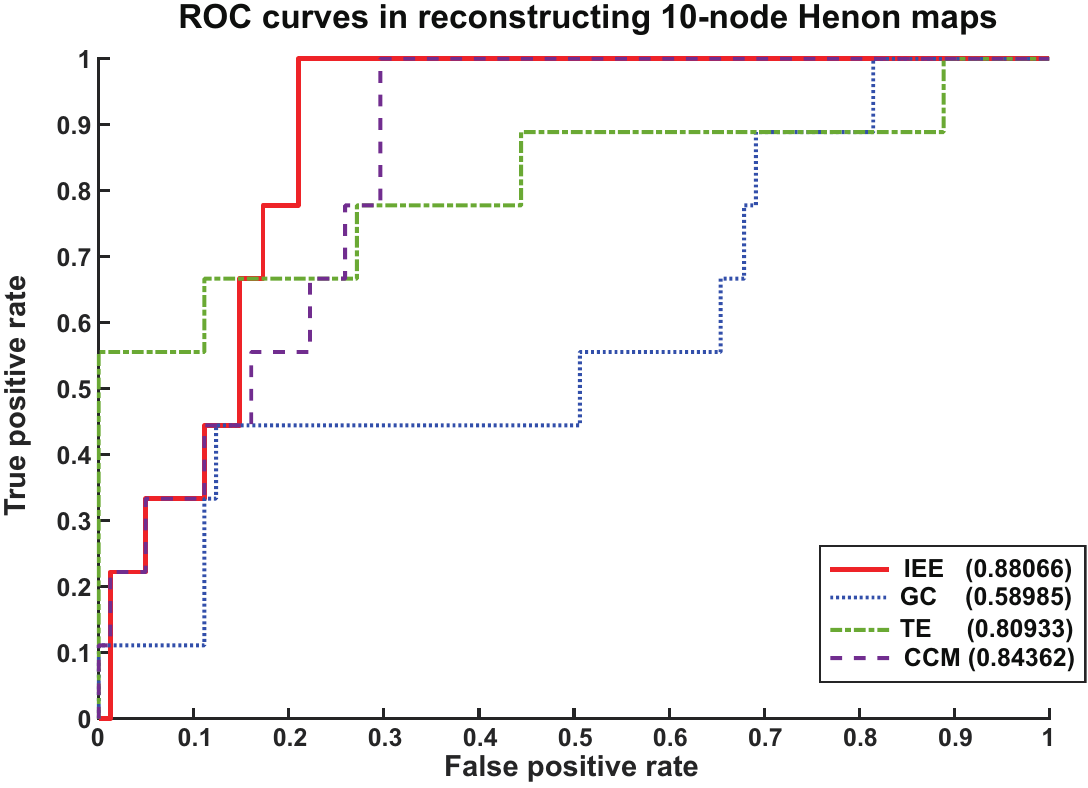}
	\caption{The ROC curves for IEE, GC, TE, and CCM when reconstructing the 10-node Honon maps in a representative simulation. The values in the legend are AUC values for the corresponding ROC curves.} \label{fig_rocHenon}
\end{figure}

In Fig.~3(B) in the main text, we simulated $20$ trials and calculated $20$ AUC values for IEE, GC, TE, and CCM in reconstructing the network. ROC curves for one trial can be found in \figref{fig_rocHenon}. Parameters were chosen as: delayed lag $L = 2$, number of nearest neighbors $K = 10$.

\begin{figure}[!htbp]
	\centering
	\includegraphics[width = \textwidth]{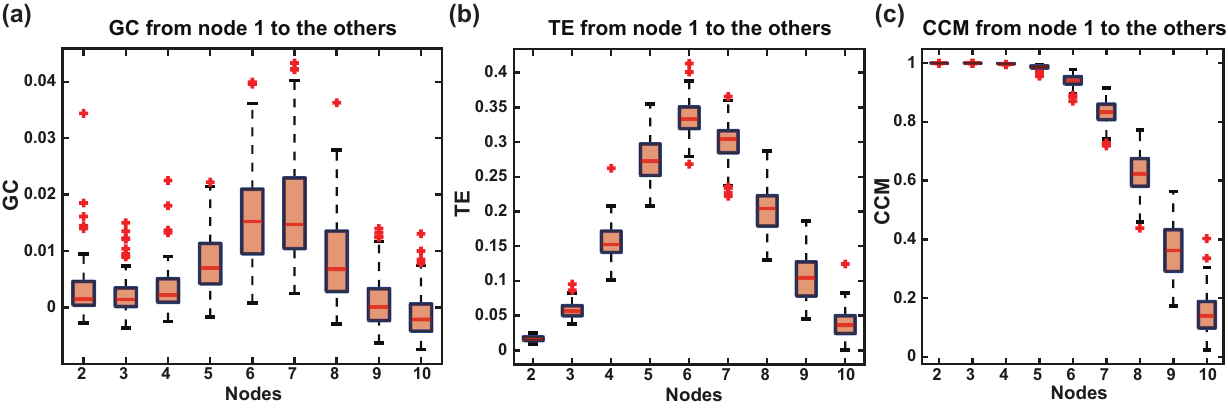}
	\caption{GC, TE, and CCM from the Node 1 to the other nine nodes in the 10-node coupled Henon-map network. GC and TE are not monotonically decreasing. CCM presents strong causalities from Node 1 to the Nodes 2-6.} \label{fig_henonfrom1}
\end{figure}

In Fig.~3(C) in the main text, we simulated $100$ trials and calculated the causal strength from node 1 to the other nine nodes. Results for GC, TE, and CCM can be found in \figref{fig_henonfrom1}. Parameters were chosen as: delayed lag $L = 2$, number of nearest neighbors $K = 10$.

\begin{figure}[!htbp]
	\centering
	\includegraphics[width = \textwidth]{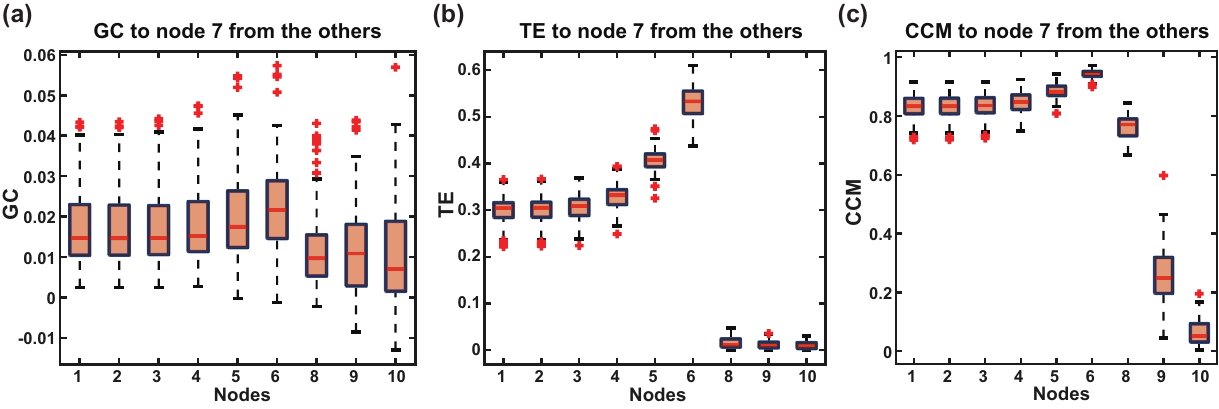}
	\caption{GC, TE, and CCM to the Node 7 from the other nine nodes in the 10-node coupled Henon-map network. GC fails the causal detection, while CCM has a false positive result from Node 8 to Node 7.} \label{fig_henonto7}
\end{figure}

In Fig.~3(D) in the main text, we simulated $100$ trials and calculated the causal strength received by node 7 from the other nine nodes. Results for GC, TE, and CCM can be found in \figref{fig_henonto7}. Parameters were chosen as: delayed lag $L = 2$, number of nearest neighbors $K = 10$. 

\subsection{Chaotic neural networks}
The chaotic neural network (chNN) is composed of $N_{\text{node}}$ chaotic neurons, each of which is described by an output variable $x_i$ and two internal state variables: a feedback state variable $y_i$, and a refractory state variable $z_i$ \cite{aihara1990chaotic, adachi1997associative}.
The dynamics of chNN is given by
\begin{subequations}
\begin{eqnarray}
	x_{i, t+1} &=& \text{tanh}[s(y_{i, t+1}+z_{i, t+1})],\\
	y_{i, t+1} &=& k_fy_{i, t} + \beta\sum_{j=1}^{N_{\text{node}}} w_{ij}x_{j,t}+\sigma_f\varepsilon_{y_i, t},\\
	z_{i, t+1} &=& k_rz_{i, t}-\alpha x_{i, t}+b_i+\sigma_r\varepsilon_{z_i,  t},
\end{eqnarray}
\end{subequations}
where $i = 1, 2, \dots, N_{\text{node}}$, $s>0$ is the steepness parameter, $k_f, k_r\in[0, 1)$ are decay parameters, $\alpha, \beta$ are coupling strengths, $w_{ij}$ is the weight between $x_j$ and $y_i$, $\vec{b} = (b_1, b_2, \dots, b_{N_{\text{node}}})^\top \in \mathbb{R}^{N_{\text{node}}}$ denotes bias, $\sigma_f, \sigma_r>0$ are noise standard deviations, and $\eps_{y_i,t}, \eps_{z_i,t}$ are standard Gaussian noise terms. The dynamics can be written into a vector-matrix form

\begin{equation}
\begin{cases}
	\vec{x}_{t+1}&= ~\text{tanh}(s(\vec{y}_{t+1}+\vec{z}_{t+1})), \\
	\vec{y}_{t+1}&= ~k_f \vec{y}_t + \beta\vec{W}\vec{x}_t + \sigma_f \vec{\eps}_{\vec{y}, t}, \\
	\vec{z}_{t+1}&= ~k_{r} \vec{z}_t - \alpha\vec{x}_t + \vec{b} + \sigma_r \vec{\eps}_{\vec{z}, t},
\end{cases}
\end{equation}
where $\vec{W}=(w_{ij})_{N_{\text{node}}\times N_{\text{node}}}$, $\vec{x} = (x_1, x_2, \dots, x_{N_{\text{node}}})^\top$, $\vec{y} = (y_1, y_2, \dots, y_{N_{\text{node}}})^\top$, and $\vec{z}=(z_1, z_2, \dots, z_{N_{\text{node}}})^\top$ .

In our simulation, we chose $N_{\text{node}}=10$ and typical parameter values $s=20$, $k_{f}=0.2$, $k_{r}=0.95$, and $b_i=0.4$ for all $i$;
the coupling matrix $\mathbf{W}=(w_{ij})$ was constructed such that every neuron $i$ received feedback inputs from two other neurons $j_1,j_2\neq i$ with nonnegative couplings $w_{ij_1} + w_{ij_2} = 1$,
where $(i, j), (w_{ij_1}, w_{ij_2})$ were generated randomly from uniform distributions; the coupling strengths and noise levels were $\alpha=4$, $\beta=0.2$ and $\sigma_{f}=\sigma_{r}=0.05$, respectively. We generated $1000$ time series $\vec{x}_{\text{obs}}$ with data length $1000$ after the relaxation time in $100$ different networks ($10$ time series for each network).  The values of IEE, GC, TE,  and CCM were calculated from $\vec{x}_{\text{obs}}$ solely. When calculating causal indices, we chose delayed lag $L = 3$, number of nearest neighbors $K = 40$. For the perturbed systems, we recorded the output $\vec{x}_{\text{per}}^{(i)}$ by removing neuron $i$ (set as a constant zero). The Kullback-Leibler divergence (KLD) was used to measure the true influence of the perturbation between neurons, i.e.
\begin{equation}
D_{ij}\triangleq\text{KLD}[p_j(\vec{x}_{\text{obs}})||p_j(\vec{x}_{\text{per}}^{(i)})] = \iint p_j(\vec{x}_{\text{obs}}) \log \frac{p_j(\vec{x}_{\text{obs}})}{p_j(\vec{x}_{\text{per}}^{(i)})}\d \vec{x}_{\text{obs}}\d\vec{x}_{\text{per}}^{(i)},
\end{equation}
where $i$ and $j$ are two different neurons, the perturbation is conducted on $i$, and $p_j(\vec{x}_{\text{obs}})$ and $p_{j}(\vec{x}_{\text{per}}^{(i)})$ represent the stationary probability density functions of $j$ before and after the perturbation, respectively.

In Figs.~4(A-D) in the main text, the linear regression $y=ax+b$ was conducted on KLD and the four causal indices, i.e. IEE, GC, TE, and CCM. In Fig.~4(E) in the main text, we used the cosine similarity between KLD and the four causal indices. The cosine similarity between vectors $\vec{A}$ and $\vec{B}$ is defined as
\begin{equation}
S_C(\vec{A}, \vec{B}) = \frac{\vec{A}\cdot\vec{B}}{||\vec{A}||~||\vec{B}||} = \frac{\sum\limits_{i=1}^n {A_iB_i}}{\sqrt{\sum\limits_{i=1}^nA_i^2}\cdot\sqrt{\sum\limits_{i=1}^nB_i^2}}, \label{eq_Sc}
\end{equation}
where $A_i$ and $B_i$ are the $i$th components of $\vec{A}$ and $\vec{B}$, respectively.

\subsection{\emph{Caenorhabditis elegans} (\emph{C.\,elegans}) neural activity dataset}
We collected the calcium fluorescence imaging time series from neurons of a freely moving \emph{C.\,elegans}, published in~\cite{Kato2015}. \emph{C.\,elegans} is a widely used model organism in neuroscience, and its nervous system connectomes, consisting of $302$ neurons in the hermaphrodite, has been comprehensively mapped~\cite{Cook2019}. Due to the potential presence of strong synchrony among different neurons, inferring the causal network over the entire system can be challenging. Recently, Banerjee et al.~\cite{Banerjee2023} attempted synaptic connection inference on a
symmetrically \emph{folded} subnetwork composed of the most active motor neurons in the system. In our study, we carried out causality inference on a \emph{reduced} network of neuron assemblies, formed by clustering neurons with highly similar activity.

\subsubsection{Data Description}
The experimental multi-neuron time series data of calcium imaging of~\emph{C.\,elegans} are obtained from \cite{Kato2015}, with a sampling rate of $2.13$Hz
and a temporal length of $18$~min (i.e. $2300$ samples). We focused on $N=31$ neurons with specified functions (such as AVAL, AVAR, etc.). The $31$ names of neurons can be found in \tabref{tab:sm_celegans_clustering}. We chose $T=1750$ time points, during which the correlation pattern of the neural activity remains stable. Then, the following preprocessing procedures were conducted:
\begin{itemize}
\item \emph{Detrending}: Considering the non-negative property of fluorescence intensity,	the time series for each neuron is first shifted to have a zero minimum; then, each time series is dynamically normalized via dividing its $5$-point moving average by its $401$-point moving average. This transformation acts similarly to a band-pass filter, removing the trend in time series and stabilizing them.
\item \emph{Dimension reduction}: We normalized the time series to have unit second raw moments and denoted it as $\vec{X} = [\vec{x}_1, \vec{x}_2, \dots, \vec{x}_N]\in \mathbb{R}^{T\times N}$, where $\vec{x}_i$ is the time series of neuron $i$. Let $\vec{S} = T^{-1}\vec{X}^\top\vec{X} = (s_{ij})$ be the cosine similarity matrix, where $s_{ij}$ represents the cosine of the angle between time series from the $i$th and $j$th neurons. Apply eigen-decomposition on $\vec{S}$ as $\vec{S}=\vec{Q}^\top\vec{\Lambda}\vec{Q}$, where $\vec{Q}$ is an orthogonal matrix and $\vec{\Lambda}$ is a non-negative diagonal matrix. Thus, $\vec{P} = \vec{\Lambda}^{1/2}\vec{Q}=[\vec{p}_1, \vec{p}_2, \dots, \vec{p}_N]$ provides a dimension reduction of $\vec{X}$. The vector $\vec{p}_i$ lies on the unit sphere in $\mathbb{R}^N$ and $s_{ij} = \vec{p}_i^\top \vec{p}_j$, where $ i, j = 1, 2, \dots, N$.
\item \emph{Clustering}: Then, we clustered the $N$ neurons into $M$ neuron assemblies, with sizes	$N_1,\,N_2,\dots, N_M$, respectively. Denote $\vec{X}_m\in\mathbb{R}^{T\times N_m}$ and $\vec{S}_m\in\mathbb{R}^{N_m\times N_m}$ as the submatrices of $\vec{X}$ and $\vec{S}$, which hold the time series data and similarity values within the $m$th assembly, respectively. The collective dynamics $\vec{Y} =[\vec{y}_1, \vec{y}_2, \dots, \vec{y}_M]\in\mathbb{R}^{T\times M}$ of each assembly in the reduced network is then represented by the dominant component of the PCA whitening of $\vec{X}_m$; that is,	given the eigen-decomposition $\vec{S}_m = \vec{Q}_m^\top\vec{\Lambda}_m\vec{Q}_m$ with $\vec{\Lambda}_m=\operatorname{diag}\{\lambda_1(\vec{S}_m),\lambda_2(\vec{S}_m),\dots,\lambda_{N_M}(\vec{S}_m)\}$ in descending order, $\vec{y}_m$ is the first column of $\vec{X}_m\vec{Q}_m^\top\vec{\Lambda}_m^{-1/2}$. The optimal clustering of neurons with similar dynamics is achieved by maximizing the fraction	$\rho$ of the energy that remains in the representing time series $\vec{Y}$, where	
\begin{equation}
	\rho = \frac{1}{N}\sum_{m=1}^M \lambda_1(\vec{S}_m).
\end{equation}
Practically, we found an approximate solution to the above problem simply by clustering points $\{\vec{p}_i\}$ using a k-means algorithm with the designated $M$, and then picked the result with the largest $\rho$	from repeated trials of k-means clustering.	In this study, $M=7$ and $\rho_{\text{max}}=0.9514$ is the best result for clustering, and the clusters are listed in \tabref{tab:sm_celegans_clustering}. Most bilaterally symmetric neuron pairs have naturally fallen into the same assemblies, as designated in~\cite{Banerjee2023}.
\end{itemize}

\begin{figure}[!htbp]
	\centering
	\includegraphics[width=0.97\textwidth]{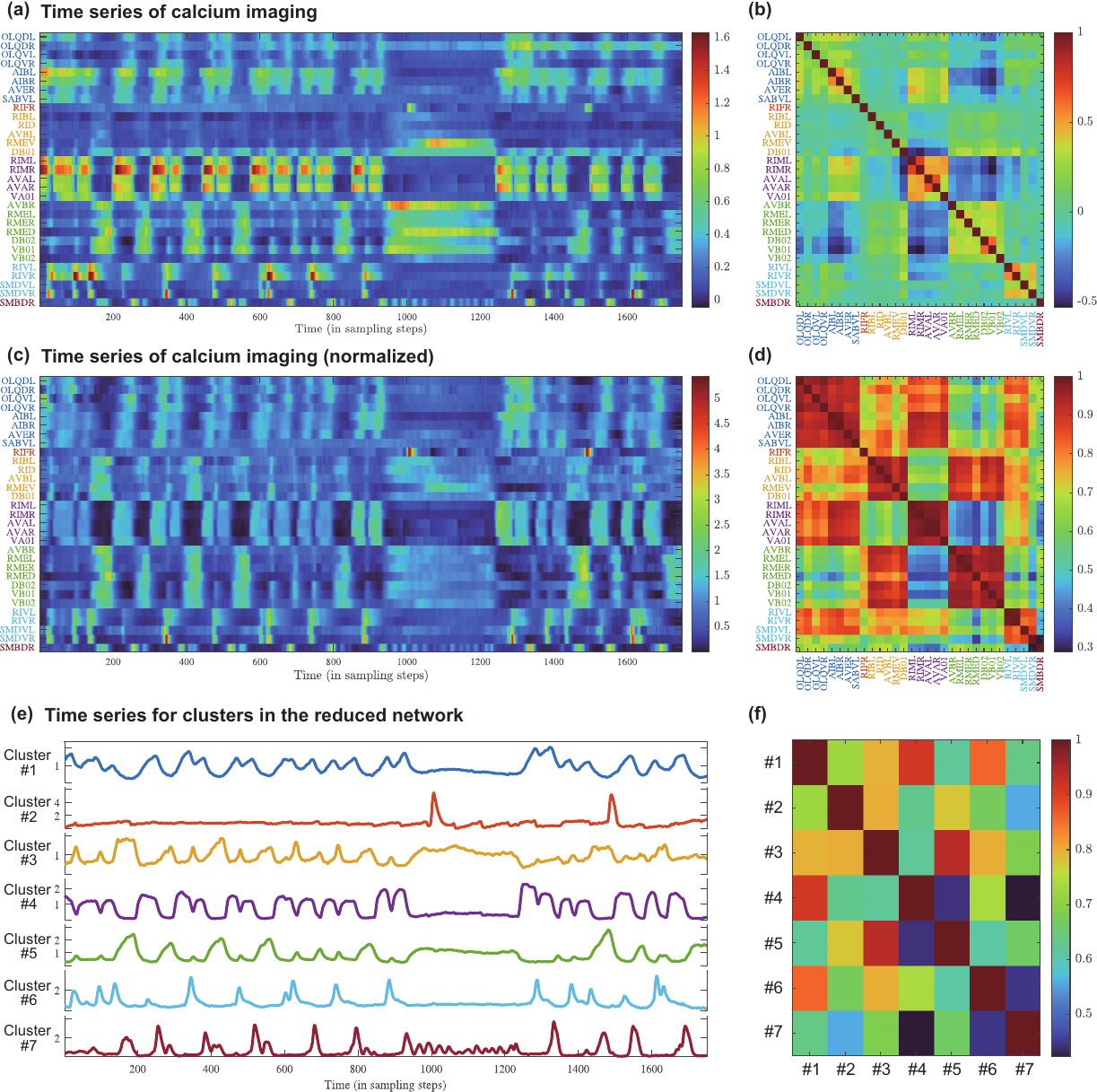}
	\caption{\label{fig:sm_celegans_prep}
		Data in the analysis of calcium imaging of a freely moving \emph{C.\,elegans} worm. (a) Time series for $31$ individual neurons as published in Ref.~\cite{Kato2015}. (b) Pearson correlation matrix for the time difference of the time series shown in (a).  (c) Time series after preprocessing. (d) Cosine similarity matrix for the time series shown in (c).
		In (a-d), the neurons belonging to the same cluster are grouped and their names are color-coded accordingly. (e) Representative time series illustrating the collective behavior of each neuron cluster. (f) Similarity matrix for the time series shown in (e).}
\end{figure}

The original time series for each neuron are displayed in \figref{fig:sm_celegans_prep}(a), which is the same as Fig.~5(B) in the main text. \figref{fig:sm_celegans_prep}(b) is the Pearson correlation matrix of the time difference. The normalized time series $\vec{X}$ are displayed in \figref{fig:sm_celegans_prep}(c). Figure~\ref{fig:sm_celegans_prep}(d) is the cosine similarity matrix $\vec{S}$, which is the same as Fig.~5(C) in the main text. The names of neurons in the same cluster are shown with the same color. The representative time series for individual neuron assemblies are shown in \figref{fig:sm_celegans_prep}(e) with the corresponding cosine similarity matrix shown in \figref{fig:sm_celegans_prep}(f).

%%%%%%%%%%%%%%%%%%%%%%%%%%%%%%%%%%%%%%%%%%%%%%%%%%%%%%%%%%%%%%%%%%%%%%%%%%%%%%%%%%%%%%%%%%%%%%%%%%%%%%%%%%%%

\subsubsection{Causality Analysis} Regarding to the ground truth of causalities between neurons, we referred to the quantitative connectomes of the adult hermaphrodite \textit{C.\,elegans} described in Supplementary Information 5 of \cite{Cook2019} and its further correction in \cite{EmmonsLab2020}. Both chemical and gap junction (electrical) synapses are provided. We denote the connection matrix for the reduced network as $\vec{C} = (c_{ij})_{M\times M}$.  The value $c_{ij}$ is simply obtained by merging the connectivity values from all neurons in the $i$th cluster to all those in the $j$th cluster. Since $M=7$, we obtained a $7\times 7$ directed network with $26$ edges, whose elements range from $1$ to $210$.

For the inferred causal networks obtained from the neural activity data, we calculated four indices, i.e. IEE, GC\cite{Granger1969}, TE~\cite{Schreiber2000}, and CCM\cite{Sugihara2012}, which are denoted as $\vec{C}^{\text{IEE}}$, $\vec{C}^{\text{GC}}$, $\vec{C}^{\text{TE}}$, and $\vec{C}^{\text{CCM}}$, respectively.

In our numerical experiments, each time series was decimated by a factor of $5$ to obtain $5$ time series, and the final causality strength was obtained by averaging the results from the $5$ evaluations. Parameters were chosen as: delayed lag $L=3$, number of nearest neighbors $K = 20$.

\begin{figure}[!hbtp]
	\centering
	\includegraphics[width=0.95\textwidth]{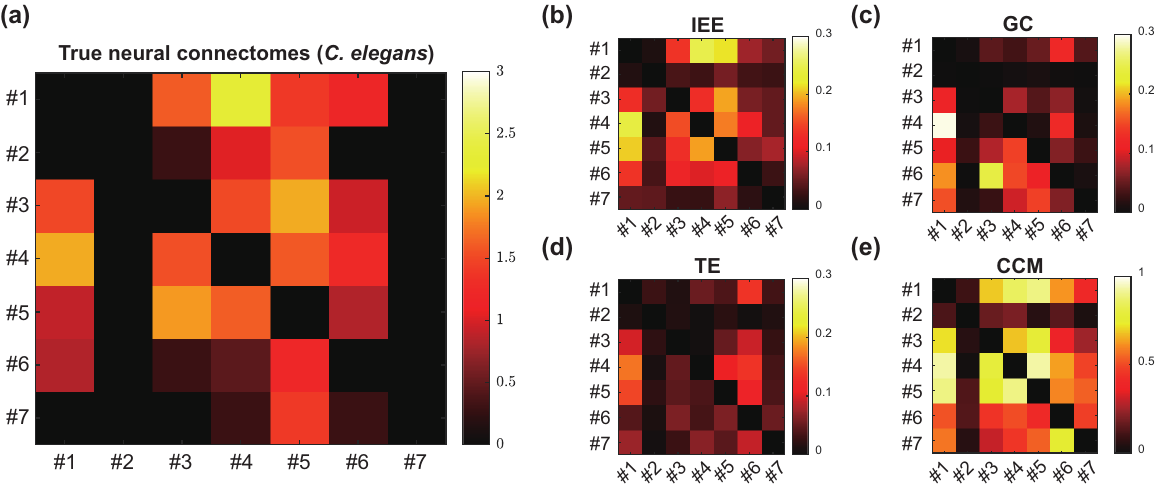}\\
	\caption{\label{fig:sm_celegans_strengths}
		Heat maps of connection strengths between neural clusters in \textit{C.\,elegans}. (a) Ground truth connectomes $\vec{C}$ (in a logarithmic scale, i.e. $\log(1+\vec{C})$). (b-e) Inferred causality strengths by IEE/GC/TE/CCM, denoted as $\vec{C}^{\text{IEE}}/\vec{C}^{\text{GC}}/\vec{C}^{\text{TE}}/\vec{C}^{\text{CCM}}$, respectively.}
\end{figure}

We demonstrated the ground truth causality strengths $\vec{C}$ (in the logarithmic scale) in \figref{fig:sm_celegans_strengths}(a), and the inferred causalities in figs.~\ref{fig:sm_celegans_strengths}(b-e). We further compared the performance of different methods from two perspectives: binary classification and cosine similarity.

\begin{figure}[!thbp]
	\centering
	\includegraphics[width=0.95\textwidth]{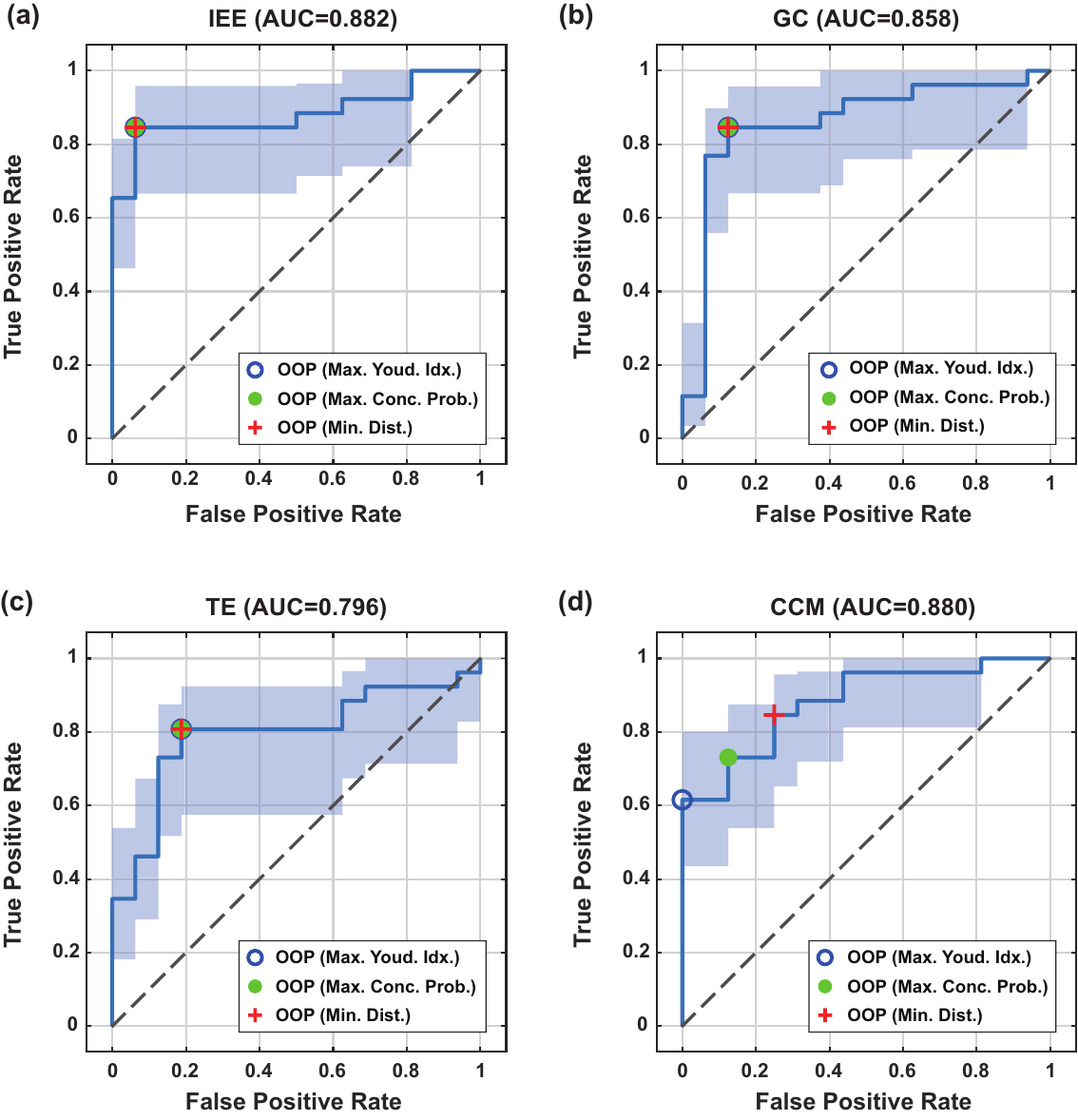}\\
	\caption{\label{fig:sm_celegans_roc}
		Receiver operating characteristic (ROC) curves in the inference of \emph{C.\,elegans} neural connectomes by (a) IEE, (b) GC, (c) TE, and (d) CCM. Optimal operating points (OOPs) obtained by the maximum Youden index (blue circles), the maximum concordance probability (green dots), and the minimum distance to the point $(0, 1)$ (red crosses) are marked. AUC values are listed. The shaded area around the ROC curve represents the $95\%$ confidence interval obtained by bootstrapping.}
\end{figure}

\begin{figure}[!htbp]
	\centering
	\includegraphics[width=0.85\textwidth]{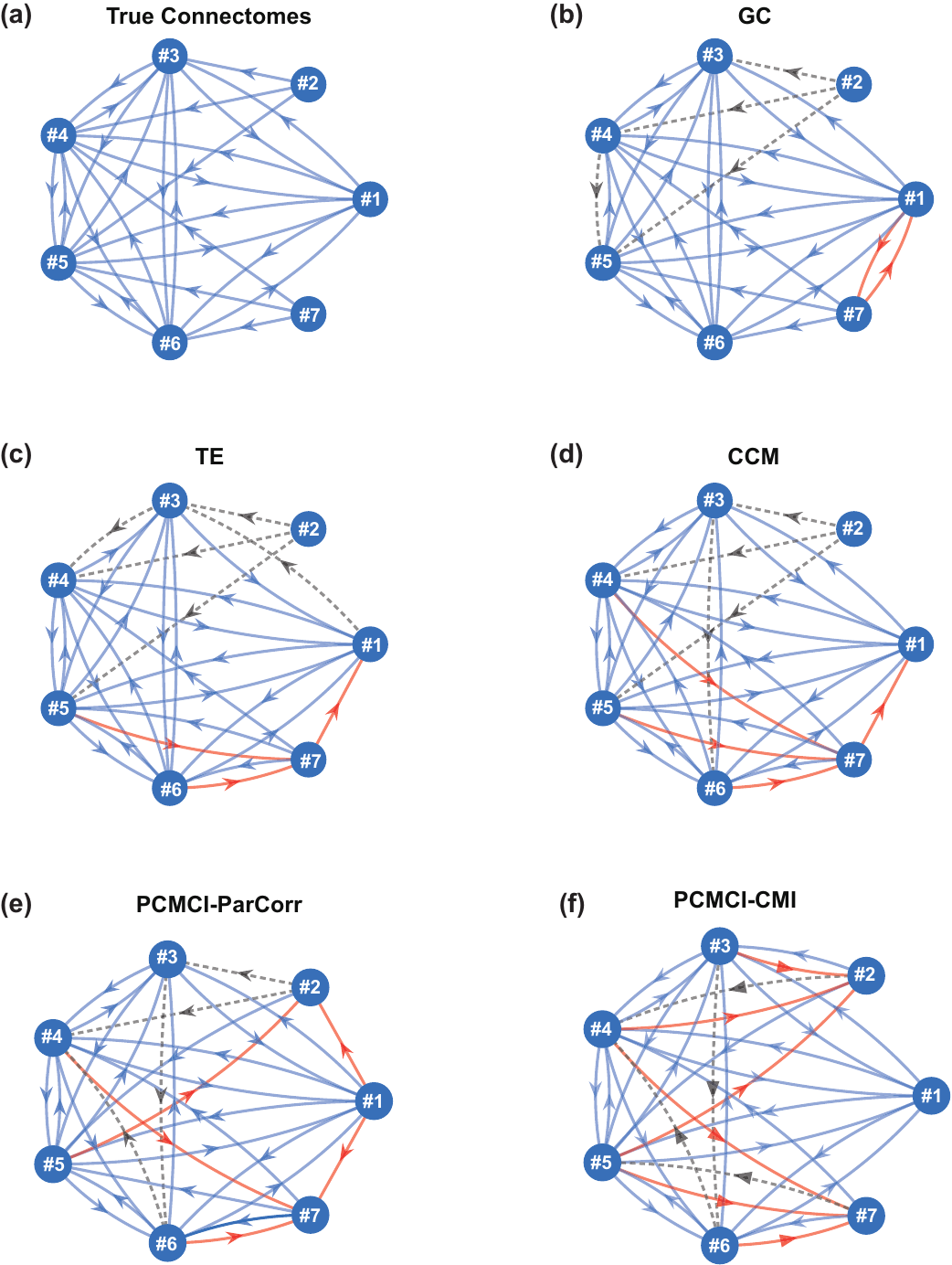}\\
	\caption{\label{fig:sm_celegans_graph}
		(a) The true neural connectomes between the seven clusters in \textit{C.\,elegans}. (b-d) The inferred causal networks by GC/TE/CCM at the optimal operating points given by the minimum distance index. (e-f) The inferred causal networks by PCMCI-ParCorr (with the significance  level $\alpha_{\text{PC}} = 0.05$) and PCMCI-CMI (with the significance level $\alpha_{\text{PC}}=0.01$). The red edges represent false positives, and the black dashed edges are false negatives.}
\end{figure}

\textit{Binary classification.} The ground truth $\vec{C}$ has $26$ causal edges (as ``$1$'') and $16$ non-causal edges (as ``$0$''). We plotted the receiver operating characteristic (ROC) curves for $\vec{C}^{\text{IEE}}$, $\vec{C}^{\text{GC}}$, $\vec{C}^{\text{TE}}$, and $\vec{C}^{\text{CCM}}$ in \figref{fig:sm_celegans_roc}, respectively. Four criteria were used to evaluate the performance of different algorithms:
\begin{itemize}
\item The area under the curve (AUC);
\item Maximum Youden index, i.e. the maximum difference between the true positive rate and the false positive rate;
\item Maximum concordance probability, i.e. the maximum product of sensitivity and specificity;
\item Minimum distance to the point $(0, 1)$ for the ROC curve.
\end{itemize}
The last three indices are proposed for choosing the \emph{optimal} operating point (OOP) on the ROC curve~\cite{Rota2015}. Figure~\ref{fig:sm_celegans_roc} marked the OOPs obtained by different criteria, and Table~2 in the main text listed the numerical values for four algorithms. We also plotted the true connectomes (Fig.~5(D) in the main text or \figref{fig:sm_celegans_graph}(a)) and inferred networks at the OOPs (Fig.~5(E) in the main text for IEE, \figref{fig:sm_celegans_graph}(b) for GC, \figref{fig:sm_celegans_graph}(c) for TE, \figref{fig:sm_celegans_graph}(d) for CCM) as graphs.  At the minimum distance OOP points, IEE/GC/TE/CCM have the number of true positive edges as $22, 22, 21, 22$, the false positive edges as $1, 2, 3, 4$, the true negative edges as $15, 14, 13, 12$, and the false negative edges as $4, 4, 5, 4$, respectively (see \tabref{tab:sm_celegans_OOPperformance}). IEE exhibited the best performance for binary classification in estimating neural connectomes of the \emph{C.\,elegans}.

\textit{Cosine similarity.} The cosine similarity values between the vectorized $\vec{C}$ and $\vec{C}^{\text{IEE}}$/$\vec{C}^{\text{GC}}$/$\vec{C}^{\text{TE}}$/$\vec{C}^{\text{CCM}}$  were calculated by \eqnref{eq_Sc}, respectively. Since $c_{ij}$ ranges widely from $1$ to $210$, we also calculated the cosine similarity values between $\ln (1+c_{ij})$ and the four inferred indices. As listed in Table~2 in the main text, in both cases $c^{\text{IEE}}_{ij}$ shows the largest similarity to the ground truth, which indicates that IEE can effectively quantify the neural connectomes in \textit{C.\,elegans}.

\subsubsection{PCMCI}
We also applied the PCMCI algorithm \cite{runge2019detecting} to infer the \textit{C.~elegans} neural connectomes.	PCMCI consists of two main steps: PC selection and Momentary Conditional Independence (MCI) test. In the PC selection step, the parent variables $\mathcal{P}(X_t^i)$ for each variable $X_t^i$ are identified through an iterative procedure. Conditional independence tests are performed at each iteration using a significance level $\alpha_{\text{PC}}$. In the MCI testing step, an independence test is conducted to determine whether 
\begin{equation}
\text{MCI}: X_{t-\tau}^i \not\perp\!\!\!\perp X_t^j~ |~ \mathcal{P}(X_t^j)\backslash \{X_{t-\tau}^i\}, \mathcal{P}(X_{t-\tau}^i).
\end{equation}
If an edge $X_{t-\tau}^j\rightarrow X_t^i$ consistently passes all tests with $p$-values larger than $\alpha_{\text{PC}}$, a causal link from $X_j$ to $X_i$ is identified. Details of the methodology can be referred to Runge et al. \cite{runge2019detecting}.

In our experiments, we first applied PCMCI with  partial correlation (denoted as PCMCI-ParCorr) and set the significance level $\alpha_{\text{PC}} = 0.05$. The inferred causal network is shown in \figref{fig:sm_celegans_graph}(e). The result includes $22$ true positive edges, $5$ false positive edges, $11$ true negative edges, and $4$ false negative edges (see \tabref{tab:sm_celegans_OOPperformance}).

Then we employed PCMCI with conditional mutual information (denoted as PCMCI-CMI), using a significance level of $\alpha_{\text{PC}} = 0.01$. The CMI test was implemented with a local permutation procedure, in which we generated $1000$ surrogate samples to approximate the null distribution. The resulting network is displayed in \figref{fig:sm_celegans_graph}(f) with $22$ true positive edges, $6$ false positive edges, $10$ true negative edges, and $4$ false negative edges (\tabref{tab:sm_celegans_OOPperformance}).

Comparing PCMCI and IEE, we give a remark that the IEE algorithm can provide comparable causal strengths across different variable pairs in one system, allowing for a unified threshold to binarize the results. In contrast, PCMCI performs edge-wise independence tests, where the choice of significance level $\alpha_{\text{PC}}$ is nontrivial, because different edges may have different null distributions.

\subsection{COVID-19 dataset}
We explored the causal relationships in the inter-regional transmission of coronavirus (COVID-19) among the prefectures of Japan.
\subsubsection{Time series data}
The data on daily confirmed new COVID-19 cases in all $47$ prefectures is available on the website of the Ministry of Health, Labour and Welfare of Japan \cite{MHLW2023}. We collected data spanning a period of $1209$ days from January 16, 2020 to May 8, 2023, as shown in Fig.~\ref{fig:sm_covid_prep}(a). During this period, there were $8$ waves of infection with the $9$th beginning by the end of this period. The overall number of cases exhibited an exponentially increasing trend. We assume that COVID-19 spread across the country in a similar fashion, differing only in the scale from wave to wave. Additionally, a consistent variation pattern within the week is observable in the data. Therefore, the time series are preprocessed by dividing their $7$-day moving averages by their $140$-day moving averages to remove both the day-of-the-week variation and the trend.

\subsubsection{Effective distance as the ground truth}
We defined the concept of \textit{effective distance} as  the ground truth of causality between prefectures based on human mobility. In previous research, the geographical connections, specifically the geodesic distance, between metropolitan areas in Japan have been considered as the ground truth \cite{Shi2022,Tao2023}. However, human mobility models can better reflect the social and economical connections between prefectures \cite{Barbosa2018}.

The definition of \emph{effective distance} in this study is based on the \emph{gravity model	of human mobility} 
\begin{equation}
F_{ij}= K m_i m_j f(d_{ij}), \label{eq_gravity}
\end{equation}
where $F_{ij}$ denotes the mobility flow from the $i$th community to the $j$th, the masses $m_i$ and $m_j$ describe the sizes of these communities, $f(d_{ij})$ represents a deterrence function which decreases with the distance $d_{ij}$ between communities, and $K$ is a constant \cite{Zipf1946,Barbosa2018}. Usually, the masses $m_i$ and $m_j$ can be various factors such as population, gdp-per-capita, etc. The distance $d_{ij}$ can be measured in terms of geodesic distance, time, or monetary cost. Typically, a power or an exponential form is assumed for the deterrence function \cite{Barbosa2018}.

In this study, we collected \textit{net} annual flows traveling between prefectures as the mobility data $F_{ij}$. The geodesic distance was used as $d_{ij}$. The proportion of the infectious population in prefecture $i$ was denoted as $\rho_i$. The function $f(\cdot)$ was assumed to follow a power law. Denote 
\begin{equation}
\mathcal{O}_i = \sum_{\substack{j=1\\j\neq i}}^{N_p}F_{ij}, \quad \mathcal{I}_j = \sum_{\substack{i=1\\i\neq j}}^{N_p}F_{ij}, \qquad i,j = 1, 2, \dots, N_p,
\end{equation}
where $N_p = 47$ is the number of prefectures, $\mathcal{O}_i$ represents the total number of travelers leaving prefecture $i$, and $\mathcal{I}_j$ is the total number of travelers arriving at prefecture $j$.
According to \eqnref{eq_gravity}, for one person leaving prefecture $i$, the probability of his/her arrival at prefecture $j$ is
\begin{equation}
P_{ij} := \frac{F_{ij}}{\mathcal{O}_i} \approx k_i\mathcal{I}_jd^{\alpha_i}_{ij},\qquad i,j = 1, 2, \dots, N_p, ~i\neq j, \label{eq_tpij}
\end{equation}
and for one person arriving at prefecture $j$, the probability of coming from prefecture $i$ is
\begin{equation}
\quad Q_{ij} := \frac{F_{ij}}{\mathcal{I}_j} \approx s_j\mathcal{O}_id^{\beta_j}_{ij},\qquad i,j = 1, 2, \dots, N_p, ~i\neq j,\label{eq_tqij}
\end{equation}
where $\mathcal{O}_i$ and $ \mathcal{I}_j$ are considered as the masses, while $k_i$ and $\alpha_i$ are coefficients describing the willingness to travel of people in prefecture $i$, and $s_j$ and $\beta_j$ are coefficients describing the ability to attract tourists of prefecture $j$. These coefficients $k_i, \alpha_i, s_j$ and $\beta_j$ were estimated by fitting from the data $F_{ij}$. We denote the estimated normalized transition probabilities as $\widehat{P}$ and $\widehat{Q}$, where 
\begin{equation}
\widehat{P}_{ij} = \frac{k_i\mathcal{I}_jd_{ij}^{\alpha_i}}{\sum_{j=1}^{N_p}k_i\mathcal{I}_jd_{ij}^{\alpha_i}}, \quad \widehat{Q}_{ij} = \frac{s_j\mathcal{O}_id_{ij}^{\beta_j}}{\sum_{i=1}^{N_p}s_j\mathcal{O}_id_{ij}^{\beta_j}}, \label{eq_tpqhat}
\end{equation}
the row-sum of $\widehat{P}\in\mathbb{R}^{N_p\times N_p}$  is $1$, and the column-sum of $\widehat{Q}\in\mathbb{R}^{N_p\times N_p}$ is $1$.

Therefore, the number of infectious individuals $T_{ij}$ traveling from prefecture $i$ to $j$ can be estimated using either $\rho_i\mathcal{O}_i\widehat{P}_{ij}$ or $\rho_i\mathcal{I}_j\widehat{Q}_{ij}$. We used their geometric mean value, i.e.
\begin{equation}
T_{ij} = \sqrt{\rho_i\mathcal{O}_i\widehat{P}_{ij}\cdot \rho_i\mathcal{I}_j\widehat{Q}_{ij}}, \label{eq_Tij}
\end{equation}
where $\rho_i$ is the infectious rate of COVID-19 in prefecture $i$. According to Eqs.~\eqref{eq_tpij}-\eqref{eq_Tij}, we can observe that $T_{ij}$ follows a distribution $T_{ij}\sim d_{ij}^{(\alpha_i+\beta_j)/2}$. The \textit{effective distance} is defined as
\begin{equation}
D_{ij}^{\text{COVID}} = T_{ij}^{2/(\alpha_i+\beta_j)} =(\rho_i^2\mathcal{O}_i\mathcal{I}_j\widehat{P}_{ij}\widehat{Q}_{ij})^{\frac{2}{\alpha_i+\beta_j}} \label{eq_Deff}
\end{equation}
to maintain the dimensionality $D_{ij}^{\text{COVID}}\sim d_{ij}$. Such an effective distance $D_{ij}^{\text{COVID}}$ considers the effects of the geodesic distance, the human mobility, and the mass of communities. We used $D^{\text{COVID}}_{ij}$ as the baseline reference for evaluating the performance of causality inference.

\vspace{2em}
\textbf{Data collection and preprocessing for estimating the effective distance:}
\begin{itemize}
\item The \emph{net} human mobility data $F_{ij}$ among prefectures is available on the website of the Ministry of Land, Infrastructure, Transport and Tourism, as the result of the \emph{Inter-Regional Travel Survey in Japan}, conducted every five years, most recently in 2015 \cite{MLIT2019}. This survey counts the number of domestic passengers traveling across the borders of prefectures, using five inter-regional transportation modes including airlines,	railways, sea lines, buses, and cars. There are missing values within travels among major metropolitan areas in Japan (i.e., Tokyo area including Tokyo, Kanagawa, Chiba, and Saitama;
Kinki area including Osaka, Kyoto, Hyogo, and Nara;
Chukyo area including Aichi, Gifu, and Mie). We filled by regression according to another survey named \emph{Passenger Regional Flow Survey} conducted
every year, which targets the \emph{gross} annual passenger flow between prefectures	(the corresponding 2015 data is available in \cite{MLIT2017}).
The net flow considers the actual origin and destination of a passenger's trip, while the same trip gets separated by intermediate stops in the gross flow.

\item The geodesic distances $d_{ij}$ (measured between prefectural offices) are acquired from the website of the Geospatial Information Authority of Japan~\cite{GSI}. 
\item Given $F_{ij}$ and $d_{ij}$, the values of $k_i, s_j, \alpha_i$ and $\beta_j$ are determined for all $i$ and $j$ by fitting according to \eqnref{eq_tpij} and \eqnref{eq_tqij}. The exponents $\alpha_i$ and $\beta_j$ range between $-2.712$ and $-0.744$,	with a median of $-1.665$.

\item The coefficient $\rho_i$, showing the proportion of infectious passengers, is defined as the ratio of the cumulative number of confirmed cases to the population in a prefecture. The cumulative number is calculated from the daily confirmed new cases from \cite{MHLW2023}, while the population were collected from the counts by Japanese government in October 1st, 2021 \cite{SBJ2022}.
\end{itemize} 

\begin{landscape}
	\begin{figure}[!htbp]
		%\begin{sidewaysfigure}[htbp]
		\centering
		\includegraphics[height=0.95\textwidth]{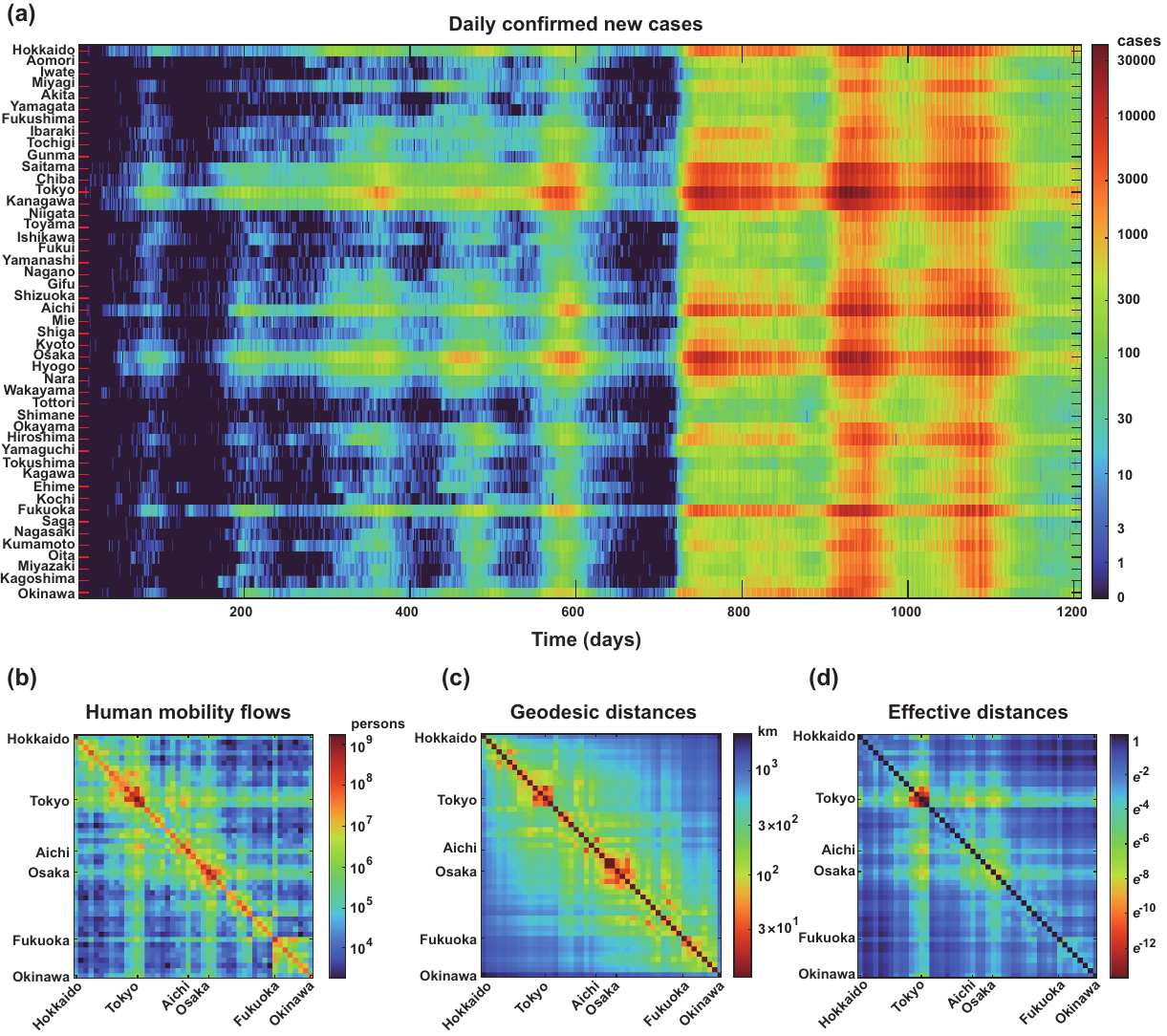}
		\caption{\label{fig:sm_covid_prep}
			Data involved in the analysis of COVID-19 transmission in Japan.
			(a) The time series of daily confirmed new cases in each prefecture, showing $8$ waves of infection from January 16, 2020 to May 8, 2023 (1209 days).
			(b) The net human mobility flows $F_{ij}$ (after completing missing values).
			(c) The geodesic distances $d_{ij}$ (in kilometers) between prefectures.
			(d) The effective distances $D^{\text{COVID}}_{ij}$ defined by \eqnref{eq_Deff} (shown in a logarithmic scale).
			$D^{\text{COVID}}_{ij}$ was used as a baseline reference for the causal inference.}
		%\end{sidewaysfigure}
	\end{figure}
\end{landscape}

The net human mobility flows $F_{ij}$, the geodesic distances $d_{ij}$, and the effective distances $D^{\text{COVID}}_{ij}$ are demonstrated in figs.~\ref{fig:sm_covid_prep}(b), (c), and (d), respectively. From \figref{fig:sm_covid_prep}(c), we observe that metropolitan areas with small geodesic distances appear as clusters of prefectures. Meanwhile, after considering the effect of human mobility, population sizes, and infectious rate, we obtained the effective distance matrix in \figref{fig:sm_covid_prep}(d), in which strips of relatively small effective distances connect those metropolitan areas and other prefectures.

\subsubsection{Causality Analysis}
We computed the causality strengths between any two nodes in the network (i.e., prefectures in Japan) with IEE/GC/TE/CCM. Each time series is divided into $3$ segments of equal lengths (keeping the sample rate) in order to achieve multiple evaluations of the indices. All segments were normalized to have zero mean and unit variance. To avoid zero values, a Gaussian noise with mean zero and standard variation $10^{-7}$ was added to the time series. Parameters in the algorithm were chosen as: delayed lag $L=3$, number of nearest neighbors $K=20$. 

\begin{figure}[htbp]
	\centering
	\includegraphics[width=\textwidth]{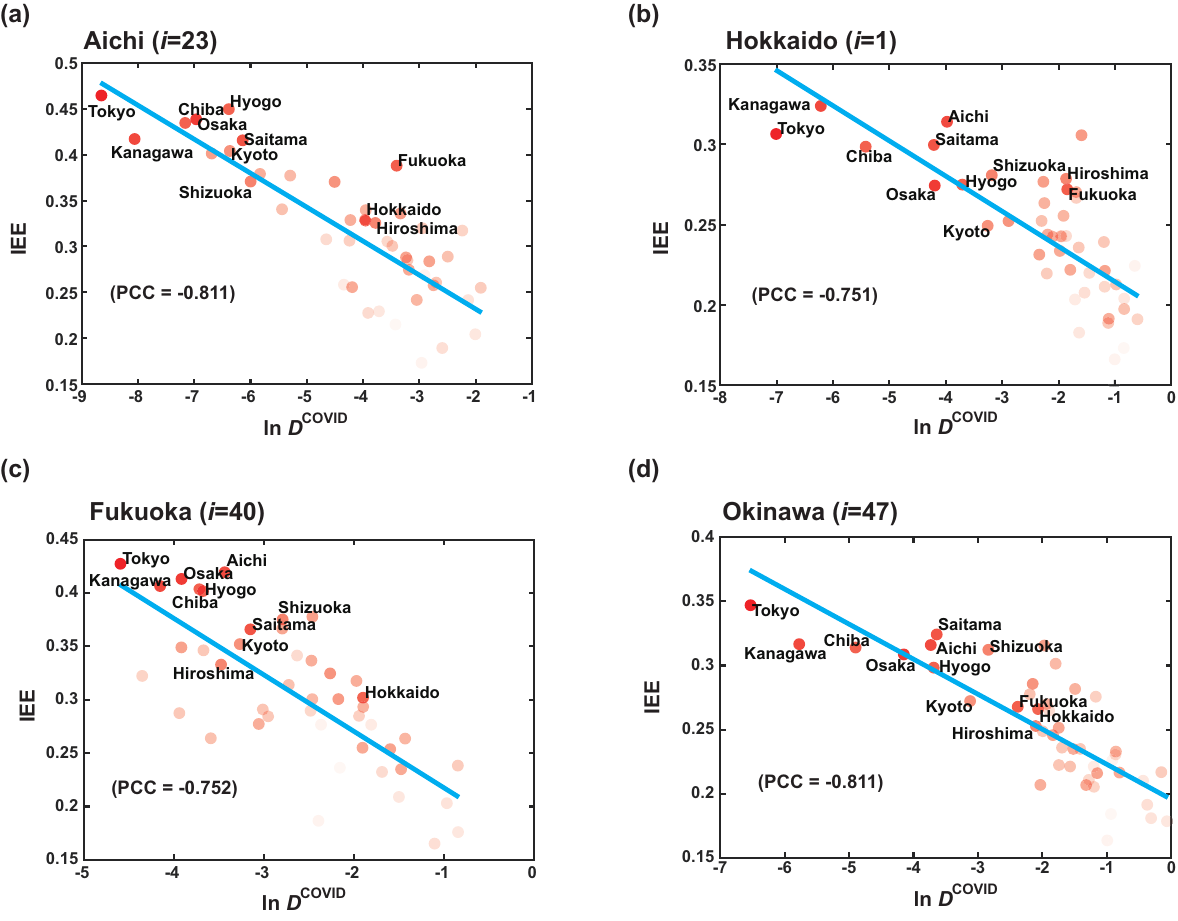}
	\caption{\label{fig:sm_covid_iee}
		Scatter plots of the inferred causal strengths IEE with respect
		to the logarithmic of effective distances $D^{\text{COVID}}_{ij}$, from Aichi/Hokkaido/Fukuoka/Okinawa to other prefectures. The color depth of dots represents the number of confirmed COVID-19 cases in the corresponding prefecture. The blue line is the least square line. Pearson correlation coefficient (PCC) is shown, and IEE can reflect the influence of disease transmission.}
\end{figure}

\begin{figure}[htbp]
	\centering
	\includegraphics[width=\textwidth]{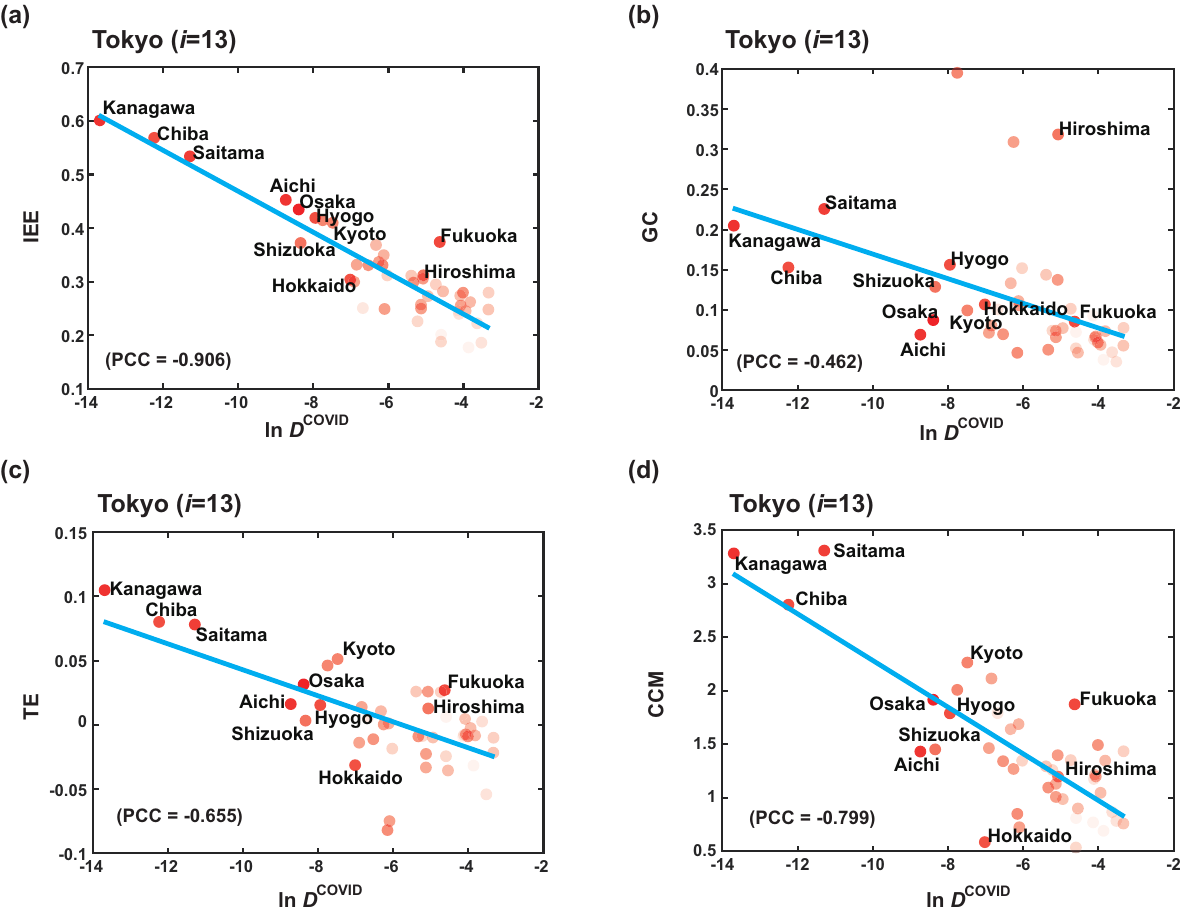}
	\caption{\label{fig:sm_covid_tokyo}
		Scatter plots of the inferred causal strengths IEE/GC/TE/CCM with respect to
		the logarithmic of effective distances, i.e. $\ln D^{\text{COVID}}_{ij}$, from Tokyo ($i=13$) to other prefectures.  The CCM index was transformed by $-\log(1-x)$ to scale its value from $0$ to $+\infty$, ensuring a consistent range as $D^{\text{COVID}}$ for comparison purposes. The color depth of dots represents the number of confirmed COVID-19 cases in the corresponding prefecture. The blue line is the least square line. Pearson correlation coefficient (PCC) is shown, and IEE is the best to reflect the influence of disease transmission.}
\end{figure}

IEE demonstrated highly linear correlation with  $\ln D^{\text{COVID}}$ with PCC $-0.906$ from Tokyo ($i=13$) to other other prefectures (Fig.~6(A) in the main text). Further, in Fig.~6(B) in the main text and \figref{fig:sm_covid_iee}, we exhibit five other prefectures including Osaka, Aichi, Hokkaido, Fukuoka, and Okinawa, where IEE showed linearity with $\ln D^{\text{COVID}}$. Figure~\ref{fig:sm_covid_tokyo} plots the causality strengths $c^{\text{Inf}}_{ij}$ given by IEE/GC/TE/CCM with respect to the effective distance $D_{ij}^{\text{COVID}}$ from Tokyo ($i=13$) to other other prefectures ($j\neq i$).  Table~3 in the main text lists an overall comparison, and IEE has an average PCC $0.748$ with the logarithm of $D^{\text{COVID}}$, higher than GC ($0.166$), TE ($0.109$), and CCM ($0.585$). In this example, the CCM index was transformed by $-\log(1-x)$ to scale its value from $0$ to $+\infty$, ensuring a consistent range as $D^{\text{COVID}}$ for comparison purposes. These results indicate that the proposed criterion $c^{\text{IEE}}$ is suitable for the quantitative analyses of the causality in the COVID-19 transmission dynamics in Japan.

\subsection{Circadian rhythm gene expression dataset}
We investigated the gene regulatory networks (GRNs) involving key genes related to circadian rhythm. The gene expression time series that were measured by Affimetrix microarray (Genechip Rat Genome 230 2.0) of the laboratory rat (Rattus norvegicus) cultured cells sampled from suprachiasmatic nucleus (SCN) for studying circadian rhythm \cite{wang2009network, kawaguchi2007establishment, morioka2008phase, ma2014detecting, leng2020partial}. We downloaded the dataset from \href{https://github.com/Partial-Cross-Mapping/circadian}{https://github.com/Partial-Cross-Mapping/circadian}, which contained the ground truth of GRN in the gene and protein level and four time series (with length 9, 16, 14, and 12) for gene expressions. After interpolating and concatenating, we obtained 98 time points for causal detection.

\begin{figure}[htbp]
	\centering
	\includegraphics[width=\textwidth]{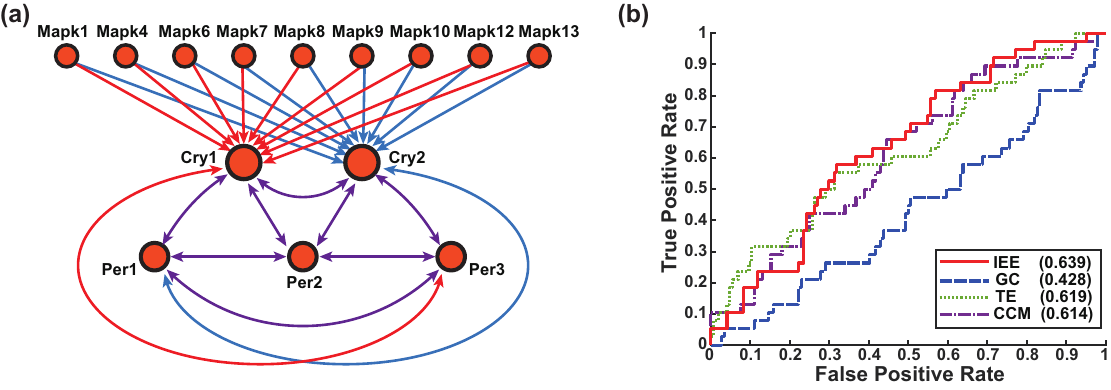}
	\caption{\label{fig:sm_circadian}
		Comparison results for detecting causality in the gene regulatory network (GRN) centered around \textit{Cry1} and \textit{Cry2}. (a) The ground truth of the GRN surrounding \textit{Cry1} and \textit{Cry2}. There are 14 genes in total. Key circadian genes \textit{Cry1, Cry2, Per1, Per2,} and \textit{Per3} are highlighted by large node sizes. Distinct colors are employed to differentiate crossing arrows. (b) The ROC curves of IEE (red solid line), GC (blue dashed line), TE (green dotted line), and CCM (purple dash-dot line). The AUC values for each method are listed in the legend.}
\end{figure}

Through decades of molecular and genetic studies \cite{ko2006molecular, ueda2005system}, lots of key circadian genes have been identified and extensively studied in mammals, including \textit{Bmal1(Arntl), Clock, Cry1, Cry2, Dec1(Bhlhb2), Dec2(Bhlhb3), Per1, Per2, Per3}. We focused on two subnetworks at the protein-protein interaction level in our experiment. One is surrounding \textit{Clock}, which is comprised by 12 genes (Fig.~6(D) in the main text). The transcription factor \textit{Clock} is phosphorylated by PFK family genes. The other network containing 14 genes is centered around \textit{Cry1/Cry2}, phosphorylated by MAPK family genes (\figref{fig:sm_circadian}(a)).

We thus applied IEE/GC/TE/CCM to the time-series data to detect the causal relationship between genes. The ROC curves and AUC values were computed and displayed in Fig.~6(E) for the \textit{Clock} network and \figref{fig:sm_circadian}(b) for the \textit{Cry1/Cry2} network. IEE designed for quantifying the IntDC performed higher AUC values than the other three ConDC indices. Parameters in algorithm were chosen as: delayed lag $L=3$, number of nearest neighbors $K=20$.

\subsection{Conditional interventional embedding entropy}
In the main text, we describe the IEE criterion for a two-variable dynamical system. To distinguish directed and indirect causality, we can generalize the system consisting of multiple variables. For simplicity, the dynamics of $y$ can be considered as
\begin{equation}
y_{t+1} = f(x_t, \dots, x_{t-p}, y_t, \dots, y_{t-p}, z_t, \dots, z_{t-p}, \varepsilon_{y,t}),
\end{equation}
and the embedding theorem ensures that
\begin{equation}
\vec{X}_t = \vec{F}(\vec{Y}_{t+1}, \vec{Z}_t),
\end{equation}
where $z_t$ is a third variable whose time-delay vector is $\vec{Z}_t$. Under an infinitesimal intervention, we can obtain
\begin{equation}
\delta\vec{X}_t = \nabla\vec{F}(\vec{Y}_{t+1}, \vec{Z}_t)\cdot(\delta\vec{Y}_{t+1}, \delta\vec{Z}_t).
\end{equation}
Thus, the IEE can be extended to its conditional version
\begin{equation}
\text{cIEE}[x\rightarrow y|z] = \text{CMI}(\delta\vec{X}_t, \delta\vec{Y}_{t+1}|\vec{Y}_{t+1},\vec{Z}_t, \delta\vec{Z}_t),
\end{equation}
where ``cIEE'' is short for conditional IEE. 
\newpage
\section{Supplementary Figures}

\newpage

\newpage

\begin{landscape}
	\section{Supplementary Tables}
	\begin{table}[!htbp]
%\begin{sidewaystable}[!htbp]
	\centering
	\caption{The mean values (standard deviations) of four causal indices from $y$ to $x$ when $\beta_{xy}=0$ in \figref{fig_2log00}.} \label{tab_s1}
	{\footnotesize{
	\begin{tabular}[\textwidth]{|c|c|c|c|c|c|}
		\hline
		\multicolumn{6}{|c|}{\makecell{When $\beta_{yx}=0$, $\text{IEE}[y\rightarrow x]= 9.42e-05~(9.16e-04)$, $\text{GC}[y\rightarrow x]= 2.67e-01~ (2.75e-02)$, \\$\text{TE}[y\rightarrow x]=2.01e-02~(7.54e-03)$,  $\text{CCM}[y\rightarrow x]=4.21e-02~(3.04e-02)$} }\\\hline
		$\beta_{yx}$	& $0.01$ & $0.02$ & $0.03$ & $0.04$ & $0.05$\\\hline
		IEE[$y\rightarrow x$] &$1.07e-04~(1.06e-03)$	&	$3.66e-04~(1.22e-03)$	&	$9.56e-04~(1.51e-03)$	&	$2.59e-03~(2.17e-03)$	&	$4.75e-03~(2.52e-03)$\\
		GC[$y\rightarrow x$]&$2.22e-01~(2.09e-02)$	&	$1.89e-01~(2.32e-02)$	&	$1.56e-01~(2.09e-02)$	&	$1.31e-01~(1.54e-02)$	&	$1.06e-01~(1.44e-02)$\\
		TE[$y\rightarrow x$]&$1.77e-02~(7.58e-03)$	&	$1.02e-02~(6.38e-03)$	&	$1.04e-02~(7.77e-03)$	&	$3.39e-02~(1.01e-02)$	&	$6.71e-02~(1.20e-02)$\\
		CCM[$y\rightarrow x$]&$6.79e-02~(4.09e-02)$	&	$2.02e-01~(4.98e-02)$	&	$3.63e-01~(4.25e-02)$	&	$4.95e-01~(3.60e-02)$	&	$5.93e-01~(2.81e-02)$\\\hline
		$\beta_{yx}$ & $0.06$ & $0.07$& $0.08$& $0.09$ &$0.10$\\\hline
		IEE[$y\rightarrow x$] & $7.86e-03~(2.96e-03)$	&	$1.24e-02~(3.98e-03)$	&	$1.91e-02~(4.72e-03)$	&	$2.67e-02~(5.69e-03)$	&	$3.69e-02~(6.33e-03)$\\
		GC[$y\rightarrow x$] &$8.50e-02~(1.38e-02)$	&	$7.00e-02~(1.23e-02)$	&	$5.77e-02~(1.08e-02)$	&	$4.34e-02~(1.02e-02)$	&	$3.44e-02~(9.11e-03)$\\
		TE[$y\rightarrow x$] &$1.09e-01~(1.27e-02)$	&	$1.52e-01~(1.37e-02)$	&	$1.95e-01~(1.54e-02)$	&	$2.40e-01~(1.53e-02)$	&	$2.80e-01~(1.63e-02)$\\
		CCM[$y\rightarrow x$] &$6.64e-01~(2.32e-02)$	&	$7.16e-01~(2.13e-02)$	&	$7.54e-01~(1.87e-02)$	&	$7.88e-01~(1.62e-02)$	&	$8.10e-01~(1.50e-02)$\\\hline
		$\beta_{yx}$	& $0.11$ & $0.12$ & $0.13$ & $0.14$ & $0.15$ \\\hline
		IEE[$y\rightarrow x$] &$4.69e-02~(8.07e-03)$	&	$5.78e-02~(7.42e-03)$	&	$7.23e-02~(9.30e-03)$	&	$8.93e-02~(1.06e-02)$	&	$1.04e-01~(1.20e-02)$\\
		GC[$y\rightarrow x$]&$2.37e-02~(8.51e-03)$	&	$1.44e-02~(7.11e-03)$	&	$8.24e-03~(4.96e-03)$	&	$3.21e-03~(2.75e-03)$	&	$1.67e-03~(1.73e-03)$\\
		TE[$y\rightarrow x$]&$3.16e-01~(1.74e-02)$	&	$3.43e-01~(1.66e-02)$	&	$3.65e-01~(1.47e-02)$	&	$3.84e-01~(1.37e-02)$	&	$3.93e-01~(1.72e-02)$\\
		CCM[$y\rightarrow x$]&$8.28e-01~(1.48e-02)$	&	$8.45e-01~(1.22e-02)$	&	$8.60e-01~(1.21e-02)$	&	$8.74e-01~(1.07e-02)$	&	$8.82e-01~(1.14e-02)$\\\hline
		$\beta_{yx}$ & $0.16$ & $0.17$& $0.18$& $0.19$&$0.20$\\\hline
		IEE[$y\rightarrow x$] &$1.19e-01~(1.16e-02)$	&	$1.32e-01~(1.47e-02)$	&	$1.45e-01~(1.18e-02)$	&	$1.56e-01~(1.19e-02)$	&	$1.72e-01~(1.33e-02)$\\
		GC[$y\rightarrow x$]&$2.31e-03~(1.64e-03)$	&	$7.33e-03~(3.81e-03)$	&	$(1.45e-02~(5.67e-03)$	&	$2.08e-02~(7.10e-03)$	&	$2.48e-02~(6.92e-03)$\\
		TE[$y\rightarrow x$]&$4.04e-01~(1.41e-02)$	&	$4.17e-01~(1.76e-02)$	&	$4.34e-01~(1.52e-02)$	&	$4.47e-01~(1.49e-02)$	&	$4.59e-01~(1.56e-02)$\\
		CCM[$y\rightarrow x$]&$8.84e-01~(9.86e-03)$	&	$8.89e-01~(1.02e-02)$	&	$8.93e-01~(9.71e-03)$	&	$9.02e-01~(8.59e-03)$	&	$9.09e-01~(9.67e-03)$\\\hline
		$\beta_{yx}$	& $0.21$ & $0.22$ & $0.23$ & $0.24$ & $0.25$ \\\hline
		IEE[$y\rightarrow x$] &$1.88e-01~(1.31e-02)$	&	$2.02e-01~(1.59e-02)$	&	$2.16e-01~(1.44e-02)$	&	$2.29e-01~(1.44e-02)$	&	$2.40e-01~(1.46e-02)$\\
		GC[$y\rightarrow x$]&$3.05e-02~(7.91e-03)$	&	$3.27e-02~(9.32e-03)$	&	$3.33e-02~(8.70e-03)$	&	$3.44e-02~(1.01e-02)$	&	$3.44e-02~(9.38e-03)$\\
		TE[$y\rightarrow x$]&$4.73e-01~(1.76e-02)$	&	$4.84e-01~(1.86e-02)$	&	$4.97e-01~(1.51e-02)$	&	$5.06e-01~(1.50e-02)$	&	$5.14e-01~(1.49e-02)$\\
		CCM[$y\rightarrow x$]&$9.17e-01~(8.29e-03)$	&	$9.25e-01~(8.79e-03)$	&	$9.32e-01~(7.10e-03)$	&	$9.39e-01~(6.87e-03)$	&	$9.46e-01~(6.60e-03)$\\\hline
		$\beta_{yx}$ & $0.26$ & $0.27$& $0.28$& $0.29$&$0.30$\\\hline
		IEE[$y\rightarrow x$] &$2.55e-01~(1.70e-02)$	&	$2.62e-01~(1.42e-02)$	&	$2.68e-01~(1.58e-02)$	&	$2.77e-01~(1.61e-02)$	&	$2.85e-01~(1.63e-02)$\\
		GC[$y\rightarrow x$]&$3.57e-02~(1.03e-02)$	&	$3.44e-02~(1.04e-02)$	&	$3.61e-02~(1.16e-02)$	&	$3.20e-02~(1.09e-02)$	&	$3.33e-02~(1.18e-02)$\\
		TE[$y\rightarrow x$]&$5.23e-01~(1.73e-02)$	&	$5.27e-01~(1.77e-02)$	&	$5.26e-01~(1.94e-02)$	&	$5.29e-01~(1.93e-02)$	&	$5.34e-01~(2.18e-02)$\\
		CCM[$y\rightarrow x$]&$9.50e-01~(6.87e-03)$	&	$9.56e-01~(7.11e-03)$	&	$9.61e-01~(6.17e-03)$	&	$9.68e-01~(5.61e-03)$	&	$9.72e-01~(5.54e-03)$\\\hline
	\end{tabular}}}
%\end{sidewaystable}
\end{table}
\end{landscape}
\newpage

\begin{landscape}
	\begin{table}[!htbp]
%\begin{sidewaystable}[!htbp]
	\centering
	\caption{The mean values (standard deviations) of four causal indices from $x$ to $y$ when $\beta_{xy}=0$ in \figref{fig_2log00}.}\label{tab_s2}
	{\footnotesize
	\begin{tabular}[\textwidth]{|c|c|c|c|c|c|}
		\hline
		\multicolumn{6}{|c|}{\makecell{When $\beta_{yx}=0$, $\text{IEE}[x\rightarrow y]= 1.50e-04~(8.61e-04)$, $\text{GC}[x\rightarrow y]= 2.60e-01~ (2.90e-02)$, \\$\text{TE}[x\rightarrow y]=2.20e-02~(7.80e-03)$,  $\text{CCM}[x\rightarrow y]=4.10e-02~(3.03e-02)$} }\\\hline
		$\beta_{yx}$	& $0.01$ & $0.02$ & $0.03$ & $0.04$ & $0.05$\\\hline
		IEE[$x\rightarrow y$] &$1.34e-04~(9.62e-04)$ & $2.23e-05~(1.17e-03)$& $2.25e-04~(1.43e-03)$& $1.97e-04~(1.52e-03)$&$3.48e-04~(1.89e-03)$\\
		GC[$x\rightarrow y$]&$2.78e-01~(2.49e-02)$ & $2.98e-01~(2.89e-02)$& $3.18e-01~(2.75e-02)$&$3.32e-01~(2.79e-02)$&$3.49e-01~(2.47e-02)$\\
		TE[$x\rightarrow y$]&$2.18e-02~(8.20e-03)$ & $2.03e-02~(8.05e-03)$ &$2.13e-02~(7.30e-03)$&$2.24e-02~(7.57e-03)$&$2.01e-02~(7.55e-03)$\\
		CCM[$x\rightarrow y$]&$4.19e-02~(2.47e-02)$ & $4.54e-02~(3.27e-02)$ &$4.93e-02~(3.15e-02)$&$4.81e-02~(2.98e-02)$&$4.74e-02~(4.32e-02)$\\\hline
		$\beta_{yx}$ & $0.06$ & $0.07$& $0.08$& $0.09$ &$0.10$\\\hline
		IEE[$x\rightarrow y$] & $5.70e-05~(2.36e-03)$&$2.18e-05~(2.19e-03)$&$9.17e-05~(2.61e-03)$&$3.04e-04~(2.64e-03)$&$1.30e-04~(3.37e-03)$\\
		GC[$x\rightarrow y$] &$3.63e-01~(2.83e-02)$&$3.78e-01~(2.85e-02)$&$3.84e-01~(3.26e-02)$&$3.92e-01~(3.11e-02)$&$3.89e-01~(2.78e-02)$\\
		TE[$x\rightarrow y$] &$2.03e-02~(7.32e-03)$&$2.04e-02~(7.72e-03)$&$1.85e-02~(7.78e-03)$&$1.89e-02~(8.19e-03)$&$1.81e-02~(6.66e-03)$\\
		CCM[$x\rightarrow y$] &$5.23e-02~(3.54e-02)$&$5.91e-02~(4.22e-02)$&$5.82e-02~(4.24e-02)$&$5.64e-02~(4.77e-02)$&$5.25e-02~(3.86e-02)$\\\hline
		$\beta_{yx}$	& $0.11$ & $0.12$ & $0.13$ & $0.14$ & $0.15$ \\\hline
		IEE[$x\rightarrow y$] &$2.44e-04~(3.40e-03)$ & $1.07e-03~(3.52e-03)$ & $4.58e-04~(3.42e-03)$ & $9.51e-04~(3.72e-03)$ & $9.80e-04~(3.42e-03)$\\
		GC[$x\rightarrow y$]&$4.00e-01~(2.99e-02)$ & $4.00e-01~(3.32e-02)$ & $4.00e-01~(3.25e-02)$ & $4.00e-01~(3.58e-02)$ & $3.82e-01~(3.85e-02)$\\
		TE[$x\rightarrow y$]&$1.77e-02~(6.98e-03)$ & $1.84e-02~(7.07e-03)$ & $1.70e-02~(7.89e-03)$ & $1.93e-02~(7.57e-03)$ & $1.76e-02~(7.63e-03)$\\
		CCM[$x\rightarrow y$]&$6.24e-02~(5.62e-02)$ & $6.75e-02~(5.63e-02)$ & $9.20e-02~(7.02e-02)$ & $9.88e-02~(7.22e-02)$ & $1.68e-01~(9.74e-02)$\\\hline
		$\beta_{yx}$ & $0.16$ & $0.17$& $0.18$& $0.19$&$0.20$\\\hline
		IEE[$x\rightarrow y$] &$6.97e-05~(2.44e-03)$ & $1.08e-03~(2.68e-03)$ & $2.12e-04~(2.82e-03)$ & $1.02e-03~(3.26e-03)$ & $7.20e-02~(2.77e-03)$\\
		GC[$x\rightarrow y$]&$3.64e-01~(2.92e-02)$ & $3.38e-01~(3.17e-02)$ & $3.25e-01~(3.12e-02)$ & $3.04e-01~(2.91e-02)$ & $2.91e-01~(2.89e-02)$\\
		TE[$x\rightarrow y$]&$1.89e-02~(6.83e-03)$ & $1.69e-02~(7.40e-03)$ & $1.85e-02~(7.63e-03)$ & $1.68e-02~(6.88e-03)$ & $1.72e-02~(8.70e-03)$\\
		CCM[$x\rightarrow y$]&$2.29e-01~(9.05e-02)$ & $3.52e-01~(8.07e-02)$ & $4.19e-01~(6.69e-02)$ & $5.08e-01~(5.44e-02)$ & $5.71e-01~(5.15e-02)$\\\hline
		$\beta_{yx}$	& $0.21$ & $0.22$ & $0.23$ & $0.24$ & $0.25$ \\\hline
		IEE[$x\rightarrow y$] &$1.17e-03~(3.00e-03)$ & $1.30e-03~(3.26e-03)$ & $2.22e-03~(4.05e-03)$ & $2.82e-03~(4.09e-03)$ & $2.89e-03~(4.70e-03)$\\
		GC[$x\rightarrow y$]&$2.80e-01~(2.65e-02)$ & $2.64e-01~(3.34e-02)$ & $2.54e-01~(2.78e-02)$ & $2.36e-01~(3.01e-02)$ & $2.22e-01~(3.01e-02)$\\
		TE[$x\rightarrow y$]&$1.77e-02~(6.70e-03)$ & $1.86e-02~(6.57e-03)$ & $1.74e-02~(7.94e-03)$ & $1.76e-02~(8.16e-03)$ & $1.73e-02~(7.51e-03)$\\
		CCM[$x\rightarrow y$]&$6.14e-01~(4.87e-02)$ & $6.67e-01~(5.24e-02)$ & $6.98e-01~(3.47e-02)$ & $7.33e-01~(3.46e-02)$ & $7.64e-01~(2.88e-02)$\\\hline
		$\beta_{yx}$ & $0.26$ & $0.27$& $0.28$& $0.29$&$0.30$\\\hline
		IEE[$x\rightarrow y$] &$3.60e-03~(4.18e-03)$ & $4.79e-03~(4.48e-03)$ & $5.39e-03~(5.48e-03)$ & $6.40e-03~(5.13e-03)$ & $7.48e-03~(5.27e-03)$\\
		GC[$x\rightarrow y$]&$2.14e-01~(3.05e-02)$ & $1.97e-01~(2.76e-02)$ & $1.85e-01~(2.83e-02)$ & $1.64e-01~(2.77e-02)$ & $1.59e-01~(2.99e-02)$\\
		TE[$x\rightarrow y$]&$1.84e-02~(8.08e-03)$ & $1.69e-02~(8.32e-03)$ & $1.75e-02~(8.22e-03)$ & $1.92e-02~(8.06e-03)$ & $1.89e-02~(8.76e-03)$\\
		CCM[$x\rightarrow y$]&$7.85e-01~(3.18e-02)$ & $8.14e-01~(3.03e-02)$ & $8.35e-01~(2.81e-02)$ & $8.60e-01~(2.55e-02)$ & $8.77e-01~(2.42e-02)$\\\hline
	\end{tabular}
}
%\end{sidewaystable}
\end{table}
\end{landscape}
\newpage

\begin{landscape}
	\begin{table}[!htbp]
%\begin{sidewaystable}[!htbp]
	\centering
	\caption{The mean values (standard deviations) of four causal indices from $y$ to $x$ when $\beta_{xy}=0.1$ in \figref{fig_2log01}.}\label{tab_s3}
	{\footnotesize
	\begin{tabular}[\textwidth]{|c|c|c|c|c|c|}
		\hline
		\multicolumn{6}{|c|}{\makecell{When $\beta_{yx}=0$, $\text{IEE}[y\rightarrow x]= 2.24e-04~(1.68e-03)$, $\text{GC}[y\rightarrow x]= 2.52e-01~ (2.38e-02)$, \\$\text{TE}[y\rightarrow x]=2.10e-02~(7.35e-03)$,  $\text{CCM}[y\rightarrow x]=2.64e-01~(5.55e-02)$} }\\\hline
		$\beta_{yx}$	& $0.01$ & $0.02$ & $0.03$ & $0.04$ & $0.05$\\\hline
		IEE[$y\rightarrow x$] &$1.16e-04~(1.67e-03)$	&	$7.81e-04~(1.82e-03)$	&	$1.79e-03~(2.01e-03)$	&	$3.68e-03~(2.66e-03)$	&	$6.24e-03~(2.71e-03)$\\
		GC[$y\rightarrow x$]&$2.27e-01~(2.43e-02)$	&	$1.98e-01~(2.07e-02)$	&	$1.77e-01~(1.93e-02)$	&	$1.52e-01~(1.85e-02)$	&	$1.26e-01~(1.96e-02)$\\
		TE[$y\rightarrow x$]&$1.81e-02~(6.47e-03)$	&	$1.03e-02~(7.21e-03)$	&	$8.56e-03~(6.59e-03)$	&	$2.48e-02~(1.14e-02)$	&	$5.14e-02~(1.17e-02)$\\
		CCM[$y\rightarrow x$]&$3.01e-01~(5.10e-02)$	&	$4.05e-01~(4.83e-02)$	&	$5.03e-01~(4.24e-02)$	&	$6.01e-01~(3.33e-02)$	&	$6.71e-01~(2.61e-02)$\\\hline
		$\beta_{yx}$ & $0.06$ & $0.07$& $0.08$& $0.09$ &$0.10$\\\hline
		IEE[$y\rightarrow x$] & $1.00e-02~(4.00e-03)$	&	$1.44e-02~(3.86e-03)$	&	$2.11e-02~(4.88e-03)$	&	$2.94e-02~(5.09e-03)$	&	$4.19e-02~(6.86e-03)$\\
		GC[$y\rightarrow x$] &$1.04e-01~(1.69e-02)$	&	$7.90e-02~(1.63e-02)$	&	$5.68e-02~(1.56e-02)$	&	$3.88e-02~(1.02e-02)$	&	$2.40e-02~(7.48e-03)$\\
		TE[$y\rightarrow x$] &$8.45e-02~(1.31e-02)$	&	$1.20e-01~(1.35e-02)$	&	$1.61e-01~(1.47e-02)$	&	$2.03e-01~(1.59e-02)$	&	$2.40e-01~(1.52e-02)$\\
		CCM[$y\rightarrow x$] &$7.23e-01~(2.39e-02)$	&	$7.66e-01~(1.91e-02)$	&	$7.98e-01~(1.81e-02)$	&	$8.20e-01~(1.83e-02)$	&	$8.45e-01~(1.75e-02)$\\\hline
		$\beta_{yx}$	& $0.11$ & $0.12$ & $0.13$ & $0.14$ & $0.15$ \\\hline
		IEE[$y\rightarrow x$] &$5.41e-02~(9.36e-03)$	&	$7.19e-02~(9.84e-03)$	&	$8.93e-02~(1.15e-02)$	&	$1.07e-01~(8.68e-03)$	&	$1.28e-01~(1.16e-02)$\\
		GC[$y\rightarrow x$]&$1.71e-02~(6.09e-03)$	&	$1.16e-02~(3.32e-03)$	&	$9.26e-03~(5.65e-03)$	&	$5.30e-03~(3.76e-03)$	&	$2.13e-03~(2.33e-03)$\\
		TE[$y\rightarrow x$]&$2.70e-01~(1.86e-02)$	&	$3.00e-01~(1.34e-02)$	&	$3.22e-01~(1.70e-02)$	&	$3.40e-01~(1.46e-02)$	&	$3.54e-01~(1.65e-02)$\\
		CCM[$y\rightarrow x$]&$8.67e-01~(1.57e-02)$	&	$8.87e-01~(1.29e-02)$	&	$9.06e-01~(9.28e-03)$	&	$9.23e-01~(9.13e-03)$	&	$9.39e-01~(7.44e-03)$\\\hline
		$\beta_{yx}$ & $0.16$ & $0.17$& $0.18$& $0.19$&$0.20$\\\hline
		IEE[$y\rightarrow x$] &$1.44e-01~(1.26e-02)$	&	$1.56e-01~(1.39e-02)$	&	$1.62e-01~(1.42e-02)$	&	$1.68e-01~(1.39e-02)$	&	$1.78e-01~(1.31e-02)$\\
		GC[$y\rightarrow x$]&$3.23e-03~(3.56e-03)$	&	$7.42e-03~(5.14e-03)$	&	$1.20e-02~(6.38e-03)$	&	$1.48e-02~(6.59e-03)$	&	$1.66e-02~(7.31e-03)$\\
		TE[$y\rightarrow x$]&$3.60e-01~(1.94e-02)$	&	$3.74e-01~(1.94e-02)$	&	$3.76e-01~(2.05e-02)$	&	$3.79e-01~(2.34e-02)$	&	$3.86e-01~(2.17e-02)$\\
		CCM[$y\rightarrow x$]&$9.49e-01~(6.94e-03)$	&	$9.57e-01~(5.78e-03)$	&	$9.64e-01~(4.15e-03)$	&	$9.70e-01~(3.69e-03)$	&	$9.74e-01~(3.43e-03)$\\\hline
		$\beta_{yx}$	& $0.21$ & $0.22$ & $0.23$ & $0.24$ & $0.25$ \\\hline
		IEE[$y\rightarrow x$] &$1.82e-01~(1.47e-02)$	&	$1.89e-01~(1.46e-02)$	&	$1.93e-01~(1.30e-02)$	&	$2.03e-01~(1.50e-02)$	&	$2.09e-01~(1.23e-02)$\\
		GC[$y\rightarrow x$]&$1.89e-02~(7.54e-03)$	&	$1.86e-02~(7.85e-03)$	&	$2.09e-02~(8.17e-03)$	&	$2.05e-02~(7.67e-03)$	&	$2.01e-02~(6.96e-03)$\\
		TE[$y\rightarrow x$]&$3.83e-01~(2.65e-02)$	&	$3.80e-01~(2.52e-02)$	&	$3.76e-01~(2.62e-02)$	&	$3.77e-01~(2.47e-02)$	&	$3.78e-01~(2.08e-02)$\\
		CCM[$y\rightarrow x$]&$9.78e-01~(2.89e-03)$	&	$9.81e-01~(2.42e-03)$	&	$9.84e-01~(1.97e-03)$	&	$9.86e-01~(1.77e-03)$	&	$9.87e-01~(1.57e-03)$\\\hline
		$\beta_{yx}$ & $0.26$ & $0.27$& $0.28$& $0.29$&$0.30$\\\hline
		IEE[$y\rightarrow x$] &$2.13e-01~(1.25e-02)$	&	$2.15e-01~(1.29e-02)$	&	$2.25e-01~(1.05e-02)$	&	$2.25e-01~(1.39e-02)$	&	$2.32e-01~(1.34e-02)$\\
		GC[$y\rightarrow x$]&$1.82e-02~(8.20e-03)$	&	$1.84e-02~(7.83e-03)$	&	$1.65e-02~(6.42e-03)$	&	$1.72e-02~(7.68e-03)$	&	$1.56e-02~(7.53e-03)$\\
		TE[$y\rightarrow x$]&$3.71e-01~(2.33e-02)$	&	$3.71e-01~(2.23e-02)$	&	$3.72e-01~(1.86e-02)$	&	$3.66e-01~(2.11e-02)$	&	$3.66e-01~(1.88e-02)$\\
		CCM[$y\rightarrow x$]&$9.89e-01~(1.34e-03)$	&	$9.90e-01~(1.08e-03)$	&	$9.92e-01~(8.64e-04)$	&	$9.93e-01~(8.94e-04)$	&	$9.93e-01~(7.65e-04)$\\\hline
	\end{tabular}
}
%\end{sidewaystable}
\end{table}
\end{landscape}
\newpage

\begin{landscape}
\begin{table}[!htbp]
%\begin{sidewaystable}[!htbp]
	\centering
	\caption{The mean values (standard deviations) of four causal indices from $x$ to $y$ when $\beta_{xy}=0.1$ in \figref{fig_2log01}.}\label{tab_s4}
	{\footnotesize
	\begin{tabular}[\textwidth]{|c|c|c|c|c|c|}
		\hline
		\multicolumn{6}{|c|}{\makecell{When $\beta_{yx}=0$, $\text{IEE}[x\rightarrow y]= 1.68e-01~(1.35e-02)$, $\text{GC}[x\rightarrow y]= 3.91e-01~ (4.07e-02)$, \\$\text{TE}[x\rightarrow y]=4.06e-01~(1.67e-02)$,  $\text{CCM}[x\rightarrow y]=9.61e-01~(3.05e-03)$} }\\\hline
		$\beta_{yx}$	& $0.01$ & $0.02$ & $0.03$ & $0.04$ & $0.05$\\\hline
		IEE[$x\rightarrow y$] &$1.67e-01~(1.41e-02)$	&	$1.59e-01~(1.17e-02)$	&	$1.58e-01~(1.31e-02)$	&	$1.49e-01~(1.18e-02)$	&	$1.46e-01~(1.26e-02)$\\
		GC[$x\rightarrow y$]&$3.85e-01~(4.48e-02)$	&	$3.97e-01~(4.46e-02)$	&	$4.02e-01~(3.77e-02)$	&	$4.15e-01~(4.60e-02)$	&	$4.19e-01~(4.78e-02)$\\
		TE[$x\rightarrow y$]&$3.99e-01~(1.96e-02)$	&	$3.79e-01~(1.75e-02)$	&	$3.61e-01~(1.99e-02)$	&	$3.35e-01~(2.36e-02)$	&	$3.17e-01~(2.28e-02)$\\
		CCM[$x\rightarrow y$]&$9.62e-01~(2.96e-03)$	&	$9.62e-01~(3.00e-03)$	&	$9.64e-01~(3.16e-03)$	&	$9.66e-01~(3.00e-03)$	&	$9.67e-01~(2.84e-03)$\\\hline
		$\beta_{yx}$ & $0.06$ & $0.07$& $0.08$& $0.09$ &$0.10$\\\hline
		IEE[$x\rightarrow y$] & $1.45e-01~(1.52e-02)$	&	$1.42e-01~(1.29e-02)$	&	$1.39e-01~(1.52e-02)$	&	$1.38e-01~(1.41e-02)$	&	$1.42e-01~(1.64e-02)$\\
		GC[$x\rightarrow y$] &$4.26e-01~(4.72e-02)$	&	$4.17e-01~(4.51e-02)$	&	$4.24e-01~(4.77e-02)$	&	$4.22e-01~(4.50e-02)$	&	$4.08e-01~(4.70e-02)$\\
		TE[$x\rightarrow y$] &$2.99e-01~(2.55e-02)$	&	$2.92e-01~(2.40e-02)$	&	$2.74e-01~(2.89e-02)$	&	$2.67e-01~(2.49e-02)$	&	$2.67e-01~(2.57e-02)$\\
		CCM[$x\rightarrow y$] &$9.69e-01~(2.65e-03)$	&	$9.70e-01~(2.34e-03)$	&	$9.71e-01~(2.61e-03)$	&	$9.73e-01~(2.37e-03)$	&	$9.75e-01~(2.44e-03)$\\\hline
		$\beta_{yx}$	& $0.11$ & $0.12$ & $0.13$ & $0.14$ & $0.15$ \\\hline
		IEE[$x\rightarrow y$] &$1.45e-01~(1.44e-02)$	&	$1.47e-01~(1.69e-02)$	&	$1.54e-01~(1.51e-02)$	&	$1.56e-01~(1.43e-02)$	&	$1.61e-01~(1.43e-02)$\\
		GC[$x\rightarrow y$]&$3.94e-01~(5.21e-02)$	&	$3.79e-01~(4.33e-02)$	&	$3.61e-01~(4.47e-02)$	&	$3.34e-01~(4.14e-02)$	&	$2.99e-01~(3.75e-02)$\\
		TE[$x\rightarrow y$]&$2.69e-01~(2.85e-02)$	&	$2.68e-01~(2.56e-02)$	&	$2.77e-01~(2.42e-02)$	&	$2.86e-01~(2.32e-02)$	&	$3.04e-01~(2.07e-02)$\\
		CCM[$x\rightarrow y$]&$9.75e-01~(2.19e-03)$	&	$9.76e-01~(2.25e-03)$	&	$9.76e-01~(2.02e-03)$	&	$9.77e-01~(1.96e-03)$	&	$9.76e-01~(2.07e-03)$\\\hline
		$\beta_{yx}$ & $0.16$ & $0.17$& $0.18$& $0.19$&$0.20$\\\hline
		IEE[$x\rightarrow y$] &$1.51e-01~(1.38e-02)$	&	$1.46e-01~(1.45e-02)$	&	$1.34e-01~(1.26e-02)$	&	$1.28e-01~(1.30e-02)$	&	$1.21e-01~(1.27e-02)$\\
		GC[$x\rightarrow y$]&$2.72e-01~(3.70e-02)$	&	$2.62e-01~(3.69e-02)$	&	$2.39e-01~(3.76e-02)$	&	$2.16e-01~(3.67e-02)$	&	$2.06e-01~(3.41e-02)$\\
		TE[$x\rightarrow y$]&$3.10e-01~(1.43e-02)$	&	$3.08e-01~(1.62e-02)$	&	$2.91e-01~(1.44e-02)$	&	$2.75e-01~(2.05e-02)$	&	$2.56e-01~(2.23e-02)$\\
		CCM[$x\rightarrow y$]&$9.75e-01~(1.87e-03)$	&	$9.76e-01~(1.77e-03)$	&	$9.77e-01~(1.77e-03)$	&	$9.77e-01~(2.07e-03)$	&	$9.78e-01~(1.70e-03)$\\\hline
		$\beta_{yx}$	& $0.21$ & $0.22$ & $0.23$ & $0.24$ & $0.25$ \\\hline
		IEE[$x\rightarrow y$] &$1.15e-01~(1.12e-02)$	&	$1.12e-01~(1.08e-02)$	&	$1.06e-01~(9.21e-03)$	&	$1.06e-01~(1.08e-02)$	&	$1.02e-01~(1.09e-02)$\\
		GC[$x\rightarrow y$]&$1.86e-01~(3.37e-02)$	&	$1.66e-01~(3.74e-02)$	&	$1.47e-01~(3.73e-02)$	&	$1.32e-01~(3.32e-02)$	&	$1.17e-01~(3.06e-02)$\\
		TE[$x\rightarrow y$]&$2.32e-01~(2.18e-02)$	&	$2.13e-01~(2.20e-02)$	&	$1.93e-01~(1.96e-02)$	&	$1.79e-01~(1.89e-02)$	&	$1.64e-01~(1.56e-02)$\\
		CCM[$x\rightarrow y$]&$9.79e-01~(1.59e-03)$	&	$9.80e-01~(1.45e-03)$	&	$9.81e-01~(1.47e-03)$	&	$9.82e-01~(1.57e-03)$	&	$9.83e-01~(1.52e-03)$\\\hline
		$\beta_{yx}$ & $0.26$ & $0.27$& $0.28$& $0.29$&$0.30$\\\hline
		IEE[$x\rightarrow y$] &$1.02e-01~(1.01e-02)$	&	$9.79e-02~(1.00e-02)$	&	$9.85e-02~(1.04e-02)$	&	$9.74e-02~(9.77e-03)$	&	$1.01e-01~(9.39e-03)$\\
		GC[$x\rightarrow y$]&$1.01e-01~(3.14e-02)$	&	$8.70e-02~(2.79e-02)$	&	$7.54e-02~(2.40e-02)$	&	$6.66e-02~(2.08e-02)$	&	$6.29e-02~(2.40e-02)$\\
		TE[$x\rightarrow y$]&$1.49e-01~(1.59e-02)$	&	$1.34e-01~(1.43e-02)$	&	$1.26e-01~(1.35e-02)$	&	$1.17e-01~(1.23e-02)$	&	$1.10e-01~(1.36e-02)$\\
		CCM[$x\rightarrow y$]&$9.84e-01~(1.46e-03)$	&	$9.85e-01~(1.15e-03)$	&	$9.86e-01~(1.40e-03)$	&	$9.87e-01~(1.33e-03)$	&	$9.88e-01~(1.03e-03)$\\\hline
	\end{tabular}
}
%\end{sidewaystable}
\end{table}
\end{landscape}
\newpage

\begin{table}[!htbp]
	\caption{\label{tab:sm_celegans_clustering}
		Clustering results of the neurons in \textit{C. elegans}.}
	\centering
	\begin{threeparttable}
		\begin{tabular}{cl}
			\toprule\hline
			\multicolumn{1}{c}{Clusters} & \multicolumn{1}{c}{Neurons}\\
			\midrule
			\#1&~~\underline{OLQDL, OLQDR}, \underline{OLQVL, OLQVR}, \underline{AIBL,  AIBR}, AVER, SABVL\\
			%%%%%%%%%%
			\#2&~~RIFR\\
			%%%%%%%%%%
			\#3&~~RIBL, RID, AVBL, RMEV, DB01\\
			%%%%%%%%%%
			\#4&~~\underline{RIML, RIMR}, \underline{AVAL, AVAR}, VA01\\
			%%%%%%%%%%
			\#5&~~AVBR, \underline{RMEL, RMER}, RMED, DB02, VB01, VB02\\
			%%%%%%%%%%
			\#6&~~\underline{RIVL,  RIVR}, \underline{SMDVL, SMDVR}\\
			%%%%%%%%%%
			\#7&~~SMBDR\\
			\hline\bottomrule
		\end{tabular}
		\begin{tablenotes}[flushleft]\footnotesize%\smallskip
			\item[$\ast$] The underlined pairs of neurons are symmetric in position and are clustered in the same cluster.
		\end{tablenotes}
	\end{threeparttable}
\end{table}

%\begin{table}[!htbp]
%	\centering
%	\caption{\label{tab:sm_celegans_performance}
	%		Numerical causal inference of the \emph{C. elegans} neural network.}
%	\begin{threeparttable}
	%		\begin{tabular}{cllllcll}
		%			\toprule\hline
		%			~& \multicolumn{4}{c}{Properties of ROC curves for binary classification}
		%			&& \multicolumn{2}{c}{Cosine Similarity} \\
		%			\cline{2-5}\cline{7-8}
		%			& \footnotesize{AUC~~} 
		%			& \footnotesize{Highest Youden index~~}
		%			& \footnotesize{Highest concordance probability~~}
		%			& \footnotesize{Minimum distance to $(0, 1)$~~}
		%			&& \footnotesize{to $c_{ij}$~~~~} & \footnotesize{to $\log(1+c_{ij})$~~}\\
		%			\midrule
		%			$\vec{C}^{\text{IEE}}$
		%			~&$0.882^\ast$   &$0.784^\ast$   &$0.793^\ast$   &$0.166^\ast$&&$0.750^\ast$ &~~$0.905^\ast$\\
		%			$\vec{C}^{\text{GC}}$
		%			~&$0.858$        &$0.721$        &$0.740$        &$0.198$        && $0.430$        &~~$0.678$      \\
		%			$\vec{C}^{\text{TE}}$
		%			~&$0.796$        &$0.620$        &$0.656$        &$0.269$        && $0.477$        &~~$0.726$      \\
		%			$\vec{C}^{\text{CCM}}$
		%			~&$0.880$        &$0.615$        &$0.639$        &$0.294$        && $0.644$        &~~$0.866$      \\
		%			\hline\bottomrule
		%		\end{tabular}
	%		\begin{tablenotes}[flushleft]\footnotesize
		%		\item $^\ast$ The best value in its column.
		%	\end{tablenotes}
	%	\end{threeparttable}
%\end{table}

\begin{table}[!htbp]
	\caption{\label{tab:sm_celegans_OOPperformance}
		The results for IEE/GC/TE/CCM at their minimum distance OOPs,  and the results for PCMCI-ParCorr (with $\alpha_{\text{PC}} = 0.05$) and PCMCI-CMI (with $\alpha_{\text{PC}} = 0.01$), when inferring the \emph{C.\,elegans} neural connectomes.}
	\centering
	\begin{threeparttable}
		\begin{tabular}{ccccc}
			\toprule\hline
			& True Positive & True Negative & False Positive & False Negative\\
			\midrule
			IEE&$22$   &$15$   &$1$   &$4$\\
			GC&$22$   &$14$   &$2$   &$4$\\
			TE &$21$   &$13$   &$3$   &$5$\\
			CCM&$22$&$12$  &$4$    &$4$\\
			PCMCI-ParCorr & $22$&$11$ &$5$ & $4$\\
			PCMCI-CMI &$22$&$10$ &$6$ & $4$\\
			\hline\bottomrule
		\end{tabular}
		\begin{tablenotes}[flushleft]\footnotesize
			\item $^\ast$ The true connectomes consist of $7$ nodes ($42$ potential pairs of neurons), with $26$ directed causal edges ($16$ non-causal pairs).
		\end{tablenotes}
	\end{threeparttable}
\end{table}
\clearpage\newpage
\bibliography{scibib}
\bibliographystyle{Science}

%\clearpage